%% file: main.tex
\documentclass{article}

\ifdefined\arxivbuild
  \usepackage[preprint]{neurips_2026}
\else
  \usepackage[main, final]{neurips_2026}
\fi

\usepackage[utf8]{inputenc}
\usepackage[T1]{fontenc}
\usepackage{microtype}
\usepackage{graphicx}
\usepackage{subcaption}
\usepackage{booktabs}
\usepackage{xcolor}
\usepackage{hyperref}
\hypersetup{pdftitle={Corrective Diffusion Language Models},pdfauthor={Shuibai Zhang, Fred Zhangzhi Peng, Yiheng Zhang, Jin Pan, Grigorios G. Chrysos}}
\usepackage{url}
\usepackage{nicefrac}
\usepackage{amsfonts}
\usepackage{amsmath}
\usepackage{amssymb}
\usepackage{mathtools}
\usepackage{amsthm}
\usepackage{bm}
\usepackage[capitalize,noabbrev]{cleveref}
\usepackage[ruled,vlined]{algorithm2e}
\usepackage{wrapfig}
\usepackage{etoolbox}
\usepackage{caption}

\theoremstyle{plain}

\theoremstyle{definition}

\newtheorem*{definition*}{Definition}
\theoremstyle{remark}

\title{Corrective Diffusion Language Models}

\author{%
  Shuibai Zhang \\
  University of Wisconsin--Madison \\
  \texttt{shuibai@cs.wisc.edu} \\
  \And
  Fred Zhangzhi Peng \\
  Duke University \\
  \And
  Yiheng Zhang \\
  University of Wisconsin--Madison \\
  \AND
  Jin Pan \\
  University of Wisconsin--Madison \\
  \And
  Grigorios G.~Chrysos \\
  University of Wisconsin--Madison \\
}

\begin{document}

\maketitle

\begin{abstract}
While Diffusion Language Models (DLMs) are theoretically well-suited for iterative refinement due to their non-causal structure, off-the-shelf masked diffusion language models (MDLMs) often fail to reliably revise incorrect tokens in practice. The key challenge lies in the model's inability to distinguish between correct and erroneous tokens in a visible sequence. Standard MDLM training supervises only masked positions, undermining the effectiveness of refinement guided by confidence. Based on this observation, we study \emph{corrective behavior} in DLMs, defined as the ability to assign lower confidence to incorrect tokens and iteratively refine them while preserving correct content. Under matched compute and data, we show that conventional masked-diffusion fine-tuning induces much less of this capability than a correction-oriented post-training principle that explicitly supervises visible corrupted tokens, which enables discriminative confidence and targeted refinement. We further find that also supervising clean visible tokens, as in BERT- or GIDD-style training, stops remask-based refinement from improving with additional steps.
To evaluate corrective behavior, we introduce the Code Revision Benchmark, a controllable and executable benchmark for assessing error localization and in-place correction. The effect transfers from a 0.5B model to LLaDA-8B and persists, with a smaller margin, when corruptions are drawn from the base model's own most plausible predictions. Beyond in-place correction, the same error-aware confidence also makes parallel decoding more effective. In HumanEval(+) code completion at 0.5B, mathematical reasoning at 16B, and high-entropy parallel decoding with remasking (ParallelBench), models trained with our objective outperform standard masked-diffusion fine-tuning under matched compute, and outperform the base model on HumanEval(+), MATH, and the high-entropy ParallelBench average; at 8B, neither objective improves from-scratch generation over the base model. Code, checkpoints and the CRB data are available at \url{https://github.com/zhangshuibai/CDLM}.
\end{abstract}

\section{Introduction}

Autoregressive (AR) language models dominate text generation by predicting tokens left to right~\citep{achiam2023gpt, guo2025deepseek, comanici2025gemini, yang2025qwen3technicalreport, grattafiori2024llama3herdmodels}, while diffusion language models (DLMs) replace causal generation with iterative denoising over complete sequences~\citep{nie2025largelanguagediffusionmodels, ye2025dream7bdiffusionlarge, labs2025mercuryultrafastlanguagemodels, song2025seeddiffusionlargescalediffusion, zhu2025lladamoesparsemoediffusion}. This non-causal formulation decouples token predictions across positions and enables parallel generation~\citep{kang2025parallelbench}, and supports localized, iterative \emph{in-place} revision~\citep{wang2025remasking, peng2025path}: arbitrary positions can be re-predicted in parallel without regenerating the surrounding context, whereas an AR model must emit any edit token by token.

\begin{figure}[!htbp]
    \centering
    \includegraphics[width=0.5\linewidth]{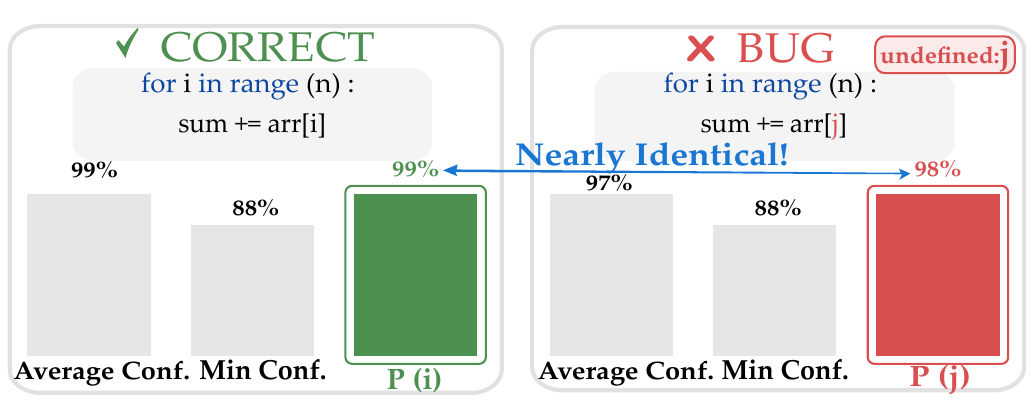}
    \caption{An illustrative example in which the model is \textbf{equally confident} in the correct variable ($i$) and the undefined bug ($j$). Global metrics (Average/Min) fail to detect this failure mode.}
    \label{fig:correct-confidence}
\end{figure}

Despite this structural advantage, off-the-shelf masked diffusion language models (MDLMs) do not reliably refine: the conventional masked-diffusion objective supervises only masked positions, leaving the model without a direct learning signal to distinguish correct from incorrect visible tokens (Figure~\ref{fig:correct-confidence}). This is a property of the supervision pattern rather than of a particular pretraining history, and it applies both to MDLMs pretrained from scratch and to MDLMs adapted from autoregressive checkpoints. To make this gap precise, we define \emph{corrective behavior} as the ability to identify unreliable tokens in a full sequence and iteratively revise them while preserving correct content.

A second challenge is evaluation: existing benchmarks focus on prefix completion~\citep{chen2021evaluatinglargelanguagemodels, austin2021mbpp, liu2023humanevalplus} and do not directly probe localization or in-place fixing. We address this with a controlled executable benchmark and a matched-compute post-training principle, and refer to MDLMs trained to exhibit corrective behavior as \emph{Corrective Diffusion Language Models (CDLMs)}.

\paragraph{Our contributions.}
\begin{itemize}
\item \textbf{Code Revision Benchmark (CRB)}\footnote{Code, checkpoints and the CRB instances: \url{https://github.com/zhangshuibai/CDLM} and \url{https://huggingface.co/datasets/Shuibai12138/crb-paper-inputs} (Appendix~\ref{app:release}).}, a controllable executable benchmark applying type-preserving corruptions of controllable difficulty to real code, with execution against the benchmark's tests as ground truth, isolating error localization and iterative in-place correction from prefix-based generation.
\item \textbf{Which positions to supervise, given the decoder.} Combining absorbing and uniform noise is not new~\citep{rutte2025generalized}. Our contribution is the supervision target: under matched data and compute, supervising visible corrupted tokens provides the missing signal for error-aware confidence, while additionally supervising clean visible tokens (as in BERT- or GIDD-style training) stops iterative remask-then-repredict refinement from improving with depth. Because only the set of supervised positions changes, the principle applies as post-training to an existing MDLM; the effect of supervising corrupted tokens holds at 0.5B, transfers to LLaDA-8B, and persists under model-plausible corruptions.
\item \textbf{Improved generation under high uncertainty.} Beyond CRB, CDLMs improve from-scratch HumanEval(+) generation at 0.5B, high-entropy parallel decoding with remasking, and mathematical reasoning at the 16B scale. At 0.5B they remain below the base model on MBPP(+), and at 8B neither fine-tuned arm improves from-scratch generation over base.
\end{itemize}

\section{Background}

\subsection{Masked Diffusion Language Models}

Let $\bm{x} = (x_1,\dots,x_n)$ be a clean token sequence over a vocabulary $\mathcal{V}$, and let $m \notin \mathcal{V}$ denote an absorbing mask token.
We focus on \emph{masked diffusion language models} (MDLMs)~\citep{sahoo2024simple}, a widely used class of DLMs~\citep{nie2025largelanguagediffusionmodels, ye2025dream7bdiffusionlarge, labs2025mercuryultrafastlanguagemodels, song2025seeddiffusionlargescalediffusion, zhu2025lladamoesparsemoediffusion} based on an absorbing corruption process in which tokens are progressively replaced by $m$.
\paragraph{Training.} For each example, a mask ratio $\lambda \sim \mathrm{Uniform}[0,1]$ is sampled and each token is independently replaced by $m$ with probability $\lambda$, giving the corrupted sequence $\bm{z}$. A Transformer $p_\theta(\bm{x}\mid\bm{z})$ is trained with the masked-token reconstruction loss $\mathcal{L}_{\mathrm{absorb}}(\theta)=\mathbb{E}_{\bm{x},\bm{z}}\big[-\sum_{i : z_i = m} \log p_\theta(x_i \mid \bm{z})\big]$: only masked positions contribute, and the logits at visible positions are ignored.

\subsection{Confidence-Based Iterative Refinement}
\label{bg:refinement}
DLMs refine sequences by iteratively predicting token distributions and selectively remasking uncertain positions~\citep{peng2025path, wang2025remasking, kim2025finetuningmaskeddiffusionprovable}. Given $\bm{z}^{(t)}$, one forward pass predicts every position; each masked position is filled with $\arg\max_v p_\theta(v\mid\bm{z}^{(t)})$ and every visible position keeps its token, giving $\tilde{\bm{z}}^{(t)}$. The \emph{confidence} of position $i$ is the probability the model assigns to the token that occupies it, $c_i^{(t)}=p_\theta(\tilde z_i^{(t)}\mid\bm{z}^{(t)})$, which for a visible token is $p_\theta(z_i\mid\bm{z})$ rather than the maximum over the vocabulary. Positions with $c_i^{(t)}\le\tau$ are remasked and all others are kept: $z_i^{(t+1)}=m$ if $c_i^{(t)}\le\tau$, else $\tilde z_i^{(t)}$. A visible token is therefore never rewritten directly; it is first remasked and then re-predicted. We call one remask-then-repredict update a refinement step; $T$ steps cost at most $T{+}1$ full-sequence forward passes, since a final pass fills the remaining masks (Algorithm~\ref{alg:confidence_refinement}).

\section{Code Revision Benchmark (CRB)}
\label{sec:crb}
The prefix-completion benchmarks used to evaluate current DLMs~\citep{nie2025largelanguagediffusionmodels, song2025seeddiffusionlargescalediffusion, ni2025trainingoptimallargediffusion} do not probe error localization or in-place fixing. A refinement benchmark needs controlled error types, fine-grained corruption severity, and deterministic verification. Code is an ideal testbed because correctness is executable. CRB builds on HumanEval~\citep{chen2021evaluatinglargelanguagemodels}, MBPP~\citep{austin2021mbpp}, HumanEval+/MBPP+~\citep{liu2023humanevalplus} and introduces controlled localized corruptions for systematic evaluation of localization and iterative in-place correction.

\paragraph{Task definition.}
Let $\bm{x}^\star=(x^\star_1,\dots,x^\star_n)$ be a canonical program. CRB selects an index set $E\subseteq\{1,\dots,n\}$ and applies a type-preserving replacement $\phi(\cdot)$, producing $z^{(0)}_i=\phi(x^\star_i)$ for $i\in E$ and $z^{(0)}_i=x^\star_i$ otherwise. The operator $\phi$ replaces a token with another token from the same lexical category while preserving token length under the tokenizer.

\subsection{Corruption Types}

\begin{figure}[!htbp]
    \centering
    \includegraphics[width=0.66\linewidth]{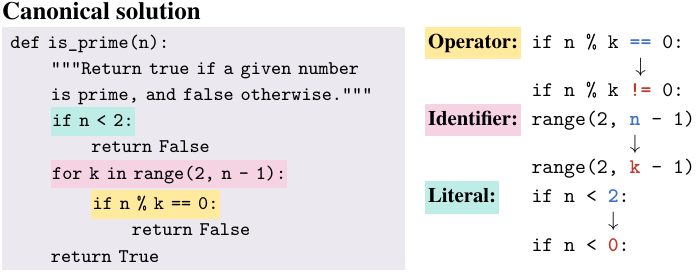}
    \caption{CRB corruption types. A canonical program (left, HumanEval/31) is corrupted by type-preserving token replacements of three kinds (right; original token in blue, replacement in red). A corrupted program enters CRB only if it fails the tests.}
    \label{fig:crb-pipeline}
\end{figure}

As shown in Figure~\ref{fig:crb-pipeline}, CRB includes three categories of token-level corruption:
\begin{itemize}
    \item \textbf{Operator substitutions}. An operator token is replaced by another operator of the same tokenized length from the set $\mathcal{O} = \{+,-,*,/,<,>,<=,>=,==,!=,<<,>>\}$.
    \item \textbf{Identifier substitutions}. A non-keyword identifier token (a variable, function or built-in name) is replaced with another identifier of the same tokenized length that appears in the same function body; Python keywords, including \texttt{True}, \texttt{False} and \texttt{None}, are never substituted. Such substitutions often induce subtle semantic inconsistencies.
    \item \textbf{Literal substitutions}. A numeric literal is replaced with another numeric literal of the same tokenized length from the same function body, introducing incorrect boundary conditions or erroneous behaviors.
\end{itemize}

\paragraph{Difficulty control and executable validation.}
Severity is controlled by the number of corrupted positions $|E|\!\in\!\{1,\ldots,5\}$, with $|E|\!=\!1$ a single-error case and $|E|\!>\!1$ requiring coordinated multi-step correction. Since not every type-preserving substitution actually breaks the program, after corruption we execute $\bm{z}^{(0)}$ against the dataset's tests and \emph{retain only those that fail}, guaranteeing that every CRB instance contains a genuine syntactic or semantic fault, verified by execution.

\section{Conventional Confidence-Based Corrective Behavior in DLMs}
\label{sec:phenomenon}

\input{Tables_Code/Benchmarking/benchmark_comparison}

To motivate our study of \emph{corrective behavior} in existing DLMs on code error correction, we evaluate several publicly available DLMs on CRB, including LLaDA-8B-Base \citep{nie2025largelanguagediffusionmodels}, Dream-7B-Base \citep{ye2025dream7bdiffusionlarge}, and Open-dCoder-0.5B \citep{opendllm2025}. LLaDA-8B-Base is pretrained from scratch as a masked diffusion model, whereas Dream-7B-Base and Open-dCoder-0.5B are obtained by continued masked-diffusion training of autoregressive Qwen2.5 checkpoints~\citep{qwen2025qwen25technicalreport}.
Our analysis focuses on two complementary aspects of corrective behavior:
(i) the ability to identify erroneous tokens within a complete input, and
(ii) the ability to correct such errors through iterative, in-place refinement.
All evaluations use the confidence-threshold-based refinement of Section~\ref{bg:refinement} on the function body, with the signature and docstring as fixed context, and greedy decoding (temperature 0) for every model (Appendix~\ref{app:hparams}).
Models are tested across multiple difficulty levels by varying both the number of corrupted tokens $|E|$ and the error type.

The results in this section serve as \emph{motivating evidence} that off-the-shelf MDLMs exhibit limited corrective behavior under controlled CRB conditions; they are not a controlled ablation of the training objective. The matched-data, matched-compute objective comparison appears in Section~\ref{sec:experiments}. Self-revision on each model's own correctly generated code (Appendix~\ref{sec:self_revision_minimal}) and model-plausible corruptions (Appendix~\ref{app:robustness}) make distributional unfamiliarity an unlikely sole explanation.

\subsection{Error-Token Identification Performance}
\label{bench:error_identification}
We evaluate whether DLMs can identify erroneous tokens via the confidence $c_i = p_{\theta}(z^{(0)}_i \mid \bm{z}^{(0)})$ of Section~\ref{bg:refinement}, computed on the corrupted program in a single forward pass. We report the \emph{confidence gap} $\mathrm{Gap}=\mathbb{E}_{i\notin E}[c_i]-\mathbb{E}_{i\in E}[c_i]$ (separation between clean and erroneous tokens), the \emph{Top-$K$ hit rate} $\mathrm{Hit@}K=\mathbb{I}(E\cap\mathrm{Bottom}_K(c)\neq\varnothing)$, where $\mathrm{Bottom}_K(c)$ is the set of the $K$ lowest-confidence body positions (whether at least one true error is among them), and, in Section~\ref{sec:ablations}, the \emph{Top-5 false positive rate}, the fraction of $\mathrm{Bottom}_5(c)$ that is clean.

\paragraph{Results.}
Table~\ref{tab:confidence_refinement_benchmark} summarizes error-token identification across corruption levels. Confidence gaps are small for LLaDA-8B-Base ($0.235$) and Open-dCoder-0.5B ($0.106$). Dream-7B-Base shows a large mean gap ($0.578$) that does not translate into reliable ranking: its top-1 hit rate ($0.094$) is the lowest of the three. In all models erroneous tokens are rarely ranked most uncertain (low top-1, higher top-5), indicating that token-level confidence in standard MDLMs is only coarsely aligned with correctness and insufficient for reliable error localization.

\subsection{Error-Correction Ability via Iterative Refinement}

We next evaluate whether DLMs can repair corrupted programs through iterative, in-place refinement. We follow the confidence-threshold-based protocol of Section~\ref{bg:refinement}: at each step $t$, positions with $c_i^{(t)}\le\tau$ are remasked and re-predicted; we set $\tau=0.9$ throughout. After $T$ steps (at most $T{+}1$ forward passes; a sample's output is final once no editable position has confidence at or below $\tau$), the program is executed and we report Pass@1, the fraction of programs that pass all tests.

\paragraph{Results.}
Table~\ref{tab:confidence_refinement_benchmark} reports Pass@1 after confidence-threshold-based iterative refinement.
Across all models and corruption levels, error-correction performance remains limited, even in the single-error setting.
No model repairs every program even with a single corrupted token (Pass@1 at $T=1$ from $0.143$ for Open-dCoder-0.5B to $0.776$ for Dream-7B-Base). Additional refinement steps help LLaDA-8B-Base and Open-dCoder-0.5B, mostly at the second step (e.g., $0.143\to0.230\to0.251$ for Open-dCoder-0.5B at $n_{\mathrm{replace}}=1$), but not Dream-7B-Base ($0.776\to0.732$): because low-confidence positions do not reliably correspond to true errors (Section~\ref{bench:error_identification}), iterative remasking often overwrites correct tokens while leaving actual errors unresolved.

\begin{figure}[!htbp]
    \centering
    \includegraphics[width=0.52\linewidth]{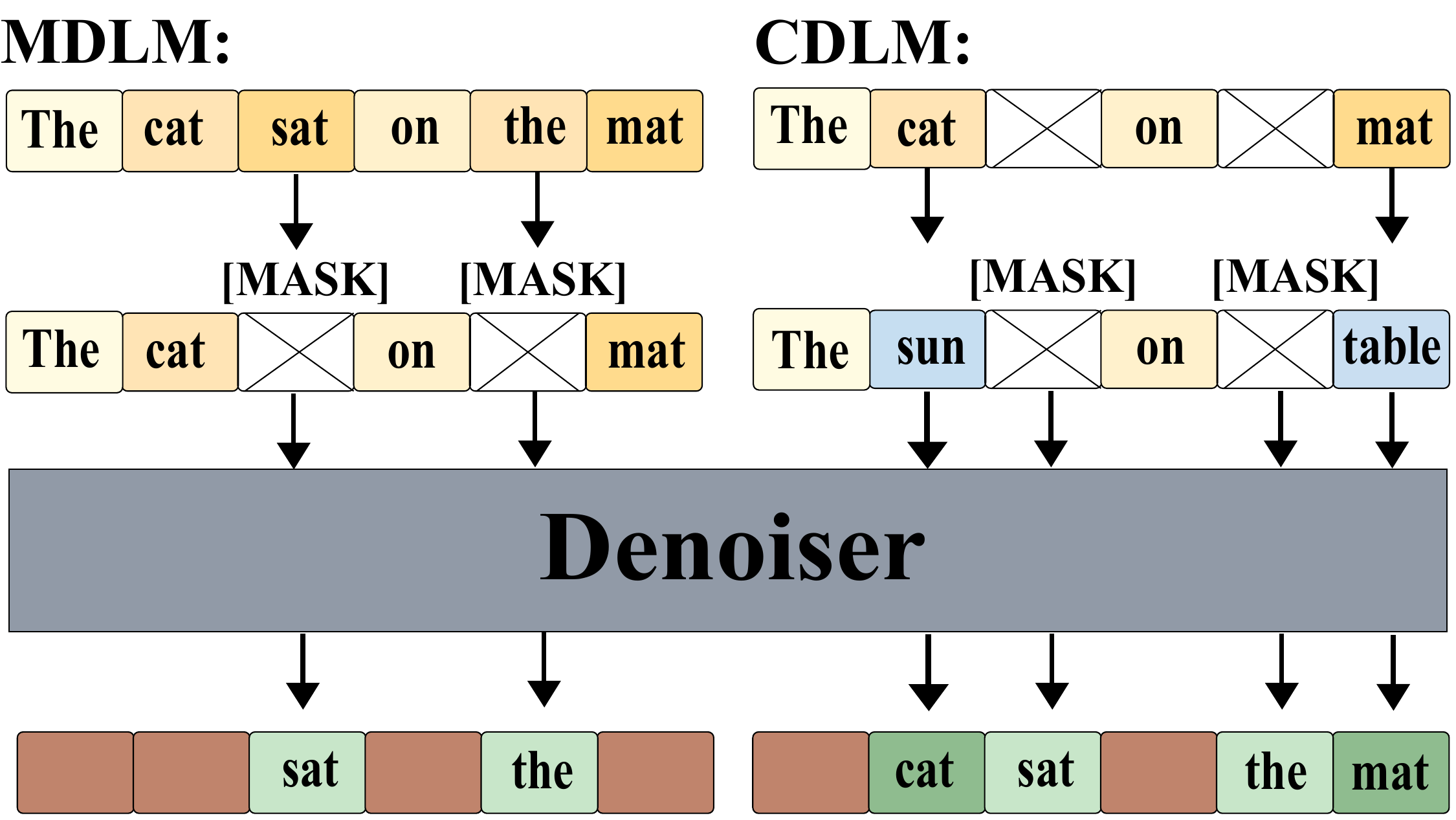}
    \caption{Comparison of training supervision in MDLM and CDLM and its implication for corrective behavior.
Cross-marked boxes denote masked tokens, and beige boxes denote visible inputs.
Standard MDLMs apply supervision only to masked tokens, leaving visible tokens unsupervised even when incorrect.
CDLM explicitly supervises corrupted visible tokens and encourages lower confidence on unreliable content.
This supervision difference enables error-aware confidence and targeted, in-place correction during iterative refinement.}
    \label{fig:confidence-illustration}
\end{figure}

\section{Corrective Post-Training via Mixture Noise}
\label{sec:mixture}

Under the absorbing-mask objective, unmasked tokens receive no direct supervision (Figure~\ref{fig:confidence-illustration}), so nothing in the objective trains the confidence of a visible token to reflect whether it is correct, and the model struggles to identify where edits are needed.

\begin{definition*}[Corrective Diffusion Language Model]
A \emph{Corrective Diffusion Language Model (CDLM)} is a DLM that (i) assigns systematically lower confidence to erroneous tokens than to correct ones, and (ii) leverages this confidence signal to localize errors and iteratively improve sequence correctness, enabling targeted, in-place refinement.
\end{definition*}

To induce this behavior, we add a uniform replacement corruption term to the standard absorbing-mask objective. Given clean $\bm{x}^\star$, we first mask a uniformly sampled subset of tokens (as in MDLM), then independently replace each remaining visible token with a uniformly sampled incorrect token with probability $\alpha\in(0,1)$. Let $\mathcal{M}$ denote the masked positions, $\mathcal{C}$ the visible positions, $n_i\sim\mathrm{Bernoulli}(\alpha)$ the replacement indicator of $i\in\mathcal{C}$, and $\mathcal{N}=\{i\in\mathcal{C}: n_i=1\}$. The training objective is
\begin{align}
\mathcal{L}
=
\mathbb{E}_{\text{corruption}}
\Bigg[
\mathcal{L}_{\mathrm{mask}}
\;+\;
\lambda_{\mathrm{noise}}
\frac{1}{|\mathcal{N}|}
\sum_{i \in \mathcal{N}} \ell_i
\Bigg],
\qquad
\mathcal{L}_{\mathrm{mask}} = \frac{1}{|\mathcal{M}|}\sum_{i \in \mathcal{M}} \ell_i ,
\label{eq:mixture-loss}
\end{align}
where $\ell_i=-\log p_\theta(x^\star_i\mid\bm{z})$ is the per-token cross-entropy against the clean token. The first term is standard MDLM denoising; the second supervises only visible tokens that have been corrupted, training the model to assign low probability to the token that is there. Clean visible tokens ($\mathcal{C}\setminus\mathcal{N}$) receive no loss; Section~\ref{sec:ctw} tests this choice by adding a clean-token term with weight $\mathrm{ctw}$. Normalizing the corruption term makes $\lambda_{\mathrm{noise}}$ independent of sequence length. Implementations differ in how $\mathcal{L}_{\mathrm{mask}}$ is weighted and normalized (the 0.5B trainer, inherited from Open-dLLM, adds a probability-weighted $1/t$ term in both arms) and in whether corrupted positions also enter its average; Appendix~\ref{app:objective_impl} gives the loss of each trainer. All language-model experiments use $\alpha=0.1$ and $\lambda_{\mathrm{noise}}=0.1$, except where the $\alpha$ sweep varies $\alpha$. We fixed $\alpha=0.1$ from preliminary experiments with $\alpha\in\{0.1,0.2,\ldots,1.0\}$ before running the sweep in Section~\ref{sec:ablations}, and did not revisit it.

\paragraph{Why uniform replacement.}
Uniform replacement is the simplest unbiased visible-corruption operator, and \citet{ni2025trainingoptimallargediffusion} report that uniform DLMs learn to tell noisy from clean tokens without being told which positions are corrupted; our objective imports this distinguishability into an MDLM.

\paragraph{Relation to GIDD.}
The Generalized Interpolating Discrete Diffusion framework \citep{rutte2025generalized} also combines absorbing and uniform noise, so the combination itself is not our contribution. What differs is the supervision target and the decoding semantics (Table~\ref{tab:gidd_vs_cdlm} in Appendix~\ref{sec:app_A}). GIDD- and BERT-style training supervise all positions, including clean visible tokens, and GIDD decoding may change every token at every step. CDLM supervises only masked and corrupted positions and keeps remask-then-repredict decoding, which is what makes localized in-place revision possible. Under such decoding, supervising clean tokens is not neutral: in Section~\ref{sec:ctw} it stops refinement from improving with depth, an effect that can only appear when edits are gated on confidence. Because only the set of supervised positions changes, CDLM applies as post-training of an existing MDLM (2000 steps at 0.5B and 8B). Since Open-dCoder-0.5B and Dream-7B are adapted from autoregressive checkpoints, we also verify the mechanism in a fully from-scratch controlled setting (Sudoku, Appendix~\ref{app:sudoku_exp}: at noise ratio $0.1$ the clean-to-noisy confidence ratio exceeds $10{,}000\times$ for CDLM against $8.2\times$ for MDLM).

\begin{figure}[!htbp]
    \centering
    \begin{subfigure}[b]{0.52\linewidth}
        \centering
        \includegraphics[width=\linewidth]{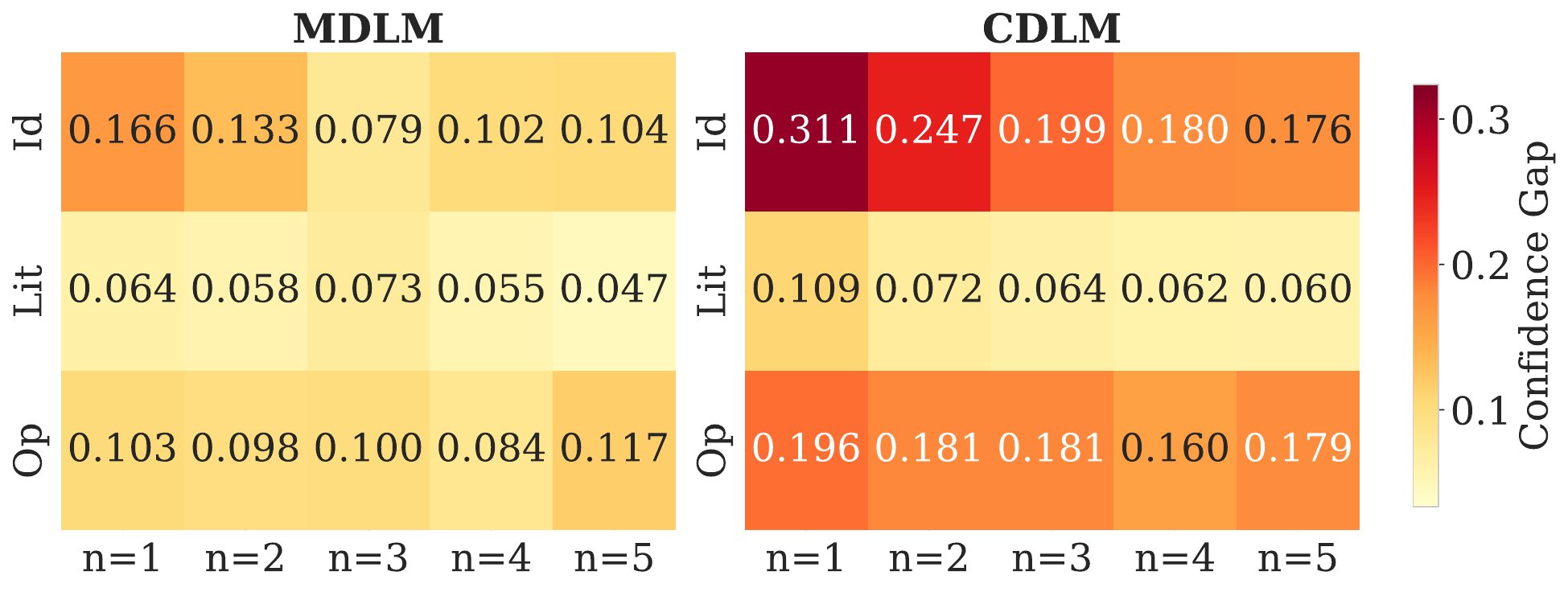}
        \caption{Confidence gap (clean vs.\ corrupted).}
        \label{fig:mixture-separation}
    \end{subfigure}
    \hfill
    \begin{subfigure}[b]{0.46\linewidth}
        \centering
        \includegraphics[width=\linewidth]{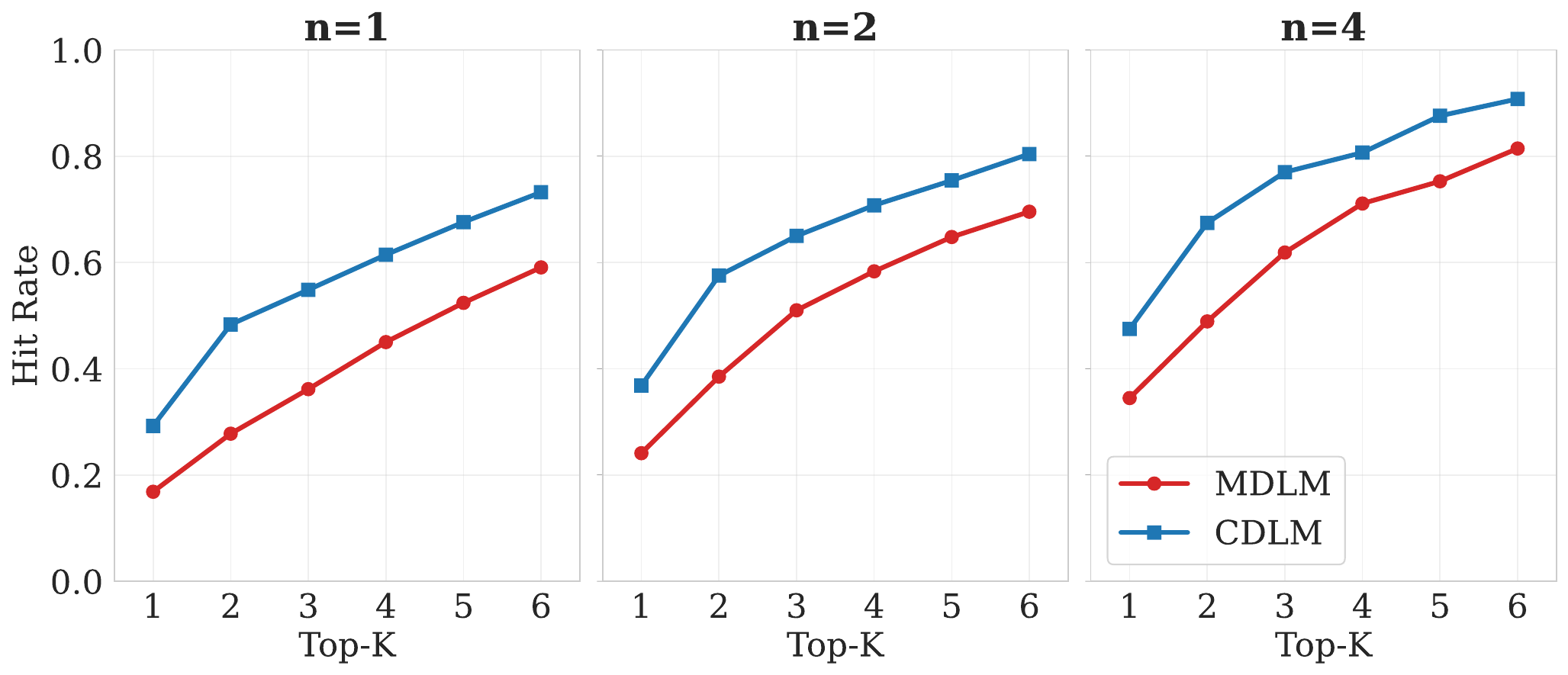}
        \caption{Top-$K$ hit rate.}
        \label{fig:topk-hit-comparison}
    \end{subfigure}
    \caption{Error localization on HumanEval. (a) Confidence gap between clean and corrupted tokens across error types (Id: identifier, Lit: literal, Op: operator) and replacement counts; CDLM's gap exceeds MDLM's in 14 of 15 cells (exception: Lit, $n=3$). (b) Top-$K$ hit rate of identifying at least one true error among the $K$ lowest-confidence positions at $n_{\text{replace}}\in\{1,2,4\}$; CDLM exceeds MDLM at every $K$.}
    \label{fig:error_localization}
\end{figure}

\begin{figure}[!htbp]
    \centering
    \includegraphics[width=0.55\linewidth]{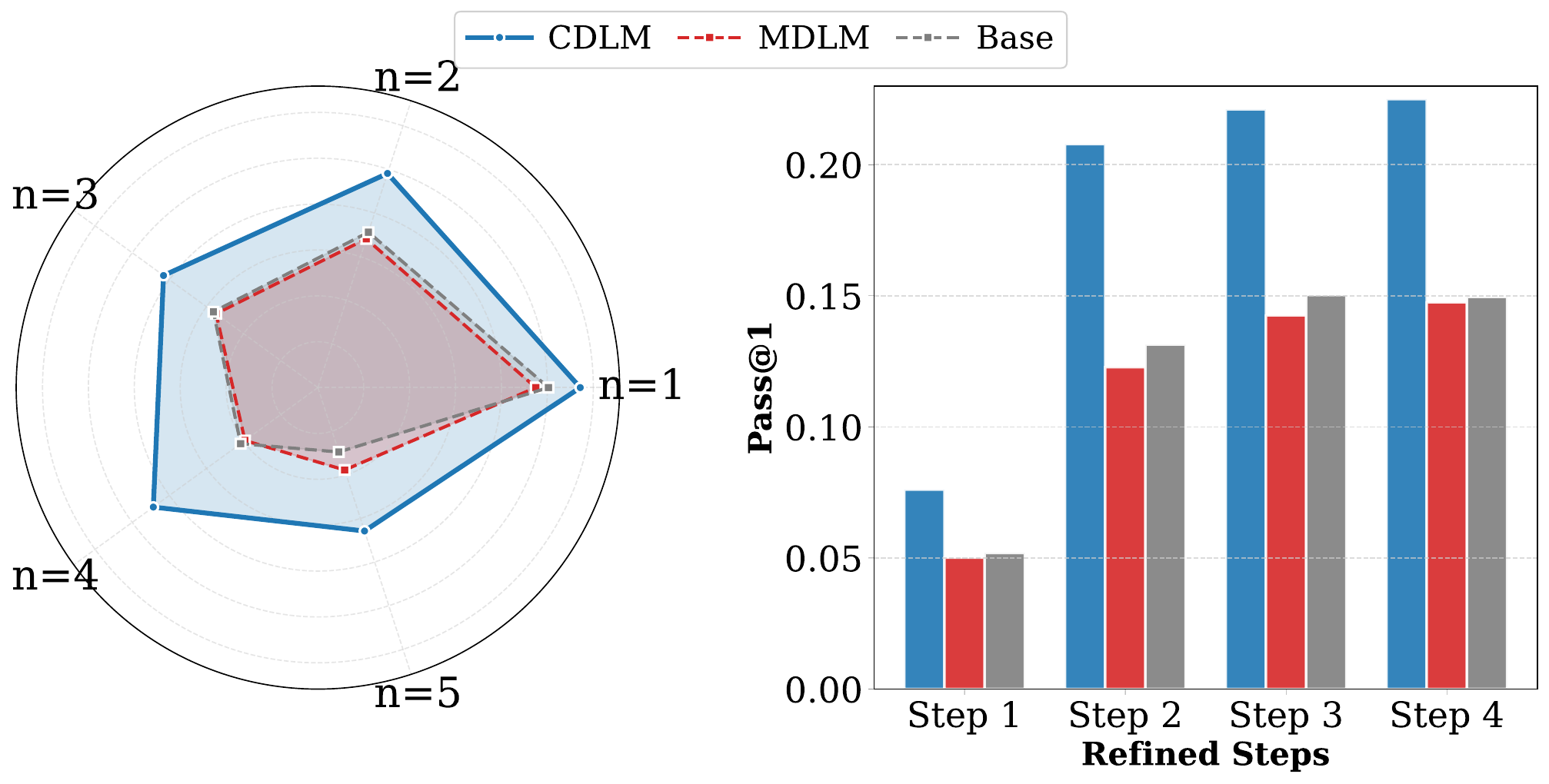}
    \caption{Pass@1 under confidence-based refinement, averaged over the four CRB datasets. Left: Pass@1 after four refinement steps ($T=4$, $\tau=0.9$) across corruption levels. Right: Mean Pass@1 over all error types and corruption levels vs.\ refinement depth $T$ (``Step $T$''). CDLM outperforms MDLM and the base model at every depth and corruption level.}
    \label{fig:pass_at_1_comparison}
\end{figure}

\section{Experiments}
\label{sec:experiments}

Unless otherwise noted, our default model is \textsc{CDLM-0.5B}, obtained by finetuning \textsc{Open-dCoder-0.5B}~\citep{opendllm2025} for $2000$ steps on Nemotron-SFT-Code, the code subset of Nemotron-Pretraining-SFT-v1~\citep{nvidia2025nvidianemotronnano2}. We compare it with the pretrained base model and with an absorbing-only fine-tuned model (MDLM) trained with identical settings except the objective; all hyperparameters are listed in Appendix~\ref{app:hparams}.

\subsection{Controlled Comparison at 0.5B}
\label{sec:main_comparison}

This is the controlled comparison referenced in Section~\ref{sec:phenomenon}: dataset, optimizer, learning-rate schedule, and number of training steps are identical between the absorbing-only MDLM and the mixture-trained CDLM-0.5B, isolating the training objective as the only variable.

\paragraph{Error Localization Improvements.}
On HumanEval CRB, the mixture objective produces larger clean-vs-erroneous confidence gaps than MDLM in 14 of the 15 error-type $\times$ $n_{\text{replace}}$ cells (Figure~\ref{fig:error_localization}a) and higher Top-$K$ hit rates for every $K\in\{1,\ldots,6\}$ (Figure~\ref{fig:error_localization}b); over the 12 CRB cells at $n_{\text{replace}}=1$ its gap and Hit@1 also exceed the base model's ($0.169$ and $0.228$ vs.\ $0.106$ and $0.147$, Table~\ref{tab:prism_h2h}).

\paragraph{Error Correction Improvements.}
Figure~\ref{fig:pass_at_1_comparison} reports deterministic Pass@1 under confidence-based refinement. Across $n_{\text{replace}}\!\in\!\{1,\ldots,5\}$, CDLM achieves higher Pass@1 than both MDLM and the base model at every refinement depth; all three arms gain most from the second step and change little afterwards, and CDLM's margin over MDLM grows from $+0.026$ at $T=1$ to $+0.078$--$+0.085$ at $T=2$--$4$ (mean over the 60 dataset $\times$ error-type $\times$ $n_{\text{replace}}$ cells). Aggregated CRB results across all error types and benchmarks are in Appendix~\ref{addditonal_pass_at_1}.
CDLM also remasks \emph{fewer} first-step tokens than base or MDLM at $\tau=0.9$ ($2.87\%$ vs.\ $3.09\%$ and $2.95\%$ of body tokens at $n_{\mathrm{replace}}=1$; Table~\ref{tab:prism_h2h}) while repairing more, so its gains come from editing the right positions rather than more of them.

\paragraph{Statistical reliability.}
Paired by task (exact McNemar test and paired bootstrap, pooled over datasets, error types and $n_{\mathrm{replace}}\in\{1,\ldots,5\}$, $N\approx 9{,}090$), CDLM exceeds MDLM by $+0.029$, $+0.070$, $+0.062$ and $+0.067$ Pass@1 at $T=1,\ldots,4$ (all $p<10^{-28}$); no cell of the full grid shows CDLM significantly below MDLM or base. With two additional training seeds per arm (three in total; $n_{\mathrm{replace}}=1$, 12-cell macro), Pass@1 is $0.189\pm0.006$ vs.\ $0.139\pm0.0003$ at $T=1$ and $0.286\pm0.002$ vs.\ $0.243\pm0.008$ at $T=4$ (mean $\pm$ SD over training seeds), with every CDLM seed above every MDLM seed (Appendix~\ref{app:stats}).

\subsection{Which Positions to Supervise}
\label{sec:ctw}

GIDD- and BERT-style training also apply loss to clean visible tokens, whereas CDLM supervises only masked and corrupted positions. To isolate this axis, we add a clean-token term with weight $\mathrm{ctw}=0.1$ (CDLM$+$ctw) and compare it with CDLM trained with the same 2000-step learning-rate schedule (Table~\ref{tab:ctw_ablation}). The two arms do not differ significantly at $T=1$ (paired $-0.012$, $p=0.08$), but clean-token supervision turns refinement from a small gain into a loss: CDLM$+$ctw declines from $0.204$ at $T=1$ to $0.182$ at $T=4$, while CDLM rises from $0.197$ to $0.205$, and CDLM is ahead at every $T\ge2$ (paired $+0.016$, $+0.021$, $+0.016$; $p=0.044$, $0.006$, $0.040$). Supervising clean tokens makes the model confident at more positions: in $53\%$ of CDLM$+$ctw programs no position falls below $\tau$ in the first step, against $44\%$ for CDLM ($n_{\mathrm{replace}}=1$, $3{,}213$ programs), so fewer errors enter the iterative loop. The difference is modest, comes from one training run per arm, and is concentrated on HumanEval+ and MBPP+ (Appendix~\ref{app:stats_ctw}). Our main checkpoint CDLM-0.5B (same step, much longer learning-rate schedule; Appendix~\ref{app:lr_horizon}) gains more from refinement ($0.188\to0.286$); comparing it with CDLM$+$ctw would mix schedule and supervision target.

\begin{table}[!htbp]
\centering
\small
\begin{minipage}[t]{0.55\linewidth}
\centering
\setlength{\tabcolsep}{4pt}
\caption{Effect of supervising clean tokens. CDLM (loss only on masked and corrupted positions) vs.\ CDLM$+$ctw (additional loss on clean visible tokens, $\mathrm{ctw}=0.1$, an approximation to GIDD/BERT-style supervision within our framework), both trained with the same 2000-step learning-rate schedule. Pass@1 macro-averaged over the 12 CRB dataset $\times$ error-type cells on the same instance files, $n_{\mathrm{replace}}=1$, $\tau=0.9$. $^{\dagger}$Main checkpoint (same 2000 steps, longer learning-rate schedule), for reference.}
\label{tab:ctw_ablation}
\begin{tabular}{lcccc}
\toprule
\textbf{Setting} & $T=1$ & $T=2$ & $T=3$ & $T=4$ \\
\midrule
CDLM            & 0.197 & \textbf{0.213} & \textbf{0.210} & \textbf{0.205} \\
CDLM$+$ctw      & \textbf{0.204} & 0.192 & 0.184 & 0.182 \\
\midrule
\textit{CDLM-0.5B}$^{\dagger}$ & \textit{0.188} & \textit{0.279} & \textit{0.281} & \textit{0.286} \\
\bottomrule
\end{tabular}
\end{minipage}\hfill
\begin{minipage}[t]{0.42\linewidth}
\centering
\setlength{\tabcolsep}{3pt}
\caption{Threshold sensitivity: Pass@1 ($n_{\mathrm{replace}}=1$, $T=1$, 12-cell macro) for $\tau\in\{0.80,0.90,0.95\}$. MDLM is evaluated on CDLM's instance files so that the two arms are paired (on its own files: $0.114/0.137/0.153$); other rows use their own files. CDLM exceeds MDLM at every $\tau$ (paired $p\le3\times10^{-17}$).}
\label{tab:tau_sensitivity}
\begin{tabular}{lccc}
\toprule
\textbf{Model} & $0.80$ & $0.90$ & $0.95$ \\
\midrule
Dream-7B-Base & \textbf{0.788} & 0.776 & 0.740 \\
LLaDA-8B-Base       & 0.423 & 0.552 & \textbf{0.636} \\
Open-dCoder-0.5B    & 0.125 & 0.143 & \textbf{0.163} \\
\midrule
MDLM (absorbing)    & 0.120 & 0.139 & \textbf{0.156} \\
CDLM (mixture)      & 0.178 & 0.188 & \textbf{0.205} \\
\bottomrule
\end{tabular}
\end{minipage}
\end{table}

\subsection{Transfer to LLaDA-8B}
\label{sec:llada8b}

\begin{table}[!htbp]
\centering
\small
\setlength{\tabcolsep}{4pt}
\caption{Controlled comparison on LLaDA-8B-Base (LoRA, 2000 steps, identical settings except the objective). Gap and Hit@1 on the 541 HumanEval localization records ($n_{\mathrm{replace}}=1$, pooled over error types; Appendix~\ref{app:llada8b_setup}); Pass@1 macro-averaged over the 12 dataset $\times$ error-type cells ($n_{\mathrm{replace}}=1$, $\tau=0.9$). Pass@1 truncates each completion at its first markdown fence, which both fine-tuned arms learned to emit from the corpus; no base completion contains a fence. Raw scores and fence counts are in Appendix~\ref{app:llada8b_fence}.}
\label{tab:llada8b}
\begin{tabular}{lcccc}
\toprule
\textbf{LLaDA-8B-Base} & \textbf{Gap} & \textbf{Hit@1} & \textbf{Pass@1 $T{=}1$} & \textbf{Pass@1 $T{=}4$} \\
\midrule
Base (no fine-tuning)        & 0.265 & 0.165 & 0.551 & 0.595 \\
+ absorbing-only (MDLM)      & 0.402 & 0.152 & 0.600 & 0.609 \\
+ mixture (CDLM)             & \textbf{0.777} & \textbf{0.608} & \textbf{0.725} & \textbf{0.623} \\
\bottomrule
\end{tabular}
\end{table}

We repeat the controlled comparison on LLaDA-8B-Base, the only one of the three backbones pretrained from scratch as an MDLM. Both arms train LoRA adapters ($r=64$) for 2000 steps on the same Nemotron-SFT-Code data, with every setting identical except the objective (Appendix~\ref{app:llada8b_setup}). Absorbing-only training raises Pass@1 ($0.551\to0.600$) and the HumanEval confidence gap ($0.265\to0.402$) but not HumanEval Hit@1 ($0.165\to0.152$). The mixture objective raises Hit@1 to $0.608$ and Pass@1 to $0.725$, $+0.124$ over the compute-matched arm (Table~\ref{tab:llada8b}). Over all four datasets MDLM's Hit@1 does rise ($0.173\to0.393$), but CDLM's rises to $0.691$ (Appendix~\ref{app:llada8b_conf}). The effect appears early: after 200 steps (1{,}200 documents, about $0.37$M tokens) the gap is already $0.790$ vs.\ $0.258$ and Hit@1 $0.530$ vs.\ $0.105$.
At corrupted positions CDLM's top prediction stays sharp (mean maximum probability $0.866$) while the probability of the token actually present collapses to $0.192$ ($0.983$ and $0.969$ at clean positions): error awareness is a collapse of $p_\theta(z_i\mid\bm{z})$, the quantity the remask rule thresholds, not a loss of predictive certainty, and under a maximum-over-vocabulary confidence CDLM would have the smallest gap of the three arms (Appendix~\ref{app:llada8b_conf}).
CDLM-8B reaches $0.725$ against Dream-7B-Base's $0.776$ (Table~\ref{tab:tau_sensitivity}): our claim concerns the objective under a controlled comparison, not the state of the art. At $T=4$ the margin over base narrows. Most of this drop comes from HumanEval+, where CDLM writes an opening fence into the blank first line of many bodies, which the fence-truncation rule then scores as an empty program (Appendix~\ref{app:llada8b_t4}).

\subsection{Robustness to the Corruption Distribution}
\label{sec:robustness}

CRB corruptions, like the training noise, are uniform within a lexical category, so the gains could reflect recognition of the training noise. We therefore rebuilt CRB with each corrupted token drawn from the base model's own top-5 predictions at that position, the distributional opposite of uniform noise (on HumanEval, with the function body as context, the substitute is about $975\times$ more probable under the base model than a random in-category token for operators and $494\times$ for identifiers, geometric means). CDLM still exceeds compute-matched MDLM in all 12 cells on the gap, Hit@1, Hit@6 and Pass@1 at $T=4$ under both corruption regimes, and at $T=1$ in 11 of 12 (random) and 10 of 12 (plausible) cells. It keeps $85\%$ of its Hit@6 margin and $72\%$ of its $T=4$ Pass@1 margin, but only about half ($48.5\%$) of its Hit@1 margin, mostly lost on operators, where every model including base degrades (Appendix~\ref{app:plausible_crb}). On the models' own failing generations, a median of 59 tokens from the canonical solution over the four datasets (109 on HumanEval; one for CRB), all arms stay at the floor, so that probe is uninformative about error awareness (Appendix~\ref{app:natural_errors}). MATH and ParallelBench (Section~\ref{sec:when_correction_helps}) contain no corrupted token at all.

\subsection{Ablations}
\label{sec:ablations}

\paragraph{Effect of Mixture Probability.}
We sweep $\alpha$ over 12 values from $0.04$ to $0.9$, one Open-dCoder-0.5B checkpoint per value with all other settings fixed, evaluated on HumanEval CRB at $T=1$ and $T=4$ (Figure~\ref{fig:mixture_prob_ablation} and Table~\ref{tab:alpha_sweep} in Appendix~\ref{app:alpha_sweep}). This is a post-hoc sensitivity analysis; $\alpha=0.1$ had been fixed beforehand (Section~\ref{sec:mixture}). The confidence gap grows almost monotonically with $\alpha$, but localization is best at $\alpha=0.7$ and degrades above it (Figure~\ref{fig:mixture_prob_ablation}a--c). Pass@1 shows the same trade-off: $\alpha\in[0.04,0.2]$, which contains $\alpha=0.1$, gives $0.091$--$0.119$ at $T=1$ and $0.234$--$0.258$ at $T=4$, while $[0.3,0.7]$ gives $0.151$--$0.207$ and $0.275$--$0.324$, so our reported gains are not a tuned maximum. With one seed per $\alpha$ we report an interval rather than a peak (the $T=1$ and $T=4$ maxima are at $0.7$ and $0.5$). Above $0.7$, Pass@1 falls most at $T=4$ ($0.165$ and $0.163$ for $\alpha=0.8, 0.9$): excessive corruption mainly removes the benefit of further refinement steps.

\paragraph{Sensitivity to the Confidence Threshold $\tau$.}
All refinement experiments in the main paper use a fixed remasking threshold $\tau=0.9$. To verify that CDLM's gains are not an artifact of this choice, we sweep $\tau\in\{0.80, 0.90, 0.95\}$ and re-evaluate Pass@1 on CRB ($n_{\mathrm{replace}}=1$, $T=1$). In Table~\ref{tab:tau_sensitivity} the ranking of the pretrained models is unchanged across these thresholds, and CDLM exceeds the compute-matched MDLM at every threshold (paired over programs, $+0.044$ to $+0.053$, $p\le3\times10^{-17}$); even the best $\tau$ for MDLM ($0.156$) stays below CDLM's worst ($0.178$), so the benefit comes from the training objective, not threshold tuning.

\paragraph{Comparison with Autoregressive Self-Editing.}
An autoregressive self-editor in the spirit of Self-Edit~\citep{zhang2023selfedit}, without test feedback (gpt-4.1-nano~\citep{openai2025gpt41}), reaches Pass@1 $0.790$ on CRB ($n_{\mathrm{replace}}=1$) when it emits only \texttt{BEFORE/AFTER} edits (median $30$ sequential forward passes) and $0.802$ when it rewrites the whole function (median $89$). Dream-7B-Base reaches $0.776$ and LLaDA-8B CDLM $0.725$ with two full-sequence passes ($T=1$), $15$--$45\times$ fewer sequential passes, though not fewer FLOPs, since a DLM pass recomputes the full sequence while an AR step uses a KV cache. With one AR model and two prompting strategies, this shows only that in-place repair reaches comparable accuracy with far fewer sequential passes (Appendix~\ref{app:ar_baseline}).

\begin{table}[!htbp]
\centering
\small
\setlength{\tabcolsep}{5pt}
\caption{From-scratch generation across code, math and parallel-decoding benchmarks. \textbf{Top:} HumanEval(+) Pass@1/Pass@10 at 0.5B under vanilla and ReMDM~\citep{wang2025remasking} decoding (MBPP(+) in Appendix~\ref{app:cdlm_codegen}). \textbf{Middle:} MATH~\citep{hendrycks2021math} suite (with the MATH-500 split of \citet{lightman2023letsverify}) for \textsc{LLaDA2.0-mini-preview} (16B) after one epoch of fine-tuning on Nemotron-Math-v2~\citep{du2025nemotronmath}; \emph{MDLM} corresponds to vanilla SFT ($\alpha=0$) and \emph{CDLM} to mixture SFT ($\alpha=0.1$), with all other training settings matched. \textbf{Bottom:} ParallelBench task-group averages, bold marking the best remask value (per-task results in Table~\ref{tab:parallelbench_final}, Appendix~\ref{app:cdlm_parallelbench}). On HumanEval(+) and MATH, CDLM improves over MDLM and base; on MBPP(+) both fine-tuned arms stay below base in Pass@1 (Appendix~\ref{app:cdlm_codegen}).}
\label{tab:pure_completion_results}

\begin{tabular}{l cccc cccc}
\toprule
\multicolumn{9}{c}{\textit{Code generation (HumanEval / HumanEval+, CDLM-0.5B)}} \\
\midrule
& \multicolumn{4}{c}{\textbf{Vanilla decoding}}
& \multicolumn{4}{c}{\textbf{ReMDM decoding}} \\
\cmidrule(lr){2-5}\cmidrule(lr){6-9}
& \multicolumn{2}{c}{HumanEval} & \multicolumn{2}{c}{HumanEval+}
& \multicolumn{2}{c}{HumanEval} & \multicolumn{2}{c}{HumanEval+} \\
\textbf{Method} & P@1 & P@10 & P@1 & P@10 & P@1 & P@10 & P@1 & P@10 \\
\midrule
Base & 0.19 & 0.37 & 0.17 & 0.32 & 0.18 & 0.33 & 0.16 & 0.29 \\
MDLM & 0.19 & 0.38 & 0.17 & 0.34 & 0.20 & 0.38 & 0.17 & 0.34 \\
CDLM & \textbf{0.22} & \textbf{0.41} & \textbf{0.21} & \textbf{0.39} & \textbf{0.21} & \textbf{0.43} & \textbf{0.20} & \textbf{0.38} \\
\bottomrule
\end{tabular}

\vspace{4pt}

\begin{tabular}{l cccc}
\toprule
\multicolumn{5}{c}{\textit{Math reasoning (LLaDA2.0-mini-preview 16B, one-epoch SFT)}} \\
\midrule
\textbf{Method} & MATH-500 & Algebra & Inter.\ Algebra & PreAlgebra \\
\midrule
Base                & 44.4 & 60.3 & 22.7 & 57.4 \\
MDLM (vanilla SFT)  & 44.6 & 60.6 & 18.5 & 57.1 \\
CDLM (mixture SFT)  & \textbf{47.8} & \textbf{63.1} & \textbf{23.8} & \textbf{62.6} \\
\bottomrule
\end{tabular}

\vspace{4pt}

\begin{tabular}{l ccc}
\toprule
\multicolumn{4}{c}{\textit{ParallelBench (LLaDA2.0-mini 16B), Fixed / Remask decoding}} \\
\midrule
\textbf{Method} & Low-entropy avg.\ (6) & High-entropy avg.\ (4) & Overall avg.\ (10) \\
\midrule
Base & 0.45 / 0.47 & 0.44 / 0.56 & 0.45 / 0.50 \\
MDLM & 0.93 / 0.93 & 0.47 / 0.45 & 0.75 / 0.74 \\
CDLM & 0.93 / \textbf{0.94} & 0.46 / \textbf{0.62} & 0.74 / \textbf{0.81} \\
\bottomrule
\end{tabular}
\end{table}

\section{When Correction-Aware Training Helps Generation}
\label{sec:when_correction_helps}
We test whether mixture training also improves \emph{from-scratch} generation in three domains, none of which contains a corrupted input. Results are summarized in Table~\ref{tab:pure_completion_results}.
\textbf{From-scratch code.} Generating the function body from the signature and docstring (HumanEval) or the whole function from the task description (MBPP) under remask-based decoding~\citep{peng2025path}, CDLM scores above MDLM and base on HumanEval(+)~\citep{chen2021evaluatinglargelanguagemodels, liu2023humanevalplus} under vanilla and ReMDM~\citep{wang2025remasking} decoding; a paired bootstrap separates it from both on HumanEval+ but not from base on HumanEval (Appendix~\ref{app:scratch_eval}). On MBPP(+) both fine-tuned arms fall below base (Appendix~\ref{app:cdlm_codegen}); a measured distribution shift of the fine-tuned models is at most a contributing factor (Appendix~\ref{app:mbpp_shift}).
\textbf{High-entropy parallel decoding.} On ParallelBench~\citep{kang2025parallelbench} with \textsc{LLaDA2.0-mini} (16B)~\citep{bie2025llada20scalingdiffusionlanguage}, CDLM lifts the high-entropy average under remasking from $0.45$ (compute-matched MDLM) to $0.62$, above the base model's $0.56$, whereas MDLM cannot exploit remasking at all ($0.47$ fixed vs.\ $0.45$ remask); low-entropy tasks are matched. With about 100 instances per task, per-task differences of a few points are not individually significant.
\textbf{Math at 16B.} We fine-tune \textsc{LLaDA2.0-mini-preview} (16B) on Nemotron-Math-v2~\citep{du2025nemotronmath} for one epoch. MDLM is vanilla SFT ($\alpha=0$) and CDLM is mixture SFT ($\alpha=0.1$) of the same checkpoint on the same corpus, with every other setting identical. Mixture SFT scores above base and vanilla SFT on MATH-500~\citep{lightman2023letsverify} and the three MATH~\citep{hendrycks2021math} subsets we evaluate; with standard errors of 1.4--2.2 points in these single runs, only the Prealgebra gains and the Intermediate Algebra gain over vanilla SFT clearly exceed the noise. MATH involves no in-place editing and no token replacement, so this gain cannot come from recognizing one corruption operator. The 16B results are single matched runs; the ablation suite was run only at 0.5B.

\section{Conclusion and Limitations}
Masked-diffusion training gives an MDLM no direct supervision on whether a visible token is wrong. Supervising visible corrupted tokens supplies it: the probability of the token actually present collapses at corrupted positions, which is exactly what remask-based refinement thresholds. Supervising clean tokens as well makes the model confident at more positions and, at 0.5B, stops iterative refinement from improving with depth, so the supervised positions should be chosen with the decoder in mind. On CRB the effect of supervising corrupted tokens holds under matched compute at 0.5B, transfers to LLaDA-8B, persists under model-plausible corruptions, and carries over to parallel decoding and 16B math.
\textbf{Limitations.} Remask-then-repredict decoding works on a fixed-length sequence, so insertions, deletions and variable-length edits are out of reach, which is why CRB is length-preserving; and in-place refinement does not repair a model's own failing generations, which differ from the canonical solution by a median of 59 tokens (109 on HumanEval). The ablations were run only at 0.5B, the 8B and 16B results are single training runs, and two of the three probed backbones, including our 0.5B base, are adapted from autoregressive checkpoints. Threshold-based refinement often runs to its full step budget, and at 8B neither fine-tuned arm improves from-scratch generation over base (Appendix~\ref{app:limitations}). \textbf{Impact.} Better error awareness may make DLMs more reliable, but they can still produce incorrect outputs and need appropriate safeguards.

\begin{ack}
This work did not receive any research funding. Experiments were run on a shared GPU cluster at the University of Wisconsin--Madison, and other costs, including API usage for the autoregressive baseline, were paid personally by the first author. The authors declare no competing interests.
\end{ack}

\bibliographystyle{plainnat}
\bibliography{ref}

\clearpage

\appendix

\section*{Contents of the Appendix}

The appendix includes the following contents:
\begin{itemize}
    \item Appendix~\ref{app:limitations} discusses the limitations of the benchmark, the training principle and the experiments.
    \item Appendix~\ref{sec:app_A} discusses related work, including temporal signals in DLM decoding.
    \item Appendix~\ref{addditonal_pass_at_1} reports aggregated Pass@1 results on CRB.
    \item Appendix~\ref{sec:self_revision_minimal} presents a self-revision analysis under minimal corruption.
    \item Appendix~\ref{sec:CDLM_training_details} gives implementation details: hyperparameters of all runs, the training objective as implemented, the learning-rate horizon, the from-scratch generation protocol, the MBPP distribution shift, and the use of LLMs.
    \item Appendix~\ref{app:stats} reports paired significance tests and training-seed variance.
    \item Appendix~\ref{app:robustness} evaluates model-plausible corruptions and naturally occurring errors.
    \item Appendix~\ref{app:llada8b} details the LLaDA-8B transfer experiment.
    \item Appendix~\ref{app:alpha_sweep} gives the full mixture-probability sweep.
    \item Appendix~\ref{app:oci} replicates the 0.5B and 8B comparisons on the public OpenCodeInstruct corpus.
    \item Appendix~\ref{app:sudoku_exp} provides a controlled from-scratch analysis of corrective behavior on Sudoku.
    \item Appendix~\ref{app:ar_baseline} compares with autoregressive self-editing.
    \item Appendix~\ref{app:concurrent} compares with concurrent corrective methods (a self-correction head and temporal features).
    \item Appendix~\ref{app:algorithms} presents the remask-based iterative refinement procedures.
    \item Appendix~\ref{app:release} describes the released code, models and data and what they reproduce.
\end{itemize}

\section{Limitations}
\label{app:limitations}
Our goal is to isolate localized corrective behavior under masked diffusion and remask-based refinement, which aligns with the formulation commonly adopted by current foundation-model-style DLMs~\citep{nie2025largelanguagediffusionmodels, ye2025dream7bdiffusionlarge, opendllm2025}. This scope brings the following limitations.

\paragraph{Fixed-length editing.} Remask-then-repredict decoding operates on a fixed-length token sequence: a position can be masked and rewritten, but no position can be created or deleted. Insertions, deletions, and more general variable-length edits are therefore outside what in-place diffusion refinement can currently express, and CRB is length-preserving for this reason rather than by benchmark design. Supporting such edits requires a variable-length formulation, as explored in recent discrete diffusion work~\citep{havasi2025edit, chao2025maskedunmaskeddiscretediffusion, zhang-etal-2025-flexible, reid2023diffuser}, together with executable revision benchmarks that cover them.

\paragraph{Localized errors only.} In-place refinement addresses localized errors. A CRB instance differs from a passing program by a single token, whereas a model's own failing generation differs from the canonical solution by a median of 109 tokens on HumanEval. On such programs all arms remain at the floor (Appendix~\ref{app:natural_errors}); end-to-end generation failure at this scale is a different regime.

\paragraph{Scale of the ablations and number of runs.} The correction-aware objective is evaluated at 0.5B, 8B and 16B, but the controlled ablations (mixture probability, clean-token supervision, threshold sensitivity) were run only on Open-dCoder-0.5B. The 8B and 16B comparisons are single training runs per arm, and the paired tests of Appendix~\ref{app:stats} bound finite-problem-sample uncertainty for one deterministic decoding run per arm. The clean-token comparison with matched schedules rests on one training run per arm, shows a modest effect ($+0.016$ to $+0.021$ Pass@1 at $T\ge2$), and its two arms were trained on different machines.

\paragraph{Backbone provenance.} Dream-7B-Base and Open-dCoder-0.5B, our main fine-tuning base, are adapted from autoregressive Qwen2.5 checkpoints. LLaDA-8B-Base (Section~\ref{sec:llada8b}) and the Sudoku DiT trained from random initialization (Appendix~\ref{app:sudoku_exp}) carry the evidence that does not depend on autoregressive pretraining.

\paragraph{Sudoku as a mechanism study.} In the Sudoku refinement experiments the editable set always contains every noisy position, which is an oracle constraint. These experiments therefore validate the confidence-separation mechanism and its use for refinement, not end-to-end localization.

\paragraph{Threshold-based refinement.} With a fixed threshold, refinement often runs to its full step budget instead of converging: on LLaDA-8B at $T=4$, 42--89\% of programs still remask a position in the last step, depending on arm and dataset. The extra steps keep rewriting low-confidence positions, and on HumanEval+ the fine-tuned CDLM fills the blank first line of many bodies with a markdown fence, which interacts with how outputs are scored and narrows its margin over base at $T=4$ (Appendix~\ref{app:llada8b_t4}). An explicit convergence or stopping criterion, or an editable set restricted to positions that carry code, is a natural extension.

\paragraph{Generation.} Fine-tuning on the Nemotron code corpus shifts the 0.5B models away from MBPP-style programs, and both fine-tuned arms stay below base on MBPP(+) (Appendix~\ref{app:mbpp_shift}). At 8B, neither the absorbing-only nor the mixture LoRA arm improves from-scratch HumanEval+ or MBPP+ generation over base, and the two arms are not significantly different (Appendix~\ref{app:llada8b_guard}); we make no claim of improved from-scratch generation at 8B.

\section{Related Work}
\label{sec:app_A}

\subsection{Development of DLMs}
The development of diffusion-based language modeling begins with continuous-space approaches such as DiffusionLM~\citep{li2022diffusionlm}, which apply Gaussian noise in continuous space over token embeddings but remain misaligned with the discrete nature of textual data. The introduction of D3PM~\citep{austin2021structured} establishes a unified framework for discrete-state diffusion through flexible transition kernels, encompassing uniform, absorbing, and structured corruption processes. Uniform-noise discrete diffusion~\citep{schiff2025simple} is easier to guide because every token can be revised, but it still trails absorbing-state diffusion in language-modeling quality~\citep{ni2025trainingoptimallargediffusion}. DiffusionBERT~\citep{he-etal-2023-diffusionbert} trains a pretrained BERT as the denoiser of an absorbing-state (\texttt{[MASK]}) diffusion and improves perplexity and BLEU over D3PM and Diffusion-LM.
Building on these developments, the masked diffusion language model (MDLM) formulation~\citep{sahoo2024simple} interprets masked diffusion as a variational objective composed of weighted masked language modeling (MLM)-style reconstruction terms and achieves strong performance among discrete DLMs. At scale, LLaDA~\citep{nie2025largelanguagediffusionmodels} performs comparably to autoregressive baselines trained under the same settings, and \citet{ni2025trainingoptimallargediffusion} derive compute- and data-constrained scaling laws for DLMs.

\subsection{Self-correction in DLMs.}
Recent work has shown that DLMs may accumulate errors during parallel decoding.
\citet{kang2025parallelbench} and \citet{chen2025optimal} demonstrate that parallel denoising steps can introduce or amplify token-level inconsistencies, highlighting the need for mechanisms that explicitly identify and correct decoding errors.
A number of subsequent approaches address this issue by incorporating remasking strategies or auxiliary guidance during inference.
\citet{peng2025path} propose Path Planning (P2), a remask-based decoding procedure in which a planner, either an external model such as BERT or the denoiser itself, selects tokens for resampling, while \citet{wang2025remasking} introduce remasking discrete diffusion models that guide inference-time remasking using token-level confidence to improve generation quality without additional training.
\citet{lee2025effective} use a reward model to guide noising and denoising transitions that refine intermediate states in place.
\citet{kim2025finetuningmaskeddiffusionprovable} further integrate remasking decisions into the model itself by introducing an internal prediction head that estimates, for each token, the probability that it should be remasked, so that remasking decisions at inference time need no external planner or scoring model. The head is trained on data constructed by comparing MDM-generated tokens against ground truth. Appendix~\ref{app:prism} compares the two approaches on CRB with the same base model, data and number of training steps. CDLM takes a different and complementary path, modifying the training objective itself without architectural changes or extra data construction; the two approaches address orthogonal aspects of correction and could in principle be combined.
\begin{table}[!htbp]
\centering
\small
\setlength{\tabcolsep}{4pt}
\caption{Methodological distinctions between GIDD~\citep{rutte2025generalized} and CDLM. While both combine absorbing and uniform noise, the two methods differ in framework, training paradigm, supervision targets, and decoding semantics.}
\label{tab:gidd_vs_cdlm}
\begin{tabular}{l l l}
\toprule
\textbf{Dimension} & \textbf{GIDD} & \textbf{CDLM} \\
\midrule
Framework & Generalized interpolating diffusion & Standard MDLM \\
Training & From scratch (as published) & Post-training of a pretrained MDLM \\
Loss & Unified ELBO (KL + Itakura--Saito) & MDLM loss + separable noise term ($\lambda_{\mathrm{noise}}$) \\
Noise schedule & Bell-shaped, varies with $t$ & Fixed mixture probability $\alpha$ on visible tokens \\
Loss on clean tokens & Yes (all positions supervised) & No (only masked + corrupted) \\
Decoding & All tokens may change at every step & Remask-then-repredict, localized in-place \\
\bottomrule
\end{tabular}
\end{table}

Most relevant to our work, \citet{rutte2025generalized} propose the Generalized Interpolating Discrete Diffusion framework, which combines absorbing and uniform noise to generalize discrete diffusion and improve generation quality. Despite the shared high-level idea of mixing absorbing and uniform noise, GIDD differs from CDLM along the six methodological dimensions enumerated in Table~\ref{tab:gidd_vs_cdlm}: framework, training paradigm, loss formulation, noise schedule, supervision targets, and decoding semantics. GIDD is trained from scratch in \citet{rutte2025generalized} and applies loss to all positions including clean visible tokens, whereas CDLM is applied as post-training of a pretrained MDLM and supervises only masked $+$ corrupted positions. The empirical consequence of supervising clean tokens is studied in Section~\ref{sec:experiments} (Table~\ref{tab:ctw_ablation}), where it stops iterative refinement from improving with depth.

\input{appendix/rw_temporal}

\subsection{Efficient Decoding in DLMs}
The inference speed of diffusion language models (DLMs) is fundamentally limited by the trade-off between parallel decoding efficiency and output quality. A growing body of work has investigated parallel decoding strategies to mitigate this issue.

One line of research adapts speculative decoding from autoregressive models to the diffusion setting, enabling simultaneous decoding of multiple tokens. Self Speculative Decoding introduces a speculative decoding framework tailored for DLMs \citep{gao2025selfspeculativedecodingdiffusion}. Reformulating the classic draft--verify paradigm, \citet{wu2025freedraftandverificationlosslessparallel} propose a self-verifiable decoding mechanism that enables lossless parallel decoding. Related approaches incorporate auxiliary verification signals: \citet{israel2025acceleratingdiffusionllmsadaptive} employ a smaller verification model to dynamically adjust the number of decoded tokens based on error estimates.

Beyond explicit verification, other methods improve efficiency through token-level commitment strategies. \citet{liang2026cd4lmconsistencydistillationadaptive} propose Confidence-Adaptive Decoding, which accelerates inference by committing high-confidence tokens early, while \citet{wang2025creditdecodingacceleratingparalleldecoding} introduce CreditDecoding, leveraging historical logits to estimate token convergence and guide more efficient parallelization.

Another line of work addresses the lack of key--value (KV) caching in standard DLMs, which stems from their use of bidirectional attention, preventing reuse of cached keys and values across refinement steps and thereby limiting inference efficiency compared to autoregressive models.
Block Diffusion introduces block-wise generation to enable KV cache reuse across decoding steps \citep{arriola2025block}. Fast-dLLM proposes a training-free block-wise parallel decoding strategy with KV caching \citep{wu2025fastdllmtrainingfreeaccelerationdiffusion}, and Fast-dLLM v2 further extends this approach with hierarchical caching mechanisms \citep{wu2025fastdllmv2efficientblockdiffusion}. Causal parallel decoding via Jacobi Forcing designs a rejection--recycling and multi-block decoding algorithm to improve inference efficiency \citep{hu2025fastaccuratecausalparallel}. TiDAR integrates speculative decoding with KV cache reuse by employing a diffusion model as a parallel drafter \citep{liu2025tidarthinkdiffusiontalk}, while WeDLM proposes a prefix-cache-compatible streaming parallel decoding strategy using causal masking and a dynamic sliding window to alleviate the bottleneck of block decoding \citep{liu2025wedlmreconcilingdiffusionlanguage}.

\subsection{Self-Correction and Refinement in Large Language Models}
A number of prior studies on large language models have explored concepts related to correction and refinement across general datasets and multiple domains, often under the formulation of self-correction, critique, or iterative improvement.
\citet{huang2024selfcorrect} study models’ ability to self-correct on reasoning-oriented benchmarks, focusing on improving final answers through iterative reasoning.
RealCritic~\citep{tang2024realcritic} proposes a critique-and-correction framework that explicitly models how the quality of critiques affects downstream solution refinement, while RefineBench~\citep{lee2025refinebenchevaluating} allows models to autonomously decide whether refinement is necessary and to generate self-feedback before revision.
In task-specific settings, particularly for code, several benchmarks have been proposed to evaluate code editing and repair capabilities under more realistic scenarios.
BugsInPy~\citep{widyasari2020bugsinpy} and pydra~\citep{kitanidis2025pydra} construct benchmarks from real or synthetic bugs, and \citet{opu2025tangled} improve method-level bug datasets by detecting tangled code changes, all to evaluate code correction capabilities.
More recent benchmarks such as CodeEditorBench~\citep{guo2025codeeditorbench}, SWE-bench~\citep{jimenez2024swebenchlanguagemodelsresolve}, and EditBench~\citep{chi2025editbenchevaluatingllmabilities} further study code editing in realistic development settings, often requiring models to perform multi-line, instruction-driven, or repository-level modifications.

While these benchmarks demonstrate that language models can revise and improve their outputs under various correction paradigms, they predominantly treat correction as a task-level or procedural process that generates revised solutions from scratch or through external critique loops.
In particular, refinement in these settings is typically not performed \emph{in place} on a complete input, but instead relies on regeneration, multi-stage prompting, or auxiliary feedback mechanisms. A representative AR self-editing approach is Self-Edit~\citep{zhang2023selfedit}, which trains an autoregressive editor to revise generated code using \emph{external} feedback signals such as test case results and runtime errors. CDLM addresses a fundamentally different setting: corrective behavior is \emph{intrinsic} to the model, relying solely on its own learned token-level confidence with no external oracle, and editing is performed in place by a few bidirectional passes rather than through sequential regeneration. Appendix~\ref{app:ar_baseline} compares DLM repair with a prompted autoregressive editor (gpt-4.1-nano, no test feedback), which the strongest DLMs match in accuracy with $15$--$45\times$ fewer sequential forward passes.
In contrast, our Code Revision Benchmark (CRB) isolates \emph{in-place refinement} by applying localized corruptions to a complete input and evaluating whether a model can identify and correct erroneous tokens while preserving correct context.

\section{Aggregated Pass@1 Results on CRB}
Table~\ref{tab:mixture_training_pass_at_1_comparison} reports aggregated Pass@1 for refinement-based code correction on standard benchmarks (HumanEval/HumanEval+/MBPP/MBPP+), using 4 refinement steps with a confidence threshold of 0.9, comparing MDLM, CDLM, and the base model.
\label{addditonal_pass_at_1}
\begin{table}[!htbp]
\centering
\small
\caption{
Pass@1 on refinement-based code correction across standard coding benchmarks for MDLM, CDLM, and the base model.
Results are aggregated over all error types and corruption levels using 4 refinement steps with a confidence threshold of 0.9.
CDLM consistently outperforms MDLM and the base model across datasets.
}
\begin{tabular}{lccc}
\toprule
\textbf{Dataset} & \textbf{MDLM} & \textbf{CDLM} & \textbf{Base} \\
\midrule
HumanEval      & 0.157 & 0.257 & 0.168 \\
HumanEval+     & 0.182 & 0.242 & 0.196 \\
MBPP           & 0.114 & 0.175 & 0.114 \\
MBPP+          & 0.137 & 0.226 & 0.120 \\
\midrule
\textbf{Average} & \textbf{0.148} & \textbf{0.225} & \textbf{0.150} \\
\bottomrule
\end{tabular}
\label{tab:mixture_training_pass_at_1_comparison}
\end{table}

\section{Self-revision under minimal corruption}
\label{sec:self_revision_minimal}
\begin{table}[!htbp]
\centering
\small
\caption{
Self-revision under minimal corruption on HumanEval.
For each model, we first generate programs and retain only those that pass all tests.
A single type-preserving token corruption ($n_{\text{replace}} = 1$) is then applied to each correct program, and the model is evaluated on its ability to revise the corrupted program back to a passing solution using iterative refinement.
Reported values correspond to Pass@1 after different numbers of refinement steps.
}
\begin{tabular}{lcccc}
\toprule
Model & T=1 & T=2 & T=3 & T=4 \\
\midrule
LLaDA-8B-Base        & 0.550 & 0.571 & 0.577 & 0.580 \\
Dream-7B-Base       & 0.528 & 0.548 & 0.402 & 0.427 \\
Open-dCoder-0.5B    & 0.203 & 0.342 & 0.344 & 0.336 \\
\bottomrule
\end{tabular}
\label{tab:self_revision_minimal}
\end{table}
The Code Revision Benchmark (CRB) constructs revision instances by introducing controlled corruptions to canonical programs.
While this design isolates error localization and correction behavior, there is a risk that a model may fail to revise a corrupted program simply because the input falls outside the distribution of code it typically generates.
In such cases, poor revision performance could reflect distributional mismatch rather than a genuine inability to detect and correct errors.

To disentangle these factors, we design a \emph{self-revision under minimal corruption} setting, in which each model is evaluated exclusively on programs that it has already demonstrated the ability to generate correctly.
By introducing only a minimal, localized perturbation to model-generated solutions, this setting makes unfamiliar inputs an unlikely explanation for revision failures and probes the model’s capacity for targeted error detection and correction.

\paragraph{Setup.}
For each model, we first identify a set of programs that the model can generate correctly under standard decoding.
Starting from these correct model-generated programs, we introduce a single type-preserving token corruption ($n_{\text{replace}} = 1$) using the CRB corruption pipeline.
Restricting to a single-token modification keeps the corrupted program close to the model’s own generation, differing from a correct solution only through a minimal and localized perturbation.
This design isolates the model’s ability to recognize and revise a small error, rather than its capacity to handle large semantic changes or distributional shifts.

We then apply iterative refinement to the corrupted program and evaluate whether the model can revise it back to a correct solution.
Performance is measured using Pass@1 after varying numbers of refinement steps, ranging from 1 to 4.

\paragraph{Results and Discussion.}
Table~\ref{tab:self_revision_minimal} evaluates self-revision performance under a single-token corruption setting.
Even when starting from programs that each model has itself generated and verified as correct, no evaluated DLM repairs more than 58\% of the corrupted programs.
Additional refinement steps help Open-dCoder-0.5B ($0.203\to0.336$) and LLaDA-8B-Base ($0.550\to0.580$) but lower Dream-7B-Base ($0.528\to0.427$), so more iterations do not reliably improve correction.
These failures persist despite the minimal and localized nature of the corruption and although the uncorrupted programs were generated by each model itself.
Although the models are capable of generating correct programs, they often fail to identify and repair even a single incorrect token once it is introduced.

\section{Implementation Details of CDLM Training}
\label{sec:CDLM_training_details}

This appendix gives the fine-tuning setups of the code-generation and ParallelBench experiments, the hyperparameters of all training runs, the objectives as implemented, the learning-rate horizon of the 0.5B runs, the from-scratch generation protocol, the MBPP distribution shift, and the use of LLMs.

\subsection{Code Generation Fine-Tuning}
\label{app:cdlm_codegen}

For the code generation experiments, we fine-tune the \textsc{Open-dCoder-0.5B} checkpoint~\citep{opendllm2025} for $2000$ optimization steps on Nemotron-SFT-Code, the code subset of Nemotron-Pretraining-SFT-v1~\citep{nvidia2025nvidianemotronnano2}.
The Nemotron corpus consists of high-quality code samples spanning multiple programming languages and problem domains.

Throughout the code generation section, we compare three models:
(i) the pretrained base model without additional fine-tuning,
(ii) a model fine-tuned using the absorbing-only objective (\textsc{MDLM}), and
(iii) the mixture-trained \textsc{CDLM-0.5B}.
The training configurations for (ii) and (iii) are identical except for the choice of training objective.
In particular, both models share the same optimizer, learning rate schedule, batch size, noise schedule, and remasking configuration, ensuring that observed differences are attributable solely to the training objective.

\paragraph{MBPP Results.}
In the main paper, we focus on HumanEval and HumanEval+ for compact presentation, and report MBPP/MBPP+ in this appendix.
Table~\ref{tab:mbpp_results_appendix} summarizes Pass@1 and Pass@10 on MBPP and MBPP+, along with an average over the two benchmarks.
Both fine-tuned arms are below the pretrained base model on these benchmarks in Pass@1 (the $\Delta$ rows of Table~\ref{tab:mbpp_results_appendix}). \textsc{CDLM} scores above \textsc{MDLM} in this table; with one training run per arm, we do not draw conclusions from these MBPP(+) differences between the two objectives.
Fine-tuning on Nemotron measurably moves both models away from all four benchmarks, and CDLM more from MBPP than from HumanEval (Appendix~\ref{app:mbpp_shift}); we regard the shift as a contributing factor rather than a full explanation of the gap to the base model.

\begin{table}[!htbp]
\centering
\small
\setlength{\tabcolsep}{5pt}
\caption{
Pass@1 and Pass@10 on MBPP and MBPP+ under two decoding variants.
\emph{Vanilla} recomputes token confidence at every refinement step, while \emph{ReMDM} fixes each token's confidence at its first unmasking step~\citep{wang2025remasking}.
The \textbf{Avg} column averages MBPP and MBPP+.
Boldface marks the higher of \textsc{CDLM} and \textsc{MDLM} under the same decoding setting (one training run per arm); $\Delta$ rows give the difference to the base model.
}
\begin{tabular}{l cccc cc}
\toprule
\textbf{Method} &
\multicolumn{2}{c}{\textbf{MBPP}} &
\multicolumn{2}{c}{\textbf{MBPP+}} &
\multicolumn{2}{c}{\textbf{Avg}} \\
& Pass@1 & Pass@10 & Pass@1 & Pass@10 & Pass@1 & Pass@10 \\
\midrule
MDLM (Vanilla)
& 0.106 & 0.312
& 0.177 & 0.449
& 0.142 & 0.381 \\
CDLM (Vanilla)
& \textbf{0.131} & \textbf{0.352}
& \textbf{0.207} & \textbf{0.472}
& \textbf{0.169} & \textbf{0.412} \\
Base (Vanilla)
& 0.157 & 0.350
& 0.231 & 0.502
& 0.194 & 0.426 \\
\quad $\Delta$ MDLM $-$ Base & $-0.051$ & $-0.038$ & $-0.054$ & $-0.053$ & $-0.053$ & $-0.046$ \\
\quad $\Delta$ CDLM $-$ Base & $-0.026$ & $+0.002$ & $-0.024$ & $-0.030$ & $-0.025$ & $-0.014$ \\
\midrule
\addlinespace[2pt]
MDLM (ReMDM)
& 0.108 & 0.328
& 0.170 & 0.444
& 0.139 & 0.386 \\
CDLM (ReMDM)
& \textbf{0.141} & \textbf{0.350}
& \textbf{0.211} & \textbf{0.476}
& \textbf{0.176} & \textbf{0.413} \\
Base (ReMDM)
& 0.160 & 0.366
& 0.232 & 0.485
& 0.196 & 0.426 \\
\quad $\Delta$ MDLM $-$ Base & $-0.052$ & $-0.038$ & $-0.062$ & $-0.041$ & $-0.057$ & $-0.040$ \\
\quad $\Delta$ CDLM $-$ Base & $-0.019$ & $-0.016$ & $-0.021$ & $-0.009$ & $-0.020$ & $-0.013$ \\
\bottomrule
\end{tabular}
\label{tab:mbpp_results_appendix}
\end{table}

\subsection{ParallelBench Fine-Tuning Details}
\label{app:cdlm_parallelbench}

For the ParallelBench experiments, we fine-tune \textsc{LLaDA2.0-mini} (16B)~\citep{bie2025llada20scalingdiffusionlanguage} on data generated from the ParallelBench benchmark~\citep{kang2025parallelbench}.
We construct a training set of $50{,}000$ examples and fine-tune the model for $6$ epochs using block diffusion~\citep{arriola2025block}.

During training, the mask ratio is sampled uniformly from $[0,1]$ for each example.
We adopt block diffusion with a block size of $32$, consistent with the decoding configuration used at evaluation time.
Since ParallelBench tasks involve short outputs, the maximum generation length is capped at $64$ tokens.

At inference time, we use confidence-based decoding with two modes, fixed and remask.
In the fixed mode, tokens are generated without further remasking.
In the remask mode, previously unmasked tokens with low confidence are remasked and regenerated in subsequent steps.
We set the confidence threshold to $0.6$ to increase parallelism during decoding.

\begin{table}[!htbp]
\centering
\small
\setlength{\tabcolsep}{6pt}
\renewcommand{\arraystretch}{1.0}
\caption{Per-task accuracy on ParallelBench~\citep{kang2025parallelbench} for \textsc{LLaDA2.0-mini} (16B)~\citep{bie2025llada20scalingdiffusionlanguage} under confidence-based decoding. Each entry reports \emph{Fixed / Remask}; bold marks the best value within each row. Tasks are grouped by output entropy: low-entropy tasks admit one valid output (e.g., Copy), high-entropy admit many (e.g., Shuffle). Implementation details are in Appendix~\ref{app:cdlm_parallelbench}.}
\label{tab:parallelbench_final}
\begin{tabular}{l c c c}
\toprule
\textbf{Task} & \textbf{Base} & \textbf{MDLM} & \textbf{CDLM} \\
& \multicolumn{3}{c}{\textit{Fixed / Remask}} \\
\midrule
\multicolumn{4}{l}{\textbf{Low-Entropy (6 tasks)}} \\
Copy            & 0.97 / 0.97 & \textbf{1.00} / \textbf{1.00} & \textbf{1.00} / \textbf{1.00} \\
Reverse         & 0.60 / 0.63 & \textbf{1.00} / \textbf{1.00} & \textbf{1.00} / \textbf{1.00} \\
Sort            & 0.15 / 0.19 & 0.61 / 0.61 & 0.60 / \textbf{0.64} \\
Insert Index    & 0.23 / 0.24 & \textbf{1.00} / \textbf{1.00} & 0.99 / 0.99 \\
Remove Index    & 0.36 / 0.37 & \textbf{1.00} / \textbf{1.00} & \textbf{1.00} / \textbf{1.00} \\
Replace Index   & 0.41 / 0.41 & \textbf{1.00} / \textbf{1.00} & \textbf{1.00} / \textbf{1.00} \\
\midrule
\multicolumn{4}{l}{\textbf{High-Entropy (4 tasks)}} \\
Shuffle         & 0.37 / 0.52 & 0.47 / 0.38 & 0.45 / \textbf{0.64} \\
Insert Random   & 0.23 / 0.35 & 0.40 / 0.44 & 0.40 / \textbf{0.72} \\
Remove Random   & 0.65 / 0.70 & 0.75 / 0.72 & 0.69 / \textbf{0.77} \\
Replace Random  & 0.51 / \textbf{0.67} & 0.25 / 0.26 & 0.30 / 0.37 \\
\midrule
Low-Entropy avg.   & 0.45 / 0.47 & 0.93 / 0.93 & 0.93 / \textbf{0.94} \\
High-Entropy avg.  & 0.44 / 0.56 & 0.47 / 0.45 & 0.46 / \textbf{0.62} \\
\textbf{Overall avg.} & 0.45 / 0.50 & 0.75 / 0.74 & 0.74 / \textbf{0.81} \\
\bottomrule
\end{tabular}
\end{table}

\paragraph{ParallelBench Data Format and Tasks.}

ParallelBench consists of structured transformation tasks designed to evaluate parallel decoding under high output uncertainty.
Each example includes a natural language instruction and a short input sequence, and the model must generate an output that satisfies the specified transformation.
Many tasks admit multiple valid outputs, making exact prediction unlikely and iterative refinement essential.

Below we illustrate three representative tasks used in our experiments with concrete examples.

\textbf{Shuffle} requires the model to randomly permute the input sequence while ensuring that the output differs from the original order.
\emph{Instruction:} Randomly shuffle the waiting line.
\emph{Input:} \texttt{["Paul Payne", "Robert Riley", "Peter Stone"]}.
\emph{Output:} Any permutation different from the input (e.g., \texttt{["Robert Riley", "Peter Stone", "Paul Payne"]}).

\textbf{Replace Index} requires replacing the element at a specified position with a given value.
\emph{Instruction:} Replace the person at position $0$ with \texttt{"Henry Warren"}.
\emph{Input:} \texttt{["Patrick Morgan", "Eric King", "Joe Reed"]}.
\emph{Output:} \texttt{["Henry Warren", "Eric King", "Joe Reed"]}.

\textbf{Replace Random} requires replacing one randomly chosen element in the sequence with a given value.
\emph{Instruction:} Replace one random person with \texttt{"Juan Torres"}.
\emph{Input:} \texttt{["David Owens", "Kelly Payne", "Aaron Freeman"]}.
\emph{Output:} Any list where exactly one name is replaced by \texttt{"Juan Torres"}.

Some of these tasks naturally induce high output entropy, particularly in the \emph{Shuffle} and \emph{Replace Random} settings.
As a result, errors arise primarily from parallel decoding rather than externally injected noise, making ParallelBench well suited for evaluating correction-aware training under high-uncertainty generation.

\input{appendix/implementation}

\input{appendix/statistics}
\input{appendix/robustness}

\input{appendix/llada8b}

\input{appendix/alpha_sweep}

\input{appendix/oci}

\section{Sudoku: A Controlled From-Scratch Proxy for Pretraining}
\label{app:sudoku_exp}

\begin{wrapfigure}{r}{0.5\linewidth}
    \centering
    \includegraphics[width=\linewidth]{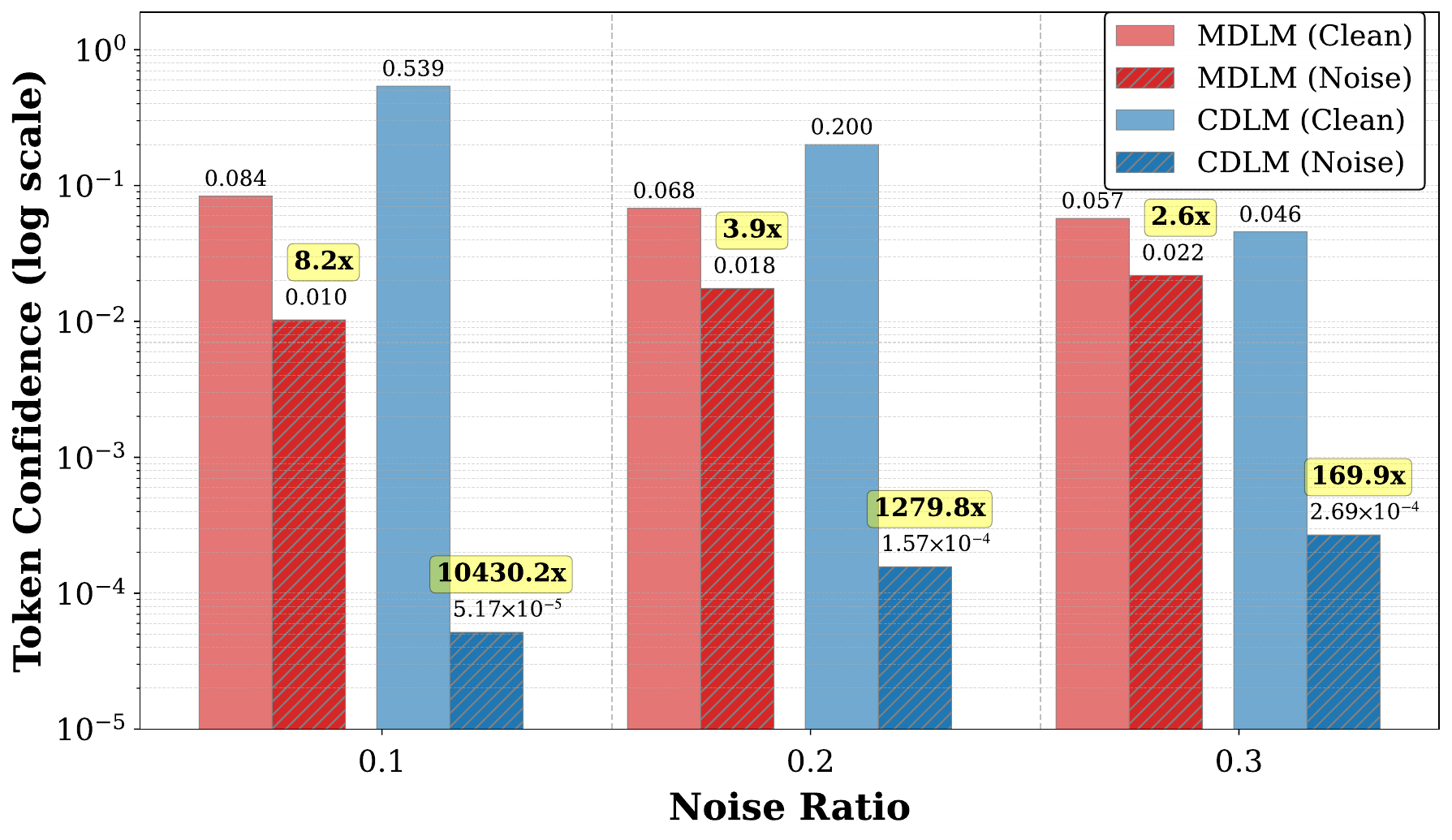}
    \caption{
    Confidence on clean and noisy digits under uniform corruption (log scale).
    We report clean--noise confidence ratios (yellow annotations) rather than absolute confidence gaps, since confidence values span multiple orders of magnitude in this setting.
    CDLM exhibits substantially larger clean--noise separation than MDLM across noise ratios, indicating much stronger error localization than the absorbing-only objective.}
    \label{fig:sudoku-uniform-noise}
\end{wrapfigure}

To further validate the proposed training principle in a fully controlled setting,
we conduct additional experiments on Sudoku and report the results.
Sudoku is a compact symbolic domain in which the full data distribution is observable and correctness is deterministically verifiable.
This environment allows us to isolate learning dynamics induced by the noising process and training objective, without interference from prior knowledge or data ambiguity.
It therefore provides a clean testbed for examining whether the proposed training objective induces error-aware confidence and iterative corrective behavior.

Our Sudoku CDLM uses a Diffusion Transformer (DiT)~\citep{peebles2023scalable} architecture with approximately 37M parameters
(12 layers, hidden size 512, 8 attention heads, MLP ratio 4, dropout 0.1),
following the experimental setup of \citet{kim2025finetuningmaskeddiffusionprovable}.
Each puzzle is represented as an 81-token sequence over a vocabulary of size 10 (a MASK token and digits 1--9).
Training uses 36{,}000 puzzles, batch size 64, 200{,}000 optimization steps,
learning rate $1\times 10^{-4}$, weight decay 0.01,
and a dynamically sampled masking ratio in $[0.2, 0.9]$.
For each objective we keep the checkpoint with the highest token accuracy on the 2{,}000 evaluation boards at mask ratio $0.5$, scored every 1{,}000 steps under that objective's own training noise (masked cells for MDLM; masked and replaced cells for CDLM). No separate validation split was held out, so absolute accuracies may be slightly optimistic.

This controlled from-scratch environment allows us to isolate and evaluate two core aspects of corrective behavior:
(i) whether training with the mixture objective induces error-awareness, namely the ability to assign lower confidence to corrupted tokens than to clean ones;
and (ii) whether improved error localization translates into more effective iterative, in-place correction.
In addition, we assess whether the mixture training objective improves standard absorbing-noise reconstruction,
corresponding to pure denoising-based generation without uniform corruptions.

\subsection{Uniform Noise Corruption: Error-Token Localization}
We begin by evaluating whether the mixture objective induces error-aware confidence calibration under uniform noise corruption.
Using a single forward pass without iterative refinement, a fraction of positions is corrupted by replacing the original digit with a uniformly sampled incorrect digit.
We then measure the model’s confidence on clean versus corrupted positions.
This setting isolates the effect of uniform-noise supervision on token-level confidence, independent of any iterative sampling dynamics.

Figure~\ref{fig:sudoku-uniform-noise} reports confidence statistics across noise ratios of $0.1$, $0.2$, and $0.3$.
Models trained with the absorbing-only objective exhibit only weak separation between clean and corrupted digits.
For example, at a noise ratio of $0.1$, the ratio between mean confidence on clean digits and noisy digits is approximately $8.2\times$.
In contrast, models trained with the mixture objective exhibit substantially stronger separation: the clean--noise confidence ratio exceeds $10{,}000\times$ at noise ratio $0.1$ and remains above $160\times$ even at noise ratio $0.3$.

These results show that mixture objective training induces strong error-awareness in the model’s confidence estimates.
Corrupted digits are consistently assigned low confidence while clean digits retain high confidence, providing a reliable signal for error localization.
This calibrated confidence forms a critical prerequisite for effective remask-based iterative correction, which we evaluate next.

\subsection{Iterative Correction under Uniform Noise}
\begin{minipage}{0.58\linewidth}
    \centering
    \includegraphics[width=\linewidth]{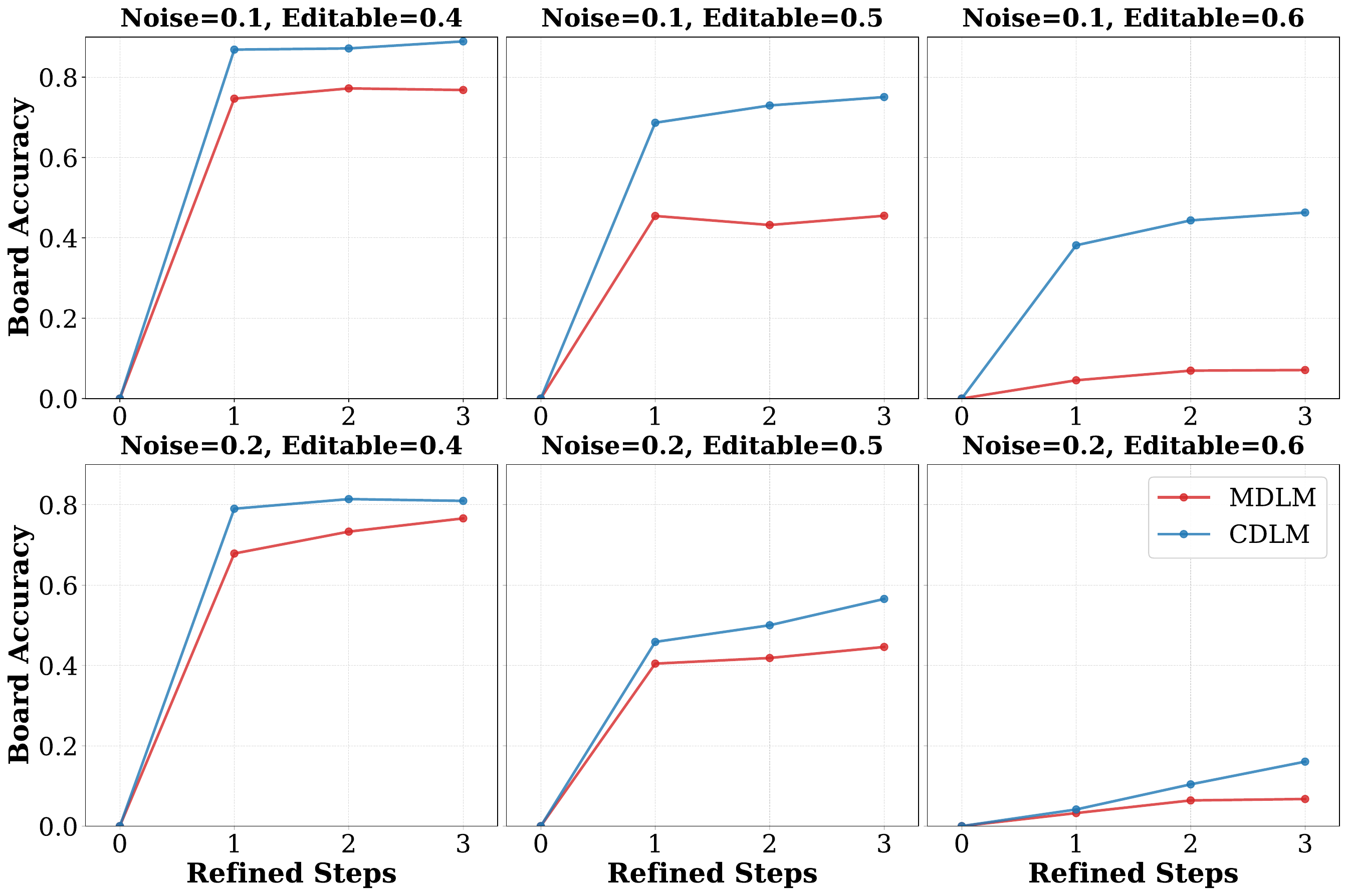}
    \captionof{figure}{
    Sudoku board accuracy under uniform corruption with iterative diffusion sampling.
    Each subplot shows accuracy as a function of refinement steps for a specific combination of noise ratio and editable ratio.
    Across all difficulty settings, CDLM consistently achieves higher accuracy than MDLM.
    }
    \label{fig:sudoku-uniform-diffusion}
\end{minipage}
\begin{minipage}{0.40\linewidth}
    \centering
    \includegraphics[width=\linewidth]{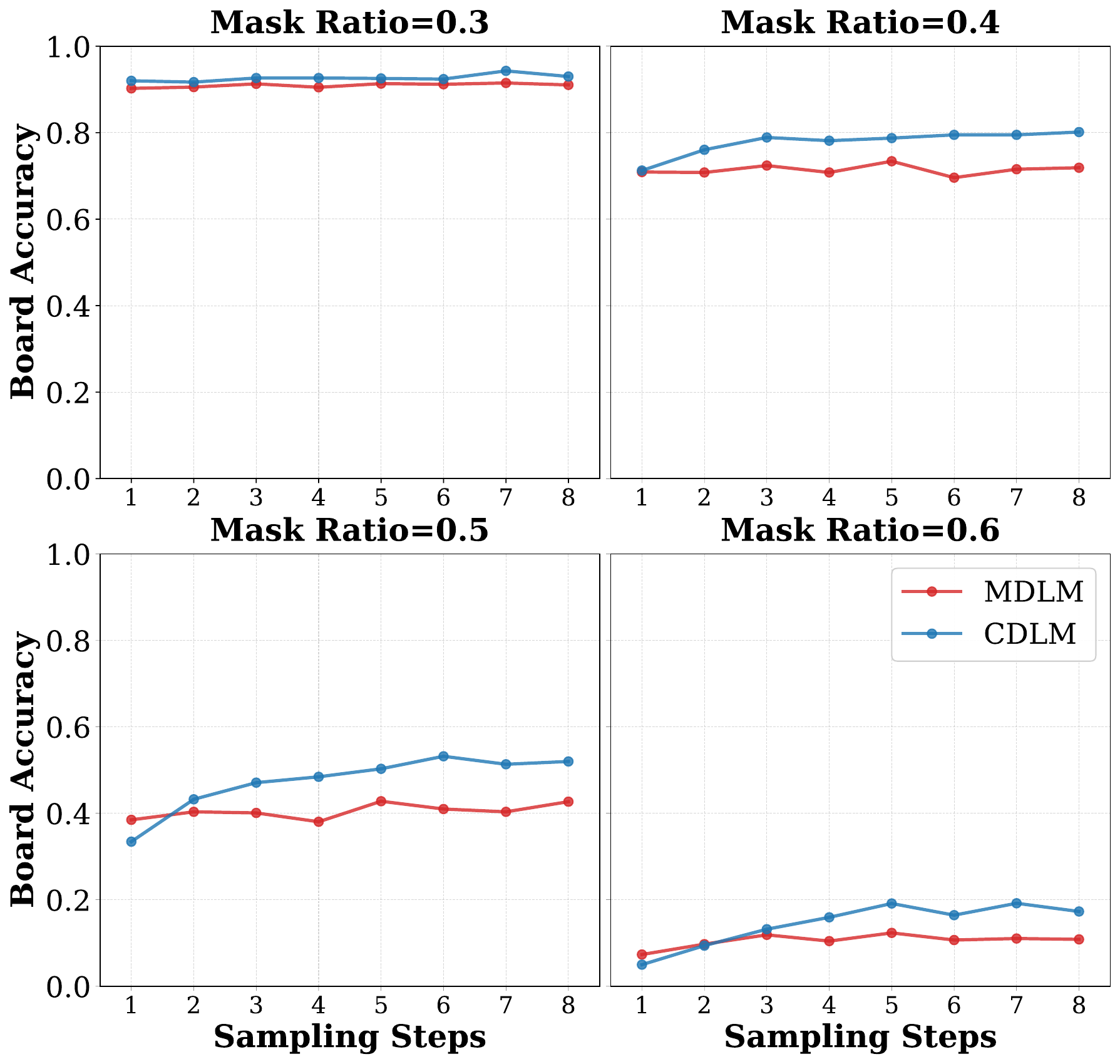}
    \captionof{figure}{
    Pure completion performance under absorbing-mask corruption.
    Each subplot shows board accuracy as a function of sampling steps for a fixed mask ratio.
    With two or more sampling steps CDLM matches or exceeds MDLM at every mask ratio; with a single step it is slightly lower at mask ratios $0.5$ and $0.6$.
    }
    \label{fig:sudoku-absorbing}
\end{minipage}
We next study whether the improved error localization learned by \textsc{CDLM}
translates into more effective iterative correction.
Starting from grids corrupted by uniform noise, we define an editable region that
always includes all noisy positions and may additionally include clean positions
according to a specified editable ratio.
Refinement is performed by iteratively remasking and resampling
tokens within the editable region, while treating all other cells as fixed
conditioning context.
The full remask-based refinement procedure, including confidence computation and
update rules, is detailed in
Appendix~\ref{app:sudoku_refinement} (Algorithm~\ref{alg:sudoku_editable_remask}).
We vary the noise ratio, the editable ratio, and the number of refinement steps,
and evaluate board accuracy after the final refinement step.

Figure~\ref{fig:sudoku-uniform-diffusion} summarizes results for two noise ratios ($0.1$ and $0.2$), three editable ratios ($0.4$, $0.5$, $0.6$), and up to three refinement steps.
Across all settings, \textsc{CDLM} consistently achieves higher board accuracy than \textsc{MDLM}.
This advantage is already visible after a single refinement step and often increases with additional steps.
These results suggest that the improved confidence calibration learned by \textsc{CDLM} facilitates not only better error localization, but also more effective guided, in-place refinement under uniform corruption.

\subsection{Pure Completion under Masked Denoising}

We finally evaluate the effect of the mixture objective in a pure completion
setting.
Unlike the refinement experiments, this setup does not involve uniformly
corrupted digits.
Instead, a fraction of positions is replaced with a mask token, and the model
performs multi-step denoising to produce a complete Sudoku grid using
confidence-guided remask-based decoding.
This setting corresponds to standard masked denoising and serves to assess
whether training for corrective behavior also benefits pure denoising-based
generation.

Figure~\ref{fig:sudoku-absorbing} reports board accuracy across mask ratios from $0.3$ to $0.6$ using confidence-guided decoding with a threshold of~$0.7$.
With two or more sampling steps, models trained with the mixture objective match or outperform those trained with the absorbing-only objective at every mask ratio; with a single step they are slightly lower at mask ratios $0.5$ and $0.6$.
The improvement is particularly pronounced at higher mask ratios, where confidence-guided refinement enables the model to identify and revise low-confidence digits during the denoising process.
These results demonstrate that incorporating uniform-noise supervision not only preserves pure completion performance, but can also improve robustness in challenging masked reconstruction regimes.

\section{Efficiency Comparison with Autoregressive Self-Editing}
\label{app:ar_baseline}

This appendix provides the detailed comparison summarized in Section~\ref{sec:ablations}. We compare DLM-based correction against an autoregressive self-editor on CRB ($n_{\mathrm{replace}}=1$, all four datasets $\times$ three error types, the $3{,}213$ test-failing programs of the Qwen2.5-tokenized instance files; a further 365 corrupted programs that already pass their tests are excluded, as in every DLM row). Self-Edit~\citep{zhang2023selfedit} is a representative AR self-editing approach that relies on \emph{external} feedback (test cases and runtime errors) to guide correction. CDLM addresses a fundamentally different setting: corrective behavior is \emph{intrinsic} to the model, using only learned token-level confidence with no external oracle. To establish a strong AR reference even without external feedback, we evaluate \texttt{gpt-4.1-nano}~\citep{openai2025gpt41} under two prompting strategies designed to favor AR:
\begin{itemize}
\item \textbf{Plan A}: AR outputs the complete corrected function (regenerative editing).
\item \textbf{Plan B}: AR outputs only the bug fix in a \texttt{BEFORE/AFTER} format, minimizing output tokens (localized editing).
\end{itemize}

We report Pass@1 and the Number of Function Evaluations (NFE), defined as the number of sequential model forward passes per sample. For AR, this equals the output token count. For DLMs, a run with $T$ refinement steps uses at most $T{+}1$ full-sequence passes (a sample's output is final after one pass if no position has confidence at or below $\tau$). At $T=1$ every sample that the DLM repairs must have been modified, and therefore used two passes; since more than half of the samples are repaired in each DLM row of Table~\ref{tab:ar_baseline_nfe} (pooled over cells, $53.1\%$ for LLaDA-8B-Base and more for the other rows), the median NFE is $2$. Median NFE is reported because the AR distribution is right-skewed. NFE counts sequential passes, not FLOPs: an AR step processes one new token with a KV cache, whereas a DLM pass recomputes the full sequence.

\begin{table}[h]
\centering
\small
\caption{Pass@1 and NFE on CRB ($n_{\mathrm{replace}}=1$). The AR model achieves slightly higher Pass@1 but uses $15$--$45\times$ more sequential forward passes than the DLMs, which repair in place with two full-sequence passes ($T=1$). All rows are macro-averaged over the 12 cells. $^{\dagger}$LoRA-trained CDLM arm of Section~\ref{sec:llada8b}, with each program truncated at its first markdown fence (Appendix~\ref{app:llada8b_fence}).}
\label{tab:ar_baseline_nfe}
\begin{tabular}{l l c c}
\toprule
\textbf{Model} & \textbf{Type} & \textbf{Pass@1} & \textbf{NFE (median)} \\
\midrule
gpt-4.1-nano (Plan A, full output)  & AR        & \textbf{0.802} & 89 \\
gpt-4.1-nano (Plan B, localize)     & AR        & 0.790          & 30 \\
\midrule
Dream-7B-Base                       & DLM ($T=1$) & 0.776          & \textbf{2} \\
LLaDA-8B-Base                       & DLM ($T=1$) & 0.552          & \textbf{2} \\
LLaDA-8B CDLM$^{\dagger}$            & DLM ($T=1$) & 0.725          & \textbf{2} \\
\bottomrule
\end{tabular}
\end{table}

The AR model achieves slightly higher Pass@1 but requires $15$--$45\times$ more sequential forward passes than the DLMs. This comparison uses a single AR model under two prompting strategies designed to favor it, so it supports only a narrow conclusion: bidirectional in-place repair reaches comparable accuracy with far fewer sequential forward passes, whereas AR editors must regenerate at least the edited span token by token. Improving DLM corrective behavior narrows the accuracy gap while keeping this property; on LLaDA-8B the mixture objective raises Pass@1 from $0.551$ to $0.725$ (Table~\ref{tab:llada8b}) at the same number of passes.

\input{appendix/concurrent}

\section{Remask-Based Iterative Refinement Algorithm}
\label{app:algorithms}
\subsection{Remask-Based Iterative Refinement Procedure in DLMs}
\label{app:algo_iter}

Algorithm~\ref{alg:confidence_refinement} summarizes the confidence-based iterative refinement procedure used throughout our experiments.
At each step, one forward pass fills the masked positions, every position is scored by the probability of the token that occupies it, and low-confidence positions are remasked; all other tokens are kept as conditioning context. A visible token is never overwritten directly: it is first remasked and then re-predicted in the next pass.
Unlike GIDD~\citep{rutte2025generalized}, which performs holistic diffusion updates over all token positions and does not employ an explicit remasking mechanism,
this procedure explicitly restricts updates to masked or remasked positions, enabling localized, in-place refinement.
In CRB the editable region $\mathcal{E}$ is the function body; the signature and docstring are fixed context.

\begin{algorithm}[t]
\caption{Confidence-Based Iterative Refinement with Remasking. $T$ refinement steps use at most $T{+}1$ forward passes; the last pass only fills masks, so the output contains no mask token.}
\label{alg:confidence_refinement}
\KwIn{
Initial sequence $\bm{z}^{(0)} \in (\mathcal{V} \cup \{\langle\mathrm{mask}\rangle\})^n$; editable positions $\mathcal{E}\subseteq\{1,\dots,n\}$; \\
Denoising model $p_\theta(\cdot \mid \cdot)$; \\
Confidence threshold $\tau \in (0,1)$; \\
Number of refinement steps $T$
}
\KwOut{Refined sequence without mask tokens}

\For{$t = 0,1,\dots,T$}{
    \tcp{One forward pass: fill masked positions, keep visible tokens}
    Compute token distributions $p_\theta(\cdot \mid \bm{z}^{(t)})$ for all positions \;
    $\tilde z_i^{(t)} \leftarrow
    \begin{cases}
    \arg\max_{v} p_\theta(v \mid \bm{z}^{(t)}), & z_i^{(t)} = \langle\mathrm{mask}\rangle \\
    z_i^{(t)}, & \text{otherwise}
    \end{cases}$ \quad for all $i$ \;
    \lIf{$t = T$}{\Return{$\tilde{\bm{z}}^{(t)}$}}

    \tcp{Confidence of the token occupying each position}
    $c_i^{(t)} \leftarrow p_\theta(\tilde z_i^{(t)} \mid \bm{z}^{(t)})$ for all $i\in\mathcal{E}$ \;
    $r^{(t)} \leftarrow \{\, i \in \mathcal{E} \mid c_i^{(t)} \le \tau \,\}$ \;
    \lIf{$r^{(t)} = \varnothing$}{\Return{$\tilde{\bm{z}}^{(t)}$}}

    \tcp{Remask; all other positions carry over}
    \ForEach{$i = 1,\dots,n$}{
        $z_i^{(t+1)} \leftarrow 
        \begin{cases}
        \langle\mathrm{mask}\rangle, & i \in r^{(t)} \\
        \tilde z_i^{(t)}, & \text{otherwise}
        \end{cases}$ \;
    }
}

\end{algorithm}

\subsection{Remask-Based Iterative Refinement for Sudoku under Uniform Noise}
\label{app:sudoku_refinement}

We consider Sudoku boards represented as sequences of length $81$, where each
visible token is a digit in $\{1,\dots,9\}$ and the model's vocabulary additionally
contains $\langle \mathrm{mask} \rangle$.
Given an initial grid corrupted by uniform replacement noise, we define an
\emph{editable set} of token positions
$\mathcal{E} \subseteq \{1,\dots,81\}$.
By construction, $\mathcal{E}$ always contains all uniformly corrupted (noisy)
positions, and may additionally include a subset of clean positions according to
a specified editable ratio.

\begin{algorithm}[t]
\caption{Remask-Based Iterative Refinement with Editable Tokens (Sudoku). Identical to Algorithm~\ref{alg:confidence_refinement} with the editable set $\mathcal{E}$ given by the evaluation protocol.}
\label{alg:sudoku_editable_remask}

\KwIn{
Initial sequence $\bm{z}^{(0)} = (z^{(0)}_1,\dots,z^{(0)}_{81})$,
$z^{(0)}_i \in \{1,\dots,9\}$,
corrupted by uniform replacement noise;\\
Denoising model $p_\theta(\cdot \mid \cdot)$ over $\{\langle \mathrm{mask} \rangle,1,\dots,9\}$;\\
Editable token set $\mathcal{E} \subseteq \{1,\dots,81\}$ containing all noisy positions;\\
Confidence threshold $\tau$;\\
Number of refinement steps $T$.
}

\KwOut{
Refined sequence without mask tokens.
}

\For{$t = 0,1,\dots,T$}{
  \tcp{One forward pass: fill masked positions, keep visible tokens}
  $\tilde z^{(t)}_i \leftarrow \arg\max_{v} p_\theta(v \mid \bm{z}^{(t)})$ if $z^{(t)}_i = \langle \mathrm{mask} \rangle$, else $z^{(t)}_i$ \quad for $i = 1,\dots,81$;\\
  \lIf{$t = T$}{\Return{$\tilde{\bm{z}}^{(t)}$}}
  \tcp{Confidence of the token occupying each editable position}
  $c^{(t)}_i \leftarrow p_\theta(\tilde z^{(t)}_i \mid \bm{z}^{(t)})$ for $i\in\mathcal{E}$;\\
  \tcp{Remasking set (restricted to editable tokens)}
  $\mathcal{R}^{(t)} \leftarrow
  \{\, i \in \mathcal{E} \;:\; c^{(t)}_i \le \tau \,\}$;\\
  \tcp{Next iterate; all non-remasked tokens carry over, so non-editable tokens never change}
  \For{$i = 1,2,\dots,81$}{
    \eIf{$i \in \mathcal{R}^{(t)}$}{
      $z^{(t+1)}_i \leftarrow \langle \mathrm{mask} \rangle$\;
    }{
      $z^{(t+1)}_i \leftarrow \tilde z^{(t)}_i$\;
    }
  }
}
\end{algorithm}

\paragraph{Notes.}
The editable set $\mathcal{E}$ is fixed throughout decoding and is constructed
\emph{a priori} for controlled evaluation.
The remasking set $\mathcal{R}^{(t)}$ is recomputed at every step based on
token-level confidence, but is always restricted to $\mathcal{E}$.
As a result, the procedure performs localized, in-place correction by allowing
updates only within the editable region, while preserving the remainder of the
grid as immutable conditioning context.

\input{appendix/release}

\newpage
\input{checklist.tex}

\end{document}

%% file: Tables_Code/Benchmarking/benchmark_comparison.tex
\begin{table*}[!htbp]
\centering
\small
\setlength{\tabcolsep}{3pt}
\renewcommand{\arraystretch}{0.9}
\caption{
Evaluation of confidence quality (top block) and iterative correction performance (bottom block) under varying numbers of corrupted tokens $n_{\text{replace}}$.
The top block reports the confidence gap between clean and erroneous tokens (\textit{gap}), together with top-1 and top-5 hit rates measuring how often error tokens appear among the lowest-confidence positions.
The bottom block reports Pass@1 after applying confidence-threshold-based iterative refinement for $T \in \{1,2,4\}$ refinement steps.
Each entry is the unweighted mean over the 12 CRB dataset $\times$ error-type cells (the gap pools body tokens within a cell), with $\tau=0.9$ and greedy decoding. LLaDA-8B-Base values differ slightly from Table~\ref{tab:llada8b}, which uses a different localization protocol (Appendix~\ref{app:llada8b_conf}).
}

\label{tab:confidence_refinement_benchmark}
\resizebox{\linewidth}{!}{%
\begin{tabular}{l *{15}{c}}
\toprule
& \multicolumn{3}{c}{$n_{\text{replace}} = 1$}
& \multicolumn{3}{c}{$n_{\text{replace}} = 2$}
& \multicolumn{3}{c}{$n_{\text{replace}} = 3$}
& \multicolumn{3}{c}{$n_{\text{replace}} = 4$}
& \multicolumn{3}{c}{$n_{\text{replace}} = 5$}
\\
\cmidrule(lr){2-4}
\cmidrule(lr){5-7}
\cmidrule(lr){8-10}
\cmidrule(lr){11-13}
\cmidrule(lr){14-16}
Model
& gap & top-1 & top-5
& gap & top-1 & top-5
& gap & top-1 & top-5
& gap & top-1 & top-5
& gap & top-1 & top-5 \\
\midrule
Dream-7B-Base         & 0.578 & 0.094 & 0.590 & 0.536 & 0.084 & 0.614 & 0.528 & 0.084 & 0.597 & 0.528 & 0.088 & 0.600 & 0.529 & 0.076 & 0.571 \\
LLaDA-8B-Base    & 0.235 & 0.178 & 0.770 & 0.209 & 0.198 & 0.801 & 0.199 & 0.172 & 0.811 & 0.185 & 0.159 & 0.784 & 0.191 & 0.162 & 0.833 \\
Open-dCoder-0.5B & 0.106 & 0.147 & 0.509 & 0.100 & 0.222 & 0.655 & 0.085 & 0.243 & 0.701 & 0.089 & 0.325 & 0.759 & 0.093 & 0.348 & 0.795 \\
\midrule
\\[-0.8em]
\midrule
& \multicolumn{3}{c}{$n_{\text{replace}} = 1$}
& \multicolumn{3}{c}{$n_{\text{replace}} = 2$}
& \multicolumn{3}{c}{$n_{\text{replace}} = 3$}
& \multicolumn{3}{c}{$n_{\text{replace}} = 4$}
& \multicolumn{3}{c}{$n_{\text{replace}} = 5$}
\\
\cmidrule(lr){2-4}
\cmidrule(lr){5-7}
\cmidrule(lr){8-10}
\cmidrule(lr){11-13}
\cmidrule(lr){14-16}
Model
& T=1 & T=2 & T=4
& T=1 & T=2 & T=4
& T=1 & T=2 & T=4
& T=1 & T=2 & T=4
& T=1 & T=2 & T=4
\\
\midrule
Dream-7B-Base
& 0.776 & 0.745 & 0.732
& 0.704 & 0.695 & 0.686
& 0.663 & 0.670 & 0.673
& 0.654 & 0.682 & 0.683
& 0.630 & 0.653 & 0.644
\\
LLaDA-8B-Base
& 0.552 & 0.587 & 0.596
& 0.419 & 0.466 & 0.483
& 0.338 & 0.405 & 0.421
& 0.319 & 0.374 & 0.407
& 0.280 & 0.327 & 0.359
\\
Open-dCoder-0.5B
& 0.143 & 0.230 & 0.251
& 0.059 & 0.162 & 0.178
& 0.036 & 0.116 & 0.141
& 0.017 & 0.089 & 0.104
& 0.005 & 0.061 & 0.074
\\
\bottomrule
\end{tabular}%
}

\end{table*}

%% file: appendix/rw_temporal.tex
\subsection{Temporal Signals in DLM Decoding}
\label{app:rw_temporal}
A separate line of work reads information from the denoising trajectory, that is, from how a DLM's predictions evolve across decoding steps, rather than from a single forward pass.
\citet{wang2025time} report \emph{temporal oscillation}: correct answers often appear at intermediate denoising steps and are then overwritten in later ones. They propose Temporal Self-Consistency Voting, a training-free decoding strategy that aggregates the predictions of all denoising steps, and Temporal Consistency Reinforcement, a post-training method that uses the semantic entropy of the intermediate predictions as a reward.
\citet{li2025diffusion} report \emph{early answer convergence}: in many cases the correct answer can already be identified halfway through decoding, under both semi-autoregressive and random remasking schedules. Their training-free decoder, Prophet, uses the confidence gap between the top two candidates to decide when to stop refining and decode all remaining tokens in one step, which reduces the number of decoding steps by up to $3.4\times$ on LLaDA-8B and Dream-7B.

Both methods read a trajectory that generation supplies: decoding starts from masked positions and fills them in step by step. In-place revision of a complete program supplies no such trajectory. The input contains no masked positions, and the positions to edit must be chosen from a single forward pass before any edit is made. Appendix~\ref{app:temporal} therefore induces a trajectory by unrolling the refinement loop and compares trajectory features with single-pass confidence on CRB.

The observation of \citet{li2025diffusion} supports our diagnosis. If predictions at masked positions stabilize early, the failure documented in Section~\ref{sec:phenomenon} is specifically a failure at visible positions, which are exactly the positions the absorbing objective never supervises (Section~\ref{sec:mixture}).

%% file: appendix/implementation.tex
\providecommand{\hpc}[1]{\parbox[t]{\linewidth}{\raggedright\strut#1\strut}}

\subsection{Hyperparameters of All Training Runs}
\label{app:hparams}

Table~\ref{tab:hparams} lists the training and decoding settings of every model in the main text; the self-correction head (Appendix~\ref{app:prism}), the 10{,}000-step CDLM checkpoint (Appendix~\ref{app:temporal}) and the OpenCodeInstruct models (Appendix~\ref{app:oci}) are described where they are used. The first column covers CDLM-0.5B and MDLM-0.5B, which are used in the main comparison (Section~\ref{sec:main_comparison}, Figures~\ref{fig:error_localization} and~\ref{fig:pass_at_1_comparison}), in Tables~\ref{tab:tau_sensitivity}, \ref{tab:pure_completion_results}, \ref{tab:mixture_training_pass_at_1_comparison} and~\ref{tab:mbpp_results_appendix}, and in the three-seed study. The second column covers the $\alpha$ sweep (Figure~\ref{fig:mixture_prob_ablation}) and the CDLM$+$ctw arm of Table~\ref{tab:ctw_ablation}; these runs use the recipe of the first column with a 2000-step learning-rate horizon (Appendix~\ref{app:lr_horizon}) and the objective weights shown, and the sweep ran on a different machine whose shard reading order was not recorded. The Sudoku model is described in Appendix~\ref{app:sudoku_exp}. Both 16B comparisons were trained on separate infrastructure. The MATH column is taken from the records of those runs. The ParallelBench models were trained in an environment whose configuration was not retained, so the table gives only the settings stated in Appendix~\ref{app:cdlm_parallelbench}, the objective weights and the seed.

In every comparison between an absorbing-only (MDLM) arm and a mixture (CDLM) arm, the two arms start from the same checkpoint, are trained on the same data, and share every setting in the table except $\alpha$ and $\lambda_{\mathrm{noise}}$. This holds at 16B as well: on MATH, MDLM is vanilla fine-tuning ($\alpha=0$) and CDLM is mixture fine-tuning ($\alpha=0.1$) of LLaDA2.0-mini-preview on the same corpus for one epoch, with all other settings identical. Every language-model CDLM in the paper uses $\alpha=0.1$, $\lambda_{\mathrm{noise}}=0.1$ and $\mathrm{ctw}=0$, except the arms of the two ablations that vary $\alpha$ or $\mathrm{ctw}$; the Sudoku model is described in Appendix~\ref{app:sudoku_exp}. $\alpha=0.1$ was fixed from preliminary experiments ($\alpha\in\{0.1,0.2,\ldots,1.0\}$) before the sweep in Figure~\ref{fig:mixture_prob_ablation} was run, and was not revisited; the sweep is a post-hoc sensitivity analysis.

\paragraph{Decoding temperature.}
All CRB repair runs decode greedily (temperature $0$) for every model: Dream-7B-Base, LLaDA-8B-Base, Open-dCoder-0.5B and all fine-tuned checkpoints. Every CRB run records its temperature in its output path or in its per-cell result file, and every recorded value is $0$.
In CRB repair the remasking threshold is $\tau=0.9$ throughout, except in the threshold sweep of Table~\ref{tab:tau_sensitivity}. From-scratch code generation samples at temperature $0.8$ (Appendix~\ref{app:scratch_eval}), and MATH decoding at 16B is greedy.

\paragraph{Compute.}
Each 2000-step 0.5B run of the main comparison took 57 to 65 minutes from the first to the last optimizer step on 4 A100-PCIe-40GB GPUs (65 to 77 minutes including model loading and the evaluations before and during training), and the CDLM$+$ctw run took 30 minutes on the same hardware. The two additional seeds of the seed study ran on 2 GPUs with gradient accumulation 2 at the same global batch. Each 2000-step LLaDA-8B LoRA run took 46 minutes on 4 A100-PCIe-40GB GPUs (1.38--1.39\,s per step, at most 23.9\,GiB allocated per GPU). The 16B MATH runs used 8 H100-80GB GPUs.

\begin{table}[!htbp]
\centering
\small
\setlength{\tabcolsep}{3pt}
\caption{Training and decoding settings per experiment. Objective weights are those of the CDLM arm; every MDLM arm has $\alpha=0$, so no position is replaced and $\lambda_{\mathrm{noise}}$ has no effect. Batch: micro-batch $\times$ GPUs $\times$ accumulation steps $=$ global batch, $\times$ maximum length. LR: peak; warmup; decay. ``--'': not recorded. The Sudoku settings are in Appendix~\ref{app:sudoku_exp}.}
\label{tab:hparams}
\begin{tabular}{p{1.55cm}p{2.4cm}p{2.3cm}p{2.3cm}p{2.15cm}p{2.0cm}}
\toprule
& \hpc{\textbf{0.5B code}\\ main comparison, seeds}
& \hpc{\textbf{0.5B code}\\ $\alpha$ sweep, CDLM$+$ctw}
& \hpc{\textbf{LLaDA-8B}\\ LoRA}
& \hpc{\textbf{16B}\\ MATH}
& \hpc{\textbf{16B}\\ ParallelBench} \\
\midrule
\multicolumn{6}{l}{\textit{Training}} \\
\hpc{Base model} & \hpc{Open-dCoder-0.5B} & \hpc{Open-dCoder-0.5B} & \hpc{LLaDA-8B-Base} & \hpc{LLaDA2.0-mini-preview} & \hpc{LLaDA2.0-mini} \\
\hpc{Data} & \hpc{Nemotron-SFT-Code} & \hpc{Nemotron-SFT-Code} & \hpc{Nemotron-SFT-Code, 12,000 documents (3.55M tokens), each seen twice} & \hpc{Nemotron-Math-v2, 300K samples} & \hpc{50,000 ParallelBench examples} \\
\hpc{Length} & \hpc{2000 steps} & \hpc{2000 steps} & \hpc{2000 steps} & \hpc{1 epoch (9,306 steps)} & \hpc{6 epochs} \\
\hpc{Batch} & \hpc{$3{\times}4{\times}1=12$, $\times\,4096$; seeds 1234, 2025: $3{\times}2{\times}2$} & \hpc{$3{\times}4{\times}1=12$, $\times\,4096$} & \hpc{$1{\times}4{\times}3=12$, $\times\,4096$} & \hpc{$4{\times}8{\times}1=32$, $\times\,2048$} & \hpc{--} \\
\hpc{Optimizer} & \hpc{AdamW (0.9, 0.95), wd 0.01, clip 1.0} & \hpc{same} & \hpc{AdamW (0.9, 0.95), wd 0.01 (not on 1-D), clip 1.0} & \hpc{AdamW (0.9, 0.999), wd 0.1, clip 1.0} & \hpc{--} \\
\hpc{LR} & \hpc{$3\times10^{-4}$; 20,345 steps; cosine to $3\times10^{-6}$ over 20,345,053 steps} & \hpc{$3\times10^{-4}$; 2 steps; cosine to $3\times10^{-6}$ over 2000 steps} & \hpc{$3\times10^{-4}$; 200 steps; cosine to 0 over 20,000 steps} & \hpc{$10^{-5}$; 3\%; cosine over the epoch} & \hpc{--} \\
\hpc{LR at step 2000} & \hpc{$2.95\times10^{-5}$ (9.83\% of peak)} & \hpc{$3\times10^{-6}$} & \hpc{$2.94\times10^{-4}$ (98.0\% of peak)} & \hpc{n/a} & \hpc{n/a} \\
\hpc{Precision} & \hpc{bf16 weights and optimizer states} & \hpc{same} & \hpc{bf16 base; fp32 LoRA weights and optimizer states} & \hpc{bf16 mixed precision} & \hpc{--} \\
\hpc{Trained weights} & \hpc{all except embeddings and LM head} & \hpc{same} & \hpc{LoRA ($r{=}64$, $\alpha{=}128$, dropout 0.05) on attention and MLP} & \hpc{--} & \hpc{--} \\
\hpc{Masking} & \hpc{$t\sim\mathcal{U}(0,1)$ per chunk, clamped to $[0.002,0.998]$} & \hpc{same} & \hpc{$p=(1{-}10^{-3})t+10^{-3}$, $t\sim\mathcal{U}(0,1)$} & \hpc{block diffusion, block size 32} & \hpc{block diffusion, block size 32, ratio $\sim\mathcal{U}[0,1]$} \\
\hpc{$\alpha$ / $\lambda_{\mathrm{noise}}$ / ctw} & \hpc{0.1 / 0.1 / 0} & \hpc{sweep: $\alpha\in\{0.04,\ldots,0.9\}$ / 0.1 / 0; ctw arm: 0.1 / 0.1 / 0.1} & \hpc{0.1 / 0.1 / 0} & \hpc{0.1 / 0.1 / 0} & \hpc{0.1 / 0.1 / 0} \\
\hpc{Seed} & \hpc{42; seed study: 1234, 2025} & \hpc{42} & \hpc{42} & \hpc{42 (framework default)} & \hpc{42 (framework default)} \\
\hpc{Hardware} & \hpc{4$\times$A100-40GB, DDP; seed study: 2$\times$} & \hpc{sweep: 4$\times$H200; ctw: 4$\times$A100-40GB; DDP} & \hpc{4$\times$A100-40GB, DDP} & \hpc{8$\times$H100-80GB, FSDP2, CPU offload} & \hpc{--} \\
\midrule
\multicolumn{6}{l}{\textit{Decoding}} \\
\hpc{CRB repair} & \hpc{$\tau=0.9$, temperature 0, batch size 1} & \hpc{same (sweep: HumanEval only)} & \hpc{$\tau=0.9$, temperature 0, batch size 1} & \hpc{n/a} & \hpc{n/a} \\
\hpc{From scratch} & \hpc{128 tokens in 128 steps, temperature 0.8, top-$k$ 200, 10 samples} & \hpc{n/a} & \hpc{Appendix~\ref{app:llada8b_guard}} & \hpc{threshold 0.9, remasking, block 32, $\le$2048 tokens, greedy, 4-shot} & \hpc{threshold 0.6, fixed or remask, block 32, $\le$64 tokens} \\
\bottomrule
\end{tabular}
\end{table}

\subsection{The Training Objective as Implemented}
\label{app:objective_impl}

Equation~\eqref{eq:mixture-loss} states the form shared by all experiments: a reconstruction term over masked positions $\mathcal{M}$ plus $\lambda_{\mathrm{noise}}$ times the mean cross-entropy over the uniformly replaced positions $\mathcal{N}$. The three trainers differ in how the reconstruction term is weighted and normalized. This subsection states what each one computes. Throughout, $\bm{x}$ is the clean sequence, $\bm{z}$ the corrupted input, $\ell_i=-\log p_\theta(x_i\mid\bm{z})$ the cross-entropy of the original token at position $i$, $\mathcal{C}$ the visible (unmasked) positions, and $\mathcal{C}\setminus\mathcal{N}$ the clean visible positions.

\paragraph{Corruption.}
Masking is applied first and replacement second, so replacement only touches visible tokens. Each visible token is selected for replacement independently with probability $\alpha$ and replaced by a token drawn uniformly from the vocabulary, excluding the mask token; the draw is repeated (up to five times) when it equals the original token, which leaves a residual collision probability below $10^{-30}$. At 0.5B the draw ranges over ids $0$ to $151{,}641$ of the tokenizer's $151{,}643$-token base vocabulary, so no added special token (including the mask) is ever drawn. At 8B it ranges over the $126{,}463$ non-mask rows of the $126{,}464$-row embedding. The target at every position is the original token.

\paragraph{Open-dCoder-0.5B.}
Documents are tokenized, terminated with an end-of-sequence token, cut into chunks of at most 4096 tokens and packed into micro-batches. For each chunk a mask ratio $t\sim\mathcal{U}(0,1)$ is drawn and clamped to $[1/500,\,1-1/500]$, and every token is masked independently with probability $t$. Open-dCoder keeps the next-token shift of its autoregressive initialization: with bidirectional attention, the output at position $i-1$ predicts token $i$, so the first token of a chunk is never a target. Let $\mathcal{S}=\mathcal{M}\cup\mathcal{N}$ be the masked and replaced target positions of a micro-batch, $t_i$ the mask ratio of the chunk containing $i$, and $\operatorname{sg}$ the stop-gradient operator. The trainer computes
\begin{equation}
\mathcal{L}_{0.5\mathrm{B}}
=\frac{1}{|\mathcal{S}|}\sum_{i\in\mathcal{S}}\ell_i
+\frac{1}{|\mathcal{S}|}\sum_{i\in\mathcal{S}}\frac{\operatorname{sg}\!\big(e^{-\ell_i}\big)}{t_i}\,\ell_i
+\lambda_{\mathrm{noise}}\,\frac{1}{|\mathcal{N}|}\sum_{i\in\mathcal{N}}\ell_i
+\mathrm{ctw}\,\frac{1}{|\mathcal{C}\setminus\mathcal{N}|}\sum_{i\in\mathcal{C}\setminus\mathcal{N}}\ell_i .
\label{eq:loss_05b}
\end{equation}
Each term is a token mean pooled over all chunks of the micro-batch (each denominator carries $+10^{-8}$), the noise and clean terms are omitted when their weight is $0$ or their set is empty, each micro-batch loss is divided by the number of gradient-accumulation steps, and losses are averaged over data-parallel ranks. The first term averages over masked and replaced positions together and carries no $1/t$ weighting. The second term is inherited from the Open-dLLM training code and is present in both arms: it is the only term weighted by $1/t$, and it weights each token by the model's detached probability $e^{-\ell_i}=p_\theta(x_i\mid\bm{z})$ of the original token. A replaced position therefore enters the first, second and third terms. Clean visible positions enter only the fourth term, whose weight is $0$ in every run except CDLM$+$ctw ($\mathrm{ctw}=0.1$). For MDLM ($\alpha=\lambda_{\mathrm{noise}}=0$), $\mathcal{S}=\mathcal{M}$ and the loss is the first two terms.

\paragraph{LLaDA-8B (LoRA).}
LLaDA is trained without a token shift. Each sequence draws $t\sim\mathcal{U}(0,1)$ and masks every token independently with probability $p=(1-10^{-3})\,t+10^{-3}$. The masking draws and the replacement draws come from separate random streams, so the masking pattern is identical in the MDLM and CDLM arms. With $B$ sequences of $L$ token positions in a micro-batch ($B=1$ here), the trainer computes LLaDA's absorbing loss plus the noise term:
\begin{equation}
\mathcal{L}_{8\mathrm{B}}
=\frac{1}{B\,L}\sum_{i\in\mathcal{M}}\frac{\ell_i}{p}
+\lambda_{\mathrm{noise}}\,\frac{1}{|\mathcal{N}|}\sum_{i\in\mathcal{N}}\ell_i ,
\label{eq:loss_8b}
\end{equation}
with the standard PyTorch cross-entropy, averaged over the three accumulation steps and over ranks. A replaced position enters only the noise term, and clean visible positions are never supervised ($\mathrm{ctw}=0$).

\paragraph{LLaDA2.0 (16B).}
The 16B runs use the block-diffusion fine-tuning code of LLaDA2.0 with block size 32 and the same two-term form.

\paragraph{Scale of $\lambda_{\mathrm{noise}}$.}
In the 0.5B and 8B implementations the noise term is a mean over the replaced positions, so its magnitude does not grow with the number of replaced tokens or with sequence length, and the reconstruction term is normalized per token as well. $\lambda_{\mathrm{noise}}$ therefore weights two per-token quantities and is comparable across sequence lengths. The same value, $\lambda_{\mathrm{noise}}=0.1$, is used for sequences of up to 4096 tokens at 0.5B and 8B and of up to 2048 tokens at 16B.

\subsection{Learning-Rate Horizon of the 0.5B Runs}
\label{app:lr_horizon}

All 0.5B runs use a peak learning rate of $3\times10^{-4}$, linear warmup over $0.1\%$ of the schedule, and cosine decay to $3\times10^{-6}$. They differ only in whether the schedule length is set explicitly, which produces two different optimization problems at the same step count. Reproducing a 0.5B result requires matching its horizon.

\paragraph{Long horizon: main comparison and seed study.}
The CDLM-0.5B and MDLM-0.5B runs, including the additional seeds, set no maximum step count. The training framework then derives the schedule length from its default token budget of $10^{12}$ tokens: $\lceil 10^{12}/(12\times4096)\rceil=20{,}345{,}053$ steps, of which $\lfloor0.001\times20{,}345{,}053\rfloor=20{,}345$ are warmup. The checkpoints used in the paper are taken at step 2000, so they are still in linear warmup: the learning rate at step 2000 is $3\times10^{-4}\times2000/20{,}345=2.949\times10^{-5}$, $9.83\%$ of the peak, and it rises linearly from $1.5\times10^{-8}$ over the first 2000 steps. Both arms logged exactly these values.

\paragraph{2000-step horizon: $\alpha$ sweep and clean-token arm.}
The $\alpha$-sweep runs of Figure~\ref{fig:mixture_prob_ablation}, including the CDLM row of Table~\ref{tab:ctw_ablation}, and the CDLM$+$ctw run set the maximum step count to 2000. Their schedule is exactly 2000 steps long: 2 warmup steps to the peak of $3\times10^{-4}$, then a full cosine decay that reaches $3\times10^{-6}$ at step 2000.

\paragraph{Consequences.}
The two regimes give different checkpoints at the same step count and are not interchangeable. The CDLM-versus-MDLM comparison is schedule-matched, since both arms use the long horizon. The $\alpha=0.1$ point of the sweep is a different model from CDLM-0.5B, trained with the 2000-step horizon, so sweep values are compared only with each other.
In Table~\ref{tab:ctw_ablation}, the CDLM row is the $\alpha=0.1$ checkpoint of the sweep, so both rows use the 2000-step horizon and differ in the training objective only by the clean-token weight (Appendix~\ref{app:stats_ctw} lists the remaining differences in training hardware and software). The main CDLM-0.5B checkpoint, shown for reference in the last row, uses the long horizon.
The LLaDA-8B runs use a third schedule, identical in both arms: 200 warmup steps and cosine decay over a 20,000-step horizon, stopped at step 2000, where the learning rate is $2.94\times10^{-4}$ ($98.0\%$ of the peak).

\subsection{From-Scratch Code Generation Protocol}
\label{app:scratch_eval}

The from-scratch results in Table~\ref{tab:pure_completion_results} (top) and Table~\ref{tab:mbpp_results_appendix} are produced with the lm-evaluation-harness~\citep{gao2024lmevalharness} (version 0.4.9.1, vendored in the Open-dLLM repository~\citep{opendllm2025}) and a model wrapper that calls Open-dCoder's diffusion sampler. The protocol is identical for the base model, MDLM-0.5B and CDLM-0.5B.

\paragraph{Tasks and prompts.}
All tasks are zero-shot. HumanEval (164 problems) and HumanEval+ (164 problems, EvalPlus tests~\citep{liu2023humanevalplus}) prompt with the function signature and docstring. MBPP (the 500 test problems of~\citet{austin2021mbpp}) and MBPP+ (378 problems) prompt with the task description, the three test assertions of the problem and an opening code fence. Both MBPP tasks are scored against those three assertions; for MBPP+ this means that the extended EvalPlus test inputs are not used, so its numbers are not comparable with EvalPlus MBPP+ scores.

\paragraph{Generation.}
The prompt is followed by 128 mask tokens, which are decoded in 128 steps. At every step all masked positions are sampled with temperature $0.8$ and top-$k$ $200$, the confidence $p_\theta(z_i\mid\bm{z})$ of every generated position is computed from the unscaled logits, and the $128-s$ lowest-confidence generated positions are masked again after step $s$. One position is therefore committed per step on net, and any generated token can be revised. This is the remask-based decoding of~\citet{peng2025path} with the model's own confidence as the remasking score. The vanilla decoder recomputes the confidences at every step; the ReMDM variant~\citep{wang2025remasking} keeps each token's confidence from the step at which it was unmasked. The run configuration also passes a top-$p$ value of $0.95$, which the wrapper does not use.

\paragraph{Stop criterion and post-processing.}
Each completion is truncated at the first occurrence of any of the task's stop strings. For HumanEval and HumanEval+ these are \texttt{"\textbackslash nclass"}, \texttt{"\textbackslash ndef"}, \texttt{"\textbackslash n\#"}, \texttt{"\textbackslash nif"} and \texttt{"\textbackslash nprint"}; for MBPP and MBPP+ they are \texttt{"\textbackslash nclass"}, \texttt{"\textbackslash n\#"}, \texttt{"\textbackslash nif"} and \texttt{"\textbackslash nprint"}, since the model writes the whole function. For HumanEval(+), the prompt and the completion (the part inside a code fence, if one is present) are passed through the harness's syntax-based sanitizer before execution; for MBPP(+), code fences are removed. The narrower stop list of the CRB repair evaluator is not used for these numbers.

\paragraph{Pass@$k$.}
Ten samples are drawn per problem, and Pass@1 and Pass@10 are the unbiased estimates of~\citet{chen2021evaluatinglargelanguagemodels} from these ten samples, averaged over problems. Evaluation ran in 4 data-parallel processes with the harness's default seeds (Python 0, NumPy 1234, PyTorch 1234) and batch size 5 for HumanEval(+) and 1 for MBPP(+). Because 378 MBPP+ problems do not split evenly over 4 processes, the harness pads the two short processes by repeating their last request, so two MBPP+ problems are scored from 20 samples. The released evaluation launcher replays this protocol and reproduces the recorded CDLM-0.5B HumanEval run sample for sample.

\paragraph{Uncertainty.}
Each arm is a single training run with ten samples per problem. A paired bootstrap over the 164 problems (10{,}000 resamples of the per-problem Pass@1 differences) gives these 95\% intervals for CDLM minus each arm in Pass@1. On HumanEval+, CDLM is above MDLM by $+0.036$ $[+0.014,+0.059]$ and above base by $+0.038$ $[+0.004,+0.072]$ under vanilla decoding, and by $+0.028$ $[+0.008,+0.050]$ and $+0.036$ $[+0.002,+0.071]$ under ReMDM. On HumanEval, CDLM is above MDLM under vanilla decoding ($+0.026$ $[+0.004,+0.049]$), but the intervals include zero against base under both decoders ($+0.027$ $[-0.005,+0.059]$; $+0.029$ $[-0.004,+0.062]$) and against MDLM under ReMDM ($+0.013$ $[-0.009,+0.035]$).

\subsection{Distribution Shift Measured by Masked-Reconstruction Loss}
\label{app:mbpp_shift}

Both fine-tuned 0.5B arms stay below the base model on MBPP and MBPP+ in Pass@1 (Table~\ref{tab:mbpp_results_appendix}). To test whether fine-tuning on Nemotron-SFT-Code shifts the model away from MBPP-style programs, we measure the masked-reconstruction loss of the base model, MDLM-0.5B and CDLM-0.5B on five distributions.

\paragraph{Setup.}
The distributions are a held-out slice of the fine-tuning corpus (2,000 Nemotron-SFT-Code documents from the end of the last shard, which the 2000-step runs do not reach) and the canonical reference programs of the four benchmarks, built with the same program constructor as CRB. The loss is the cross-entropy at masked positions, averaged over masked positions, with bidirectional attention and the next-token shift used in training. Sequences are tokenized as in training and capped at 512 tokens. Every sequence is evaluated at mask ratios $\{0.15,0.3,0.5,0.7\}$ with three mask seeds each, and the 12 losses are averaged. Mask patterns depend only on the sequence, the ratio and the seed, so all three models see identical corruptions and the model differences are paired. Because the loss falls as more context is visible, each Nemotron document is truncated to a length drawn from the token-length distribution of the four benchmarks.

The masked-reconstruction loss is the quantity the MDLM objective trains on, which makes it the appropriate likelihood proxy for a masked diffusion model; a left-to-right perplexity does not correspond to how these models are trained or decoded. Absolute losses should not be compared across columns: the Nemotron slice contains natural-language reasoning traces as well as code and is intrinsically harder to reconstruct than a short canonical Python solution. The informative quantity is the change within each column.

\begin{table}[!htbp]
\centering
\small
\setlength{\tabcolsep}{3.5pt}
\caption{Masked-reconstruction loss (nats/token) on a held-out slice of the fine-tuning corpus (Nemotron) and on the reference programs of the four code benchmarks. \textbf{Top:} mean $\pm$ standard deviation over sequences. \textbf{Middle:} paired difference to the base model, mean $\pm$ half-width of the 95\% confidence interval over sequences; a positive value means the model became worse at reconstructing that distribution. \textbf{Bottom:} two-sided one-sample $t$-test of the paired differences. Each sequence's loss averages mask ratios $\{0.15,0.3,0.5,0.7\}\times3$ mask seeds.}
\label{tab:mbpp_shift}
\begin{tabular}{lccccc}
\toprule
& \textbf{Nemotron} & \textbf{HumanEval} & \textbf{HumanEval+} & \textbf{MBPP} & \textbf{MBPP+} \\
\midrule
Sequences & 2000 & 164 & 163 & 499 & 378 \\
Mean tokens & 91.8 & 165.8 & 165.8 & 79.0 & 65.3 \\
\midrule
Base & $1.341\pm0.670$ & $0.771\pm0.338$ & $0.838\pm0.330$ & $0.834\pm0.398$ & $0.889\pm0.373$ \\
MDLM & $1.097\pm0.646$ & $0.785\pm0.340$ & $0.850\pm0.334$ & $0.845\pm0.406$ & $0.899\pm0.381$ \\
CDLM & $1.072\pm0.660$ & $0.808\pm0.352$ & $0.883\pm0.350$ & $0.886\pm0.434$ & $0.950\pm0.412$ \\
\midrule
MDLM$-$Base & $-0.243\pm0.007$ & $+0.013\pm0.004$ & $+0.012\pm0.004$ & $+0.011\pm0.003$ & $+0.010\pm0.005$ \\
CDLM$-$Base & $-0.269\pm0.008$ & $+0.037\pm0.006$ & $+0.045\pm0.006$ & $+0.053\pm0.006$ & $+0.061\pm0.010$ \\
\midrule
$p$, MDLM & $<10^{-300}$ & $5\times10^{-10}$ & $1\times10^{-7}$ & $5\times10^{-10}$ & $1.1\times10^{-4}$ \\
$p$, CDLM & $<10^{-300}$ & $6\times10^{-22}$ & $3\times10^{-29}$ & $2\times10^{-49}$ & $1\times10^{-29}$ \\
\bottomrule
\end{tabular}
\end{table}

\paragraph{Results.}
Fine-tuning moves both arms toward the fine-tuning corpus: the loss on the held-out Nemotron slice falls by $0.243$ (MDLM) and $0.269$ (CDLM) nats/token. Both arms also move away from all four benchmarks, and every difference in Table~\ref{tab:mbpp_shift} is statistically significant. For CDLM the movement away from the benchmarks is asymmetric in the direction of the MBPP regression: the MBPP family (MBPP and MBPP+ pooled) degrades by $+0.0562$ nats/token against $+0.0407$ for the HumanEval family (Welch's $t$-test on the per-sequence differences, $p=2.1\times10^{-5}$). MBPP programs are shorter than HumanEval programs (65 to 79 tokens on average against 166), and the loss depends on how much context is visible, so we repeated the comparison on programs of 40 to 200 tokens only; the asymmetry remains ($+0.0515$ against $+0.0386$, $p=7.3\times10^{-4}$). The asymmetry is specific to CDLM. MDLM shows none: its MBPP family moves slightly less than its HumanEval family (difference $-0.0021$, $p=0.31$).

The shift is mild. CDLM's MBPP loss rises by about 6\% of the base model's MBPP loss ($+0.053$ on $0.834$), and we do not consider it sufficient on its own to account for the Pass@1 gap to the base model. We therefore treat it as a contributing factor rather than as the explanation. Relative to the base model, both fine-tuned arms regress on MBPP and MBPP+ under identical data and compute (Table~\ref{tab:mbpp_results_appendix}).

\subsection{Use of Large Language Models}
\label{app:llm_use}

LLM coding agents were used to implement and orchestrate the evaluation pipelines and analysis scripts, and to assist with writing and editing. No reported number comes from an LLM judgment: every Pass@1 and Pass@$k$ is obtained by executing the benchmark's test suite, MATH accuracy by programmatic answer matching, and ParallelBench accuracy by the benchmark's programmatic scorer, and all localization metrics are computed from model probabilities. The only LLM that produces reported outputs is gpt-4.1-nano~\citep{openai2025gpt41}, used as the autoregressive self-editing baseline of Appendix~\ref{app:ar_baseline}, whose repaired programs are scored by the same test suites.

%% file: appendix/statistics.tex
\section{Statistical Reliability}
\label{app:stats}

This appendix quantifies two sources of uncertainty in the 0.5B controlled comparison (Section~\ref{sec:main_comparison}) and in the clean-token ablation (Section~\ref{sec:ctw}). Appendix~\ref{app:stats_paired} treats the finite sample of CRB problems with paired tests over per-problem outcomes, Appendix~\ref{app:stats_exact} repeats the main test on corrupted programs that are byte-identical across arms, Appendix~\ref{app:stats_ctw} tests the clean-token ablation of Table~\ref{tab:ctw_ablation}, and Appendix~\ref{app:stats_seeds} measures variation across training seeds. Appendix~\ref{app:stats_scope} states what these analyses do not cover.

\subsection{Paired tests over CRB problems}
\label{app:stats_paired}

\paragraph{Design.}
The unit of analysis is one corrupted program. We compare CDLM-0.5B with three arms built on the same Open-dCoder-0.5B checkpoint: the checkpoint itself (Base), the absorbing-only post-trained model (MDLM), and the clean-token ablation (CDLM$+$ctw, Table~\ref{tab:ctw_ablation}). Every arm is decoded with the confidence-thresholded remask rule at $\tau=0.9$ and temperature $0$, so each arm has one deterministic pass/fail outcome per program and refinement depth $T\in\{1,2,3,4\}$. The test family covers the four CRB datasets (HumanEval, HumanEval+, MBPP, MBPP+), the three error types (operator, identifier, literal) and $n_{\mathrm{replace}}\in\{1,\dots,5\}$. Within each dataset, error type and $n_{\mathrm{replace}}$, we pair the outcomes of two arms by task identifier and occurrence index, keeping the tasks present in both arms' instance files (the files differ slightly in size across arms; see Appendix~\ref{app:stats_exact}). This gives $N=9{,}087$ pairs for CDLM versus MDLM, $9{,}092$ for CDLM versus Base and $9{,}052$ for CDLM versus CDLM$+$ctw at each $T$.

For each comparison we report the difference in Pass@1, always CDLM minus the comparator and pooled over items, an exact two-sided McNemar test on the discordant pairs~\citep{mcnemar1947note}, and a 95\% percentile interval from 10{,}000 paired bootstrap resamples of items~\citep{efron1993introduction}. The tables in the main text macro-average over the 12 dataset $\times$ error-type cells, whereas these tests pool items, which is the unit McNemar's test requires. The two views describe the same outcomes: macro-averaging the per-file outcomes used here at $n_{\mathrm{replace}}=1$ reproduces the main-text CDLM-0.5B values exactly ($0.188/0.279/0.281/0.286$ at $T=1,\dots,4$).

LLaDA-8B-Base and Dream-7B-Base are not arms of this controlled comparison (they are different backbones, evaluated without post-training), so we do not pair them with the 0.5B arms. The LLaDA-8B comparison of Section~\ref{sec:llada8b} uses CRB files built with the LLaDA tokenizer, which yields a different item set, and is not part of this family.

\paragraph{CDLM versus MDLM and Base.}
Table~\ref{tab:stats_paired} pools all four datasets, three error types and five corruption levels. CDLM exceeds the matched MDLM at every depth, by $+0.029$ at $T=1$ and by $+0.062$ to $+0.070$ at $T=2,3,4$ (all $p<10^{-28}$), and exceeds Base by $+0.026$ to $+0.061$ (all $p<10^{-20}$). Pooled over error types and corruption levels, the difference from MDLM is significant ($p<0.05$) in each of the four datasets at every depth; the weakest of these 16 tests is MBPP+ at $T=1$ ($+0.014$, $p=0.014$). The same holds against Base (weakest: MBPP+ at $T=1$, $+0.012$, $p=0.038$). Pooled over datasets and corruption levels instead, the only slices that are not significant are literal corruptions at $T=1$ ($+0.006$, $p=0.13$ against MDLM; $+0.002$, $p=0.73$ against Base).

\begin{table}[!htbp]
\centering
\small
\caption{Paired tests over CRB problems, pooled over 4 datasets $\times$ 3 error types $\times$ $n_{\mathrm{replace}}\in\{1,\dots,5\}$ ($\tau=0.9$, greedy decoding). $\Delta$ is CDLM minus the comparator in Pass@1 (pooled over items); $b/c$ are the discordant counts (only CDLM passes / only the comparator passes); CI from 10{,}000 paired bootstrap resamples; $p$ from the exact McNemar test.}
\label{tab:stats_paired}
\begin{tabular}{crccrcrr}
\toprule
$T$ & $N$ & CDLM & Comparator & \multicolumn{1}{c}{$\Delta$} & 95\% CI & \multicolumn{1}{c}{$b/c$} & \multicolumn{1}{c}{$p$} \\
\midrule
\multicolumn{8}{l}{\emph{CDLM vs.\ MDLM}} \\
1 & 9{,}087 & 0.095 & 0.066 & $+0.029$ & $[+0.024, +0.035]$ & 424/158 & $4.6\times10^{-29}$ \\
2 & 9{,}087 & 0.216 & 0.147 & $+0.070$ & $[+0.062, +0.078]$ & 1032/399 & $8.4\times10^{-65}$ \\
3 & 9{,}087 & 0.225 & 0.163 & $+0.062$ & $[+0.054, +0.070]$ & 1018/453 & $3.6\times10^{-50}$ \\
4 & 9{,}087 & 0.231 & 0.164 & $+0.067$ & $[+0.059, +0.075]$ & 1052/442 & $1.6\times10^{-57}$ \\
\midrule
\multicolumn{8}{l}{\emph{CDLM vs.\ Base}} \\
1 & 9{,}092 & 0.095 & 0.069 & $+0.026$ & $[+0.020, +0.031]$ & 420/188 & $2.4\times10^{-21}$ \\
2 & 9{,}092 & 0.215 & 0.154 & $+0.061$ & $[+0.053, +0.070]$ & 1033/477 & $2.5\times10^{-47}$ \\
3 & 9{,}092 & 0.224 & 0.169 & $+0.055$ & $[+0.046, +0.063]$ & 1037/541 & $3.2\times10^{-36}$ \\
4 & 9{,}092 & 0.229 & 0.170 & $+0.060$ & $[+0.051, +0.069]$ & 1069/526 & $9.2\times10^{-43}$ \\
\bottomrule
\end{tabular}
\end{table}

\paragraph{The 60-test family.}
Table~\ref{tab:stats_grid} breaks the three comparisons down by $n_{\mathrm{replace}}$ and $T$, each test pooled over the 4 datasets and 3 error types. These $3\times5\times4=60$ tests form one family, and we control the false discovery rate with the Benjamini--Hochberg procedure~\citep{benjamini1995controlling} at $q<0.05$. Fifty of the 60 tests reject: 18 of 20 against MDLM, 18 of 20 against Base and 14 of 20 against CDLM$+$ctw. Two of the 14 rejections against CDLM$+$ctw favor CDLM$+$ctw ($n_{\mathrm{replace}}=1$ and $3$ at $T=1$); Appendix~\ref{app:stats_ctw} separates the clean-token term from the learning-rate schedule in this comparison. No test against MDLM or Base finds CDLM significantly worse, in this family or at any other pooling level we computed (per dataset, per error type, and per dataset $\times$ error type $\times$ $n_{\mathrm{replace}}$ $\times$ $T$ without multiplicity correction).

\begin{table}[!htbp]
\centering
\small
\caption{The 60-test family: Pass@1 difference (CDLM minus comparator) for each $n_{\mathrm{replace}}$ and refinement depth $T$, pooled over 4 datasets $\times$ 3 error types. $N$ is the number of paired programs (constant across $T$). CDLM is the main CDLM-0.5B checkpoint; its comparison with CDLM$+$ctw also reflects their different learning-rate schedules (Appendix~\ref{app:stats_ctw}). Entries in parentheses do not survive Benjamini--Hochberg correction over the 60 tests at $q<0.05$; all other entries do.}
\label{tab:stats_grid}
\begin{tabular}{llrrrrr}
\toprule
Comparator & $n_{\mathrm{replace}}$ & $N$ & $T=1$ & $T=2$ & $T=3$ & $T=4$ \\
\midrule
MDLM & 1 & 3{,}180 & $+0.045$ & $+0.042$ & $+0.036$ & $+0.045$ \\
     & 2 & 2{,}429 & $+0.030$ & $+0.089$ & $+0.079$ & $+0.081$ \\
     & 3 & 1{,}478 & $+0.020$ & $+0.085$ & $+0.070$ & $+0.076$ \\
     & 4 & 1{,}154 & $(+0.011)$ & $+0.097$ & $+0.089$ & $+0.092$ \\
     & 5 &    846 & $(+0.007)$ & $+0.053$ & $+0.058$ & $+0.060$ \\
\midrule
Base & 1 & 3{,}174 & $+0.039$ & $+0.040$ & $+0.032$ & $+0.035$ \\
     & 2 & 2{,}436 & $+0.030$ & $+0.076$ & $+0.074$ & $+0.079$ \\
     & 3 & 1{,}475 & $(+0.010)$ & $+0.070$ & $+0.055$ & $+0.064$ \\
     & 4 & 1{,}154 & $+0.011$ & $+0.077$ & $+0.076$ & $+0.079$ \\
     & 5 &    853 & $(+0.009)$ & $+0.061$ & $+0.053$ & $+0.064$ \\
\midrule
CDLM$+$ctw & 1 & 3{,}164 & $-0.026$ & $+0.059$ & $+0.074$ & $+0.078$ \\
     & 2 & 2{,}404 & $(-0.006)$ & $+0.060$ & $+0.062$ & $+0.071$ \\
     & 3 & 1{,}474 & $-0.016$ & $+0.033$ & $+0.039$ & $+0.051$ \\
     & 4 & 1{,}151 & $(-0.006)$ & $+0.048$ & $+0.045$ & $+0.048$ \\
     & 5 &    859 & $(-0.009)$ & $(+0.026)$ & $(+0.026)$ & $(+0.022)$ \\
\bottomrule
\end{tabular}
\end{table}

\paragraph{Two notes on the non-significant tests.}
First, ten of the 60 tests do not reject. Seven are at $T=1$: against MDLM at $n_{\mathrm{replace}}=4,5$, against Base at $n_{\mathrm{replace}}=3,5$, and against CDLM$+$ctw at $n_{\mathrm{replace}}=2,4,5$. In six of these seven, the pooled Pass@1 of both arms lies between $0.007$ and $0.041$, so few programs separate the arms; the seventh (CDLM$+$ctw, $n_{\mathrm{replace}}=2$) has Pass@1 $0.081$ against $0.087$. The other three are the comparison with CDLM$+$ctw at $n_{\mathrm{replace}}=5$ and $T=2,3,4$, the smallest CDLM$+$ctw slice ($N=859$), where CDLM leads by $+0.022$ to $+0.026$ without reaching significance.

Second, the difference between CDLM-0.5B and CDLM$+$ctw, which also reflects their different learning-rate schedules (Appendix~\ref{app:stats_ctw}), is not uniform across datasets (Table~\ref{tab:stats_ctw_dataset}). Pooled over error types and corruption levels, CDLM's advantage over CDLM$+$ctw at $T\ge2$ is large on HumanEval+ ($+0.202$ to $+0.230$, $p<10^{-55}$), clear on MBPP+ ($+0.034$ to $+0.060$, $p\le0.0012$), small on MBPP ($+0.015$ to $+0.021$, $0.004<p\le0.042$), and absent on HumanEval ($-0.007$ to $+0.012$, $p\ge0.40$). At $T=1$, CDLM$+$ctw is ahead on all four datasets, significantly on HumanEval ($-0.020$, $p=0.037$) and MBPP ($-0.018$, $p=0.001$). The $T\ge2$ side of this difference therefore rests mainly on HumanEval+, MBPP+ and MBPP, and the $T=1$ side on HumanEval and MBPP.

\begin{table}[!htbp]
\centering
\small
\caption{CDLM-0.5B (main checkpoint, longer schedule) versus CDLM$+$ctw by dataset: Pass@1 difference CDLM minus CDLM$+$ctw (first row) and exact McNemar $p$ (second row, in parentheses), pooled over 3 error types $\times$ $n_{\mathrm{replace}}\in\{1,\dots,5\}$. Negative values favor CDLM$+$ctw. $N$ is constant across $T$.}
\label{tab:stats_ctw_dataset}
\begin{tabular}{lrrrrr}
\toprule
Dataset & $N$ & $T=1$ & $T=2$ & $T=3$ & $T=4$ \\
\midrule
HumanEval  & 1{,}473 & $-0.020$ & $-0.007$ & $-0.001$ & $+0.012$ \\
           &         & ($0.037$) & ($0.60$) & ($1.0$) & ($0.41$) \\
HumanEval+ & 1{,}567 & $-0.010$ & $+0.202$ & $+0.229$ & $+0.230$ \\
           &         & ($0.25$) & ($1.3{\times}10^{-56}$) & ($1.5{\times}10^{-69}$) & ($4.9{\times}10^{-69}$) \\
MBPP       & 3{,}983 & $-0.018$ & $+0.021$ & $+0.015$ & $+0.017$ \\
           &         & ($0.0010$) & ($0.0050$) & ($0.042$) & ($0.021$) \\
MBPP+      & 2{,}029 & $-0.009$ & $+0.034$ & $+0.047$ & $+0.060$ \\
           &         & ($0.23$) & ($0.0012$) & ($3.2{\times}10^{-6}$) & ($3.0{\times}10^{-9}$) \\
\bottomrule
\end{tabular}
\end{table}

\subsection{Identical-stimulus subset}
\label{app:stats_exact}

CRB instances were generated separately for each model's evaluation, with the same generator and settings. Operator corruptions do not depend on process state, so every 0.5B arm received byte-identical operator instances. The identifier and literal paths of the generator build their candidate lists by iterating over Python sets, whose iteration order for strings changes from one Python process to the next (string hash randomization); runs with the same random seed therefore drew different identifier and literal corruptions for the same task, and because whether an instance is retained depends on the corruption drawn, the instance files also differ slightly in size. Over $n_{\mathrm{replace}}=1,\dots,5$, the fraction of CDLM's instances that are byte-identical to MDLM's is 100\% for operator, 26.7\% for identifier and 45.2\% for literal corruptions. The released benchmark files (Appendix~\ref{app:release}) are the canonical CRB instances.

Restricting the comparison to programs whose corrupted text is byte-identical in both arms gives $N=6{,}204$ pairs for CDLM versus MDLM at each $T$ ($4{,}385$ operator, $554$ identifier and $1{,}265$ literal) and $N=6{,}133$ for CDLM versus Base. The conclusions do not change (Table~\ref{tab:stats_exact}). At $T=1$, CDLM exceeds MDLM by $+0.027$ ($[+0.021,+0.032]$, $p=4.4\times10^{-22}$), and at $T=2,3,4$ by $+0.058$ to $+0.068$ (all $p<10^{-32}$). The difference between CDLM-0.5B and CDLM$+$ctw, which also reflects their learning-rate schedules (Appendix~\ref{app:stats_ctw}), has the same sign on this subset ($N=6{,}220$; $T=1$: $-0.016$, $p=1.5\times10^{-4}$; $T=4$: $+0.066$, $p=6.8\times10^{-26}$).

\begin{table}[!htbp]
\centering
\small
\caption{Identical-stimulus subset: only programs whose corrupted text is byte-identical in both arms, pooled over 4 datasets $\times$ 3 error types $\times$ $n_{\mathrm{replace}}\in\{1,\dots,5\}$. $\Delta$ is CDLM minus the comparator in Pass@1; $p$ from the exact McNemar test.}
\label{tab:stats_exact}
\begin{tabular}{crcrrcr}
\toprule
& \multicolumn{3}{c}{CDLM vs.\ MDLM ($N=6{,}204$)} & \multicolumn{3}{c}{CDLM vs.\ Base ($N=6{,}133$)} \\
\cmidrule(lr){2-4}\cmidrule(lr){5-7}
$T$ & \multicolumn{1}{c}{$\Delta$} & 95\% CI & \multicolumn{1}{c}{$p$} & \multicolumn{1}{c}{$\Delta$} & 95\% CI & \multicolumn{1}{c}{$p$} \\
\midrule
1 & $+0.027$ & $[+0.021, +0.032]$ & $4.4\times10^{-22}$ & $+0.023$ & $[+0.017, +0.028]$ & $2.1\times10^{-15}$ \\
2 & $+0.068$ & $[+0.059, +0.077]$ & $3.9\times10^{-46}$ & $+0.055$ & $[+0.045, +0.065]$ & $1.5\times10^{-28}$ \\
3 & $+0.058$ & $[+0.049, +0.068]$ & $6.3\times10^{-33}$ & $+0.051$ & $[+0.041, +0.060]$ & $3.3\times10^{-23}$ \\
4 & $+0.065$ & $[+0.056, +0.075]$ & $1.4\times10^{-40}$ & $+0.056$ & $[+0.046, +0.066]$ & $1.3\times10^{-27}$ \\
\bottomrule
\end{tabular}
\end{table}

\subsection{Clean-token ablation}
\label{app:stats_ctw}

\paragraph{Matched learning-rate schedules (Table~\ref{tab:ctw_ablation}).}
Both arms of Table~\ref{tab:ctw_ablation} were fine-tuned for 2000 steps with the same fully decaying cosine schedule (Appendix~\ref{app:lr_horizon}) and differ in the training objective only by the clean-token term. Both refine the same 3{,}213 CRB programs ($n_{\mathrm{replace}}=1$), so every program is paired. Table~\ref{tab:stats_ctw} gives the paired tests. At $T=1$ the two arms do not differ significantly ($-0.012$, $p=0.081$). At every $T\ge2$ CDLM is ahead by $+0.016$ to $+0.021$ ($p=0.044$, $0.0055$ and $0.040$); only $T=3$ would survive a Bonferroni correction over the four depths. The difference comes mainly from HumanEval+ and MBPP+ (macro over error types at $T=4$: $+0.054$ and $+0.033$), while HumanEval and MBPP differ by less than $0.005$. The CDLM arm of this comparison is the $\alpha=0.1$ checkpoint of the mixture-probability sweep; it was trained on a different machine (4 H200 GPUs, transformers 4.54.1) from the CDLM$+$ctw arm (4 A100 GPUs, transformers 4.57.0), with the same configuration, and the shard reading order of the sweep runs was not recorded. Each arm is a single training run.

\begin{table}[!htbp]
\centering
\small
\caption{Clean-token ablation with matched learning-rate schedules (Table~\ref{tab:ctw_ablation}; $n_{\mathrm{replace}}=1$, pooled over 4 datasets $\times$ 3 error types, $N=3{,}213$ paired programs). $\Delta$ is CDLM minus CDLM$+$ctw in Pass@1 (pooled over items, so the levels differ from the macro averages of Table~\ref{tab:ctw_ablation}); $b/c$ are the discordant counts; CI from 10{,}000 paired bootstrap resamples; $p$ from the exact McNemar test.}
\label{tab:stats_ctw}
\begin{tabular}{crccrcrr}
\toprule
$T$ & $N$ & CDLM & CDLM$+$ctw & \multicolumn{1}{c}{$\Delta$} & 95\% CI & \multicolumn{1}{c}{$b/c$} & \multicolumn{1}{c}{$p$} \\
\midrule
1 & 3{,}213 & 0.185 & 0.196 & $-0.012$ & $[-0.025, +0.001]$ & 206/244 & $0.081$ \\
2 & 3{,}213 & 0.214 & 0.198 & $+0.016$ & $[+0.000, +0.031]$ & 334/283 & $0.044$ \\
3 & 3{,}213 & 0.211 & 0.190 & $+0.021$ & $[+0.006, +0.036]$ & 335/266 & $0.0055$ \\
4 & 3{,}213 & 0.205 & 0.190 & $+0.016$ & $[+0.001, +0.031]$ & 323/272 & $0.040$ \\
\bottomrule
\end{tabular}
\end{table}

\paragraph{Comparison with the main checkpoint.}
The main CDLM-0.5B checkpoint was trained with a much longer schedule (Appendix~\ref{app:lr_horizon}). Compared directly with CDLM$+$ctw, it shows a larger and two-sided difference: CDLM$+$ctw is better at $T=1$ ($-0.026$, $[-0.041,-0.011]$, $p=8.1\times10^{-4}$; $N=3{,}164$ programs paired by task) and worse at every $T\ge2$, by up to $+0.078$ at $T=4$ ($[+0.060,+0.096]$, $p=3.4\times10^{-17}$). On the same instance files, CDLM-0.5B is ahead of the matched-schedule CDLM arm by $+0.050$ to $+0.067$ at $T\ge2$ ($p\le3.8\times10^{-8}$), so most of this larger difference reflects the learning-rate schedule rather than the clean-token term. The CDLM$+$ctw comparisons in Tables~\ref{tab:stats_grid} and~\ref{tab:stats_ctw_dataset} use the main checkpoint and should be read in the same way.

\subsection{Training seeds}
\label{app:stats_seeds}

\paragraph{Setup.}
We retrained both arms of the 0.5B controlled comparison with two further seeds (1234 and 2025), which together with the published seed-42 runs gives three training seeds per arm. All other settings match the published runs: base checkpoint, corpus, objective settings ($\alpha$, $\lambda_{\mathrm{noise}}$, ctw), frozen layers, sequence length, optimizer, learning-rate schedule (the derived schedule is identical, with the same learning rate at step 2000), global batch size 12, and the step-2000 checkpoint. Seeds 1234 and 2025 were trained on 2 GPUs (micro-batch $3\times2$ GPUs $\times$ 2 accumulation steps), whereas both seed-42 runs used 4 GPUs ($3\times4\times1$); global batch, schedule and corpus are identical, and only the assignment of data to GPUs differs. The new runs used a later revision of the training code whose changes are confined to logging, communication timeouts and checkpoint naming.

All six checkpoints were evaluated with one frozen, hash-verified copy of the evaluation code (batch size 1, temperature $0$, $\tau=0.9$, $n_{\mathrm{replace}}=1$, $T\in\{1,4\}$) on one fixed set of CRB instance files, byte-identical across the six evaluations ($3{,}213$ programs per checkpoint and depth). Metrics are macro-averaged over the 12 dataset $\times$ error-type cells, as in the main text. The confidence gap (computed per program as the mean confidence $c_i=p_\theta(z_i\mid\bm{z})$ over clean body positions minus the confidence at the corrupted position, then averaged over programs; Table~\ref{tab:confidence_refinement_benchmark} instead pools all body positions of a cell) and Hit@1 (top-1 hit rate) are computed from the first forward pass over the corrupted program, before any edit, and therefore do not depend on $T$.

\begin{table}[!htbp]
\centering
\small
\caption{Three training seeds per arm (0.5B, $n_{\mathrm{replace}}=1$, $\tau=0.9$, macro average over the 12 CRB cells). The number of training GPUs is given in parentheses; the global batch is 12 in every run. Values have four decimals so that the smallest SD is visible. SD is the sample standard deviation across the three training seeds, not the standard error of the mean. Pooled SD is $\sqrt{(s^2_{\mathrm{CDLM}}+s^2_{\mathrm{MDLM}})/2}$; the min./max.\ row is the worst CDLM seed minus the best MDLM seed. Gap and Hit@1 come from the first forward pass and do not depend on $T$.}
\label{tab:stats_seeds}
\setlength{\tabcolsep}{4pt}
\begin{tabular}{llrrrr}
\toprule
& & \multicolumn{2}{c}{Pass@1} & & \\
\cmidrule(lr){3-4}
Arm & Seed (GPUs) & $T=1$ & $T=4$ & Gap & Hit@1 \\
\midrule
CDLM & 42 (4) & 0.1888 & 0.2853 & 0.1636 & 0.2277 \\
     & 1234 (2) & 0.1829 & 0.2884 & 0.1497 & 0.2126 \\
     & 2025 (2) & 0.1943 & 0.2836 & 0.1664 & 0.2295 \\
     & mean $\pm$ SD & $0.1887{\pm}0.0057$ & $0.2858{\pm}0.0024$ & $0.1599{\pm}0.0090$ & $0.2233{\pm}0.0093$ \\
\midrule
MDLM & 42 (4) & 0.1386 & 0.2445 & 0.0976 & 0.1470 \\
     & 1234 (2) & 0.1393 & 0.2510 & 0.0951 & 0.1430 \\
     & 2025 (2) & 0.1389 & 0.2349 & 0.0944 & 0.1435 \\
     & mean $\pm$ SD & $0.1390{\pm}0.0003$ & $0.2435{\pm}0.0081$ & $0.0957{\pm}0.0017$ & $0.1445{\pm}0.0022$ \\
\midrule
\multicolumn{2}{l}{Difference of means} & $+0.0497$ & $+0.0423$ & $+0.0642$ & $+0.0788$ \\
\multicolumn{2}{l}{Difference / pooled SD} & 12.4 & 7.1 & 9.9 & 11.7 \\
\multicolumn{2}{l}{Min.\ CDLM $-$ max.\ MDLM} & $+0.0436$ & $+0.0326$ & $+0.0520$ & $+0.0656$ \\
\multicolumn{2}{l}{Welch $p$}                      & 0.0042 & 0.0077 & 0.0051 & 0.0031 \\
\bottomrule
\end{tabular}
\end{table}

\paragraph{Results.}
Table~\ref{tab:stats_seeds} lists every run. Error bars are the standard deviation across training seeds, not the standard error of the mean. The difference between the arm means is 12.4, 7.1, 9.9 and 11.7 times the pooled between-seed SD for Pass@1 at $T=1$, Pass@1 at $T=4$, the confidence gap and Hit@1. The two arms separate completely: the worst CDLM seed exceeds the best MDLM seed by $+0.0436$ (Pass@1, $T=1$), $+0.0326$ (Pass@1, $T=4$), $+0.0520$ (gap) and $+0.0656$ (Hit@1), so all nine cross-arm pairs of seeds favor CDLM on each metric. Welch's $t$-test~\citep{welch1947generalization} gives $p$ between $0.003$ and $0.008$ on the four metrics. A two-sided exact Mann--Whitney test~\citep{mann1947test} with three runs per arm cannot return a $p$-value below $2/\binom{6}{3}=0.10$; it returns exactly $0.10$ on every metric, which is the value attained under complete separation and not a null result. Per cell, the seed-mean Pass@1 of CDLM exceeds that of MDLM in 11 of 12 cells at $T=1$ (the exception is MBPP+ literal, $-0.013$) and in 11 of 12 cells at $T=4$ (the exception is HumanEval+ identifier, $-0.063$).

\paragraph{Pipeline identity and GPU count.}
Evaluated with the frozen pipeline, the seed-42 CDLM checkpoint produces refined programs identical to the published run in every cell, and Pass@1 differs only through one HumanEval+ program whose test times out in one of the two executions ($0.1888$ against $0.1883$ at $T=1$; $0.2853$ against $0.2858$ at $T=4$). For MDLM, the seed-42 value here ($0.1386$ at $T=1$) differs slightly from the original evaluation of the same checkpoint ($0.137$), because that evaluation used MDLM's own CRB instance files, whose identifier and literal corruptions differ (Appendix~\ref{app:stats_exact}); on the operator cells, whose instances are identical, the two evaluations agree exactly. Seed 42 differs from the mean of the two 2-GPU seeds by at most $0.0015$ in Pass@1, in either arm and at either depth, against differences between the arm means of $0.042$ to $0.050$. Each arm contains one 4-GPU run and two 2-GPU runs, so an effect of the GPU count that is common to both arms enters the between-seed SD and not the difference of means.

\subsection{Scope}
\label{app:stats_scope}

Each arm was decoded once with greedy decoding, so the paired tests bound only the uncertainty from the finite sample of CRB problems; training-seed variation is measured only for the 0.5B comparison of CDLM with MDLM, while the clean-token ablation and the 8B (Section~\ref{sec:llada8b}) and 16B (Section~\ref{sec:when_correction_helps}) comparisons are single training runs per arm, and ParallelBench outcomes (Table~\ref{tab:parallelbench_final}) were retained only as per-task aggregates, so no paired test is possible there.

%% file: appendix/robustness.tex
\section{Robustness to the Corruption Distribution}
\label{app:robustness}

CRB replaces each corrupted token with a token of the same lexical category (for identifiers and literals, one that appears elsewhere in the same program), drawn uniformly at random, and mixture training replaces visible tokens uniformly at random. This section asks whether CDLM's advantage depends on that similarity. Appendix~\ref{app:plausible_crb} keeps CRB's construction but draws each corrupted token from the base model's own top predictions, which is the opposite end of the model's belief from uniform replacement; it gives the full results behind the paragraph in Section~\ref{sec:robustness}. Appendix~\ref{app:natural_errors} drops the corruption operator altogether and repairs programs that the models generated themselves and that fail their tests. Both studies use the three 0.5B arms of Section~\ref{sec:main_comparison}: the Open-dCoder-0.5B base model, the absorbing-only fine-tuned MDLM and \textsc{CDLM-0.5B}, all with the same refinement procedure as the main CRB experiments ($\tau=0.9$, greedy decoding, $T\in\{1,4\}$ rounds, that is, two and five forward passes).

\subsection{Plausible CRB}
\label{app:plausible_crb}

\paragraph{Construction.}
Positions are selected as in CRB with $n_{\mathrm{replace}}=1$. For each selected position, we replace the token with the mask token, keep the rest of the function body clean (the signature and docstring are not part of the proposer's input), and read the top-5 predictions of the shared base checkpoint (Open-dCoder-0.5B) at that position. We discard the original token, keep only single-token candidates whose decoded form has the same leading and trailing whitespace (so the tokenized length is unchanged) and that belong to the same lexical category as the original token (an operator from the CRB operator set, a non-keyword identifier, or a numeric literal), and substitute the highest-ranked remaining candidate. The base checkpoint is the common ancestor of all three arms, so the corruption favors none of them; masking the position asks which tokens the model expects there, rather than whether it would copy the token it sees. The category constraint is never relaxed: a position whose top-5 contains no admissible candidate of the right category is skipped. Over the 12 dataset $\times$ error-type files, 4{,}382 of 4{,}715 queried positions received a substitution and 333 were skipped. After removing duplicate programs (two variants are drawn per task and the substitution is deterministic), 3{,}200 plausible programs remain. As in CRB, corrupted programs that still pass their tests are discarded; 2{,}761 of them (86.3\%) fail and form the benchmark. Typical substitutions are \texttt{distance < threshold} $\to$ \texttt{distance >= threshold}, \texttt{\textquotesingle\textquotesingle.join(...)} $\to$ \texttt{\textquotesingle\textquotesingle.append(...)}, and \texttt{balance = 0} $\to$ \texttt{balance = 1} (HumanEval/0, /1 and /3).

\paragraph{Matched random control.}
The random arm was regenerated with the same generator, the same tokenizer (that of the base checkpoint), the same random seed and Python hash seed, and the same code path, so the two regimes use the same procedure and differ in how the replacement token is drawn. Because the random draw consumes random numbers that the top-$k$ draw does not, the corrupted positions are not the same (for example, they are identical in 58 of the 127 HumanEval operator tasks present in both regimes), and the task sets differ (random identifier and literal replacement needs another token of that category in the same body). 3{,}213 of the random arm's 3{,}568 programs (90.1\%) fail their tests; this control is a different sample from the pinned set of Appendix~\ref{app:release} even though both contain 3{,}213 failing programs, so CDLM's random-regime values here differ slightly from Table~\ref{tab:oci_05b_crb}. Within each regime, all three arms are evaluated on the same files. This differs from the original CRB files behind Table~\ref{tab:confidence_refinement_benchmark}, in which identifier and literal corruptions were drawn separately for each model (Appendix~\ref{app:stats_exact}).

\paragraph{How plausible the corruptions are.}
Table~\ref{tab:plausible_crb_shift} recomputes, on HumanEval, the base model's distribution at each corrupted position (position masked, rest of the function body clean, as in the construction) and locates the substituted token in it. For operators and identifiers the two regimes sit at opposite ends of the model's belief: the median rank falls from 20.5 to 2 and from 39 to 2, and the geometric-mean probability of the substituted token rises by factors of about 975 and 494. The median plausible rank is 2 rather than 1 because the original token, which usually holds rank 1, is excluded. Literal corruptions shift much less: 81.3\% of random literal substitutions are already in the base model's top 5 (median rank 3), because a random literal replacement is another numeric literal of the same program, and programs contain few, mostly small, literals. The literal cells are therefore a weak test of distribution shift, and we report the aggregate retention figures below both with and without them.

\begin{table}[!htbp]
\centering
\small
\setlength{\tabcolsep}{4pt}
\caption{How plausible the corruptions are, under the shared base checkpoint (Open-dCoder-0.5B) with the corrupted position masked. Each entry gives random / plausible. Position counts, ranks, top-5 fractions and probabilities are measured on HumanEval (up to 200 corrupted programs per file); rank 1 is the most probable token. Ratio: plausible over random geometric-mean probability. The last column counts the test-failing programs over all four datasets that form the benchmark. Measured with the function body as the only context, as in the construction; with the full program visible, the plausible token is in the top 5 for 87.5\% (operator), 75.5\% (identifier) and 98.0\% (literal) of positions.}
\label{tab:plausible_crb_shift}
\begin{tabular}{lcccccc}
\toprule
Error type & Positions & Median rank & In top 5 & Geo.\ mean prob. & Ratio & Programs \\
\midrule
Operator & 200 / 176 & 20.5 / 2 & 19.5\% / 100\% & $9.7\times10^{-6}$ / $9.5\times10^{-3}$ & 975.5 & 1{,}465 / 997 \\
Identifier & 131 / 98 & 39 / 2 & 14.5\% / 100\% & $7.0\times10^{-6}$ / $3.5\times10^{-3}$ & 494.5 & 934 / 728 \\
Literal & 155 / 200 & 3 / 2 & 81.3\% / 100\% & $5.4\times10^{-4}$ / $7.1\times10^{-3}$ & 13.2 & 814 / 1{,}036 \\
\bottomrule
\end{tabular}
\end{table}

\paragraph{Metrics.}
Confidence is $c_i=p_\theta(z_i\mid\bm{z})$, the probability the model assigns to the token currently at body position $i$, read at the first forward pass over the unmodified corrupted program; the confidence gap and Hit@$K$ therefore do not depend on $T$. The gap is computed per program (mean confidence over clean body positions minus confidence at the corrupted position) and then averaged over programs. Hit@$K$ (top-$K$ hit rate) is the fraction of programs whose corrupted position is among the $K$ lowest-confidence body positions. Pass@1 is measured after $T=1$ and $T=4$ rounds. Unless stated otherwise, every number is the macro average, the unweighted mean over the 12 dataset $\times$ error-type cells at $n_{\mathrm{replace}}=1$. The \emph{retained} fraction of a metric is the CDLM$-$MDLM difference under plausible corruption divided by the same difference under random corruption, both macro-averaged.

\begin{table}[!htbp]
\centering
\small
\setlength{\tabcolsep}{3.5pt}
\caption{Random versus plausible CRB ($n_{\mathrm{replace}}=1$, $\tau=0.9$). Every entry is the macro average over the 12 dataset $\times$ error-type cells; within a regime all three arms see the same corrupted programs (3{,}213 random, 2{,}761 plausible). Gap and Hit@$K$ are read at the first forward pass. $\Delta$: CDLM$-$MDLM, computed before rounding. Retained: plausible $\Delta$ divided by random $\Delta$. The literal cells, where random replacement was already close to plausible (Table~\ref{tab:plausible_crb_shift}), are included; without them the retained fractions are 52.5\% (gap), 48.6\% (Hit@1), 68.2\% (Hit@3), 85.9\% (Hit@6), 64.5\% (Pass@1, $T{=}1$) and 64.8\% (Pass@1, $T{=}4$). One training seed per arm and one corruption draw.}
\label{tab:plausible_crb}
\begin{tabular}{l cccc cccc c}
\toprule
& \multicolumn{4}{c}{Random CRB} & \multicolumn{4}{c}{Plausible CRB} & \\
\cmidrule(lr){2-5}\cmidrule(lr){6-9}
Metric & Base & MDLM & CDLM & $\Delta$ & Base & MDLM & CDLM & $\Delta$ & Retained \\
\midrule
Gap & 0.099 & 0.094 & \textbf{0.158} & $+$0.064 & 0.054 & 0.051 & \textbf{0.084} & $+$0.033 & 52.1\% \\
Hit@1 & 0.148 & 0.147 & \textbf{0.222} & $+$0.075 & 0.093 & 0.092 & \textbf{0.129} & $+$0.036 & 48.5\% \\
Hit@3 & 0.337 & 0.343 & \textbf{0.452} & $+$0.109 & 0.242 & 0.240 & \textbf{0.309} & $+$0.069 & 63.2\% \\
Hit@6 & 0.570 & 0.570 & \textbf{0.662} & $+$0.092 & 0.456 & 0.434 & \textbf{0.512} & $+$0.078 & 85.3\% \\
Pass@1, $T{=}1$ & 0.140 & 0.140 & \textbf{0.185} & $+$0.044 & 0.099 & 0.087 & \textbf{0.116} & $+$0.029 & 66.2\% \\
Pass@1, $T{=}4$ & 0.259 & 0.243 & \textbf{0.288} & $+$0.045 & 0.212 & 0.187 & \textbf{0.219} & $+$0.033 & 72.5\% \\
\bottomrule
\end{tabular}
\end{table}

\paragraph{Results.}
Table~\ref{tab:plausible_crb} gives the macro averages. Plausible corruptions are harder for every arm, including the untrained base model, whose gap falls from $0.099$ to $0.054$ and whose Pass@1 at $T=4$ falls from $0.259$ to $0.212$. CDLM keeps the highest value on every metric in both regimes, but its margin over MDLM shrinks. Of the random-regime margin, 85.3\% is retained at Hit@6, 66.2\% and 72.5\% for Pass@1 at $T=1$ and $T=4$, 63.2\% at Hit@3, 52.1\% for the gap and 48.5\% at Hit@1. Roughly half of the localization margin at small $K$ is therefore specific to the random corruption distribution, while most of the Hit@6 and Pass@1 margins carry over.

Per cell, CDLM exceeds compute-matched MDLM in all 12 cells on the confidence gap, Hit@1 and Hit@6 under both regimes, and in all 12 on Pass@1 at $T=4$. At $T=1$ it leads in 11 of 12 cells under random corruption and in 10 of 12 under plausible corruption (one of the two remaining cells is a tie). Hit@3 is 12 of 12 under random and 11 of 12 under plausible corruption. Every exception is an MBPP+ cell: random literal and plausible operator at $T=1$ (the tie is plausible identifier), and plausible literal at Hit@3.

Relative to the base model, the absorbing-only arm is not beneficial for in-place repair in either regime. Averaged over cells, MDLM is above base on three of the six random-regime entries of Table~\ref{tab:plausible_crb}, by at most $0.006$ (Hit@3; $+0.001$ at Hit@6 and less than $0.001$ for Pass@1 at $T=1$), and below base on all six plausible-regime entries. On Pass@1 it never exceeds base by more than $0.001$ and is below base in three of the four combinations of regime and depth. CDLM is above base on all twelve macro entries. This holds in aggregate rather than per cell: CDLM exceeds base in 8 to 12 of the 12 cells, depending on metric and regime, and the smallest count is for Pass@1 at $T=4$ under plausible corruption.

\begin{table}[!htbp]
\centering
\small
\setlength{\tabcolsep}{4pt}
\caption{Plausible CRB by error type. Each entry is the unweighted mean over the four datasets ($n_{\mathrm{replace}}=1$, $\tau=0.9$); Rand.\ = random CRB, Plaus.\ = plausible CRB. \emph{Retained}: CDLM$-$MDLM under plausible corruption divided by CDLM$-$MDLM under random corruption. For literal corruptions, random replacement was already close to plausible (Table~\ref{tab:plausible_crb_shift}), and the random-regime Pass@1 margin at $T=1$ is only $+0.013$, so the literal retained fractions carry little information.}
\label{tab:plausible_crb_types}
\begin{tabular}{ll cc cc cc cc}
\toprule
& & \multicolumn{2}{c}{Gap} & \multicolumn{2}{c}{Hit@1} & \multicolumn{2}{c}{Pass@1, $T{=}1$} & \multicolumn{2}{c}{Pass@1, $T{=}4$} \\
\cmidrule(lr){3-4}\cmidrule(lr){5-6}\cmidrule(lr){7-8}\cmidrule(lr){9-10}
Type & Arm & Rand. & Plaus. & Rand. & Plaus. & Rand. & Plaus. & Rand. & Plaus. \\
\midrule
Operator & Base & 0.086 & 0.034 & 0.140 & 0.083 & 0.143 & 0.087 & 0.280 & 0.194 \\
 & MDLM & 0.080 & 0.029 & 0.137 & 0.076 & 0.144 & 0.075 & 0.264 & 0.168 \\
 & CDLM & 0.146 & 0.049 & 0.221 & 0.101 & 0.181 & 0.085 & 0.307 & 0.187 \\
 & \emph{retained} & \multicolumn{2}{c}{31.1\%} & \multicolumn{2}{c}{30.0\%} & \multicolumn{2}{c}{26.7\%} & \multicolumn{2}{c}{45.4\%} \\
\midrule
Identifier & Base & 0.134 & 0.111 & 0.188 & 0.148 & 0.164 & 0.176 & 0.279 & 0.267 \\
 & MDLM & 0.124 & 0.108 & 0.187 & 0.150 & 0.159 & 0.153 & 0.279 & 0.262 \\
 & CDLM & 0.232 & 0.178 & 0.296 & 0.218 & 0.242 & 0.220 & 0.327 & 0.301 \\
 & \emph{retained} & \multicolumn{2}{c}{65.5\%} & \multicolumn{2}{c}{62.9\%} & \multicolumn{2}{c}{81.7\%} & \multicolumn{2}{c}{82.2\%} \\
\midrule
Literal & Base & 0.079 & 0.017 & 0.115 & 0.048 & 0.113 & 0.034 & 0.217 & 0.173 \\
 & MDLM & 0.078 & 0.015 & 0.119 & 0.052 & 0.118 & 0.033 & 0.187 & 0.131 \\
 & CDLM & 0.097 & 0.024 & 0.150 & 0.067 & 0.131 & 0.043 & 0.231 & 0.170 \\
 & \emph{retained} & \multicolumn{2}{c}{48.4\%} & \multicolumn{2}{c}{47.7\%} & \multicolumn{2}{c}{82.7\%} & \multicolumn{2}{c}{88.5\%} \\
\bottomrule
\end{tabular}
\end{table}

\paragraph{By error type.}
Table~\ref{tab:plausible_crb_types} splits the comparison by error type. The attenuation of the localization margin is concentrated in operator corruptions, and identifier corruptions retain the most: operator retains 31.1\% of the gap margin and 30.0\% of the Hit@1 margin, identifier 65.5\% and 62.9\%. (Pooled over programs instead of averaged over datasets, the same fractions are 35.0\% and 36.7\% for operator, 70.3\% and 76.1\% for identifier.) At Hit@6 all three types retain between 83.8\% and 88.3\%. A plausible operator swap such as \texttt{<} $\to$ \texttt{>=} can yield idiomatic code that only the docstring shows to be wrong, and every arm degrades on it: the base model's own operator gap falls from $0.086$ to $0.034$. On plausible operator corruptions CDLM's Pass@1 is below the base model's at both depths ($0.085$ against $0.087$ at $T=1$, $0.187$ against $0.194$ at $T=4$), while remaining above MDLM; on plausible literal corruptions it is below base at $T=4$ ($0.170$ against $0.173$). Under plausible corruption, CDLM's aggregate Pass@1 lead over the base model therefore comes mainly from identifier corruptions.

Whether retention follows the size of the distribution shift depends on how the shift is measured (Table~\ref{tab:plausible_crb_shift}). By rank, identifier corruptions move furthest (median 39 to 2) and retain the most of the gap and Hit@1 margins, which is the opposite of what a detector of the training noise alone would predict. By probability ratio, operator corruptions move furthest (about 975 against 494) and retain the least of these margins, which is consistent with such a detector. We therefore do not use the split by error type as evidence either way. The aggregate result is what we rely on: under corruptions drawn from the opposite end of the model's belief, CDLM keeps the highest value on every metric and retains 48.5--85.3\% of its margin over MDLM, so its advantage is not only a detector of its own training noise; but about half of its localization margin at small $K$ on the original CRB depends on the corrupted token being implausible.

\paragraph{Relation to Table~\ref{tab:confidence_refinement_benchmark}.}
The Base row of the random regime differs slightly from the Open-dCoder-0.5B row of Table~\ref{tab:confidence_refinement_benchmark} at $n_{\mathrm{replace}}=1$ (gap $0.099$ against $0.106$, Pass@1 $0.140$ against $0.143$ at $T=1$ and $0.259$ against $0.251$ at $T=4$) for two reasons. First, the instances differ: the regenerated control shares all operator instances with the original files, but per cell only 31--46\% of the identifier instances and 58--91\% of the literal instances are byte-identical. Second, the gap is averaged per program here, whereas Table~\ref{tab:confidence_refinement_benchmark} pools all body positions of a cell before differencing; on the original files the per-program gap of the base model is $0.100$ against $0.106$ pooled. Run on the original files, the scripts used here reproduce the base model's Hit@1 ($0.147$), top-5 hit rate ($0.509$) and Pass@1 at $T=1$ ($0.143$) and $T=4$ ($0.251$) in Table~\ref{tab:confidence_refinement_benchmark}.

\paragraph{Not run.}
Plausible CRB was run at $n_{\mathrm{replace}}=1$ only; $n_{\mathrm{replace}}=3$ was not run because of the compute budget. In the top-$k$ construction it would consist of three independent single-token substitutions, which differs from random CRB at $n_{\mathrm{replace}}=3$, where one multi-token identifier can account for all three positions. A stricter adversary, which would pick the token that maximizes $p_\theta(z_i\mid\bm{z})$ with the position visible (the exact quantity the remasking rule thresholds), is implemented in our generator but was not run. All results use one training seed per arm and one draw of corruption positions; the substitution itself is deterministic given the position.

\subsection{Naturally Occurring Errors}
\label{app:natural_errors}

This study repairs programs that the models generated themselves and that fail their tests, with no corruption operator involved. It is the complement of Appendix~\ref{sec:self_revision_minimal}, which applies one CRB corruption to programs that each model generated \emph{correctly}; here the input is the model's own \emph{failing} program, unchanged.

\paragraph{Setup.}
For every task, the generator receives the CRB function header (signature and docstring) with the whole body masked and fills it with a single greedy sample from its masked-diffusion decoder (128 tokens in 128 steps, confidence-based remasking). Each body is cut at the first markdown fence and at the first of the lm-evaluation-harness stop sequences \texttt{\textbackslash nclass}, \texttt{\textbackslash ndef}, \texttt{\textbackslash n\#}, \texttt{\textbackslash nif} and \texttt{\textbackslash nprint} (Appendix~\ref{app:scratch_eval}), and then executed against the tests; the failing programs form the benchmark. We built two failing sets. The base-generated set has 940 programs (136 HumanEval, 136 HumanEval+, 404 MBPP, 264 MBPP+). The CDLM-generated set has 865 (114, 120, 380 and 251); 56 CDLM generations whose body was empty after truncation were dropped, and none of the base model's. The base and CDLM arms thus each repair one set they generated and one they did not, and MDLM repairs two sets it did not generate. All three arms repair both sets with the refinement procedure of the main CRB experiments at $T=1$ and $T=4$, and the base-generated set also at $T=16$.

\paragraph{Metric.}
We report Pass@1 only. Confidence gap and Hit@$K$ need a ground-truth error position, which a natural failure does not have. Aligning a failing program to the canonical solution would not supply one: the canonical solution is one correct program among many, and the alignment marks most of the body (Table~\ref{tab:natural_distance}).

\begin{table}[!htbp]
\centering
\small
\setlength{\tabcolsep}{4pt}
\caption{In-place repair of naturally occurring failures (Open-dCoder-0.5B arms, $\tau=0.9$). Pass@1 is pooled over the programs of the four datasets; the number of repaired programs is in parentheses. Modified: fraction of programs whose output differs from the input after $T$ rounds. $T=16$ was run on the base-generated set only.}
\label{tab:natural_errors}
\begin{tabular}{ll ccc cc}
\toprule
& & \multicolumn{3}{c}{Pass@1} & \multicolumn{2}{c}{Modified} \\
\cmidrule(lr){3-5}\cmidrule(lr){6-7}
Failing set & Arm & $T{=}1$ & $T{=}4$ & $T{=}16$ & $T{=}1$ & $T{=}4$ \\
\midrule
Base-generated ($n=940$) & Base & 0.000 (0) & 0.000 (0) & 0.000 (0) & 20.6\% & 28.2\% \\
 & MDLM & 0.001 (1) & 0.001 (1) & 0.001 (1) & 12.4\% & 20.3\% \\
 & CDLM & 0.001 (1) & 0.001 (1) & 0.001 (1) & 16.0\% & 19.8\% \\
\midrule
CDLM-generated ($n=865$) & Base & 0.001 (1) & 0.003 (3) & n/a & 23.6\% & 30.1\% \\
 & MDLM & 0.002 (2) & 0.001 (1) & n/a & 15.5\% & 22.2\% \\
 & CDLM & 0.002 (2) & 0.002 (2) & n/a & 14.7\% & 19.9\% \\
\bottomrule
\end{tabular}
\end{table}

\paragraph{Results.}
Table~\ref{tab:natural_errors} shows that every arm is at the floor: pooled Pass@1 lies between $0.000$ and $0.003$ for all three arms on both failing sets and at every depth. The differences between arms are one or two programs and carry no information about the arms.

\paragraph{Why the floor.}
Table~\ref{tab:natural_distance} gives the token-level Levenshtein distance from each failing body to the canonical body (Open-dCoder tokenizer). A CRB instance at $n_{\mathrm{replace}}=1$ is one token from the canonical program: all 5{,}974 instances of Appendix~\ref{app:plausible_crb} are at distance exactly 1. A naturally occurring failure is not. On HumanEval the median base-generated failure is 109 token edits from the canonical body, with failing bodies 113 tokens long on average; on MBPP the median is 41.5. Only 4 of the 940 base-generated failures (0.4\%) are within four tokens of the canonical body, and 12 (1.3\%) within eight. The canonical solution is only one correct program, so these distances overstate the repair a failing program needs whenever it is close to a different correct solution; the contrast with CRB is nevertheless large. These failures are mostly not a correct program with a few wrong tokens, and a confidence-gated edit of a few positions is unlikely to turn them into one. The distances are for the 940 programs that were repaired. A count on an earlier version of the base-generated set (996 programs, truncated with a narrower stop list) gave a median of 52.5 tokens on MBPP and 3 programs (0.3\%) within four tokens; the HumanEval median (109) is the same in both.

\begin{table}[!htbp]
\centering
\small
\caption{Token-level Levenshtein distance from the failing (or corrupted) body to the canonical body, Open-dCoder tokenizer. Quartiles by linear interpolation. Within 4: programs at distance at most 4 tokens. CRB rows: all test-failing $n_{\mathrm{replace}}=1$ instances of the plausible-CRB study, over the four datasets and three error types.}
\label{tab:natural_distance}
\begin{tabular}{lrrcrr}
\toprule
Set & Programs & Median & Quartiles & Within 4 & Mean length \\
\midrule
Base-generated, HumanEval & 136 & 109 & 94--116 & 0 (0.0\%) & 113 \\
Base-generated, HumanEval+ & 136 & 110 & 96--117 & 0 (0.0\%) & 112 \\
Base-generated, MBPP & 404 & 41.5 & 26--97 & 0 (0.0\%) & 61 \\
Base-generated, MBPP+ & 264 & 34 & 23--71 & 4 (1.5\%) & 55 \\
Base-generated, all & 940 & 59 & 29--110 & 4 (0.4\%) & 74 \\
CDLM-generated, all & 865 & 55 & 29--104 & 5 (0.6\%) & 68 \\
\midrule
CRB, random & 3{,}213 & 1 & 1--1 & 3{,}213 (100\%) & 56 \\
CRB, plausible & 2{,}761 & 1 & 1--1 & 2{,}761 (100\%) & 57 \\
\bottomrule
\end{tabular}
\end{table}

\paragraph{Controls.}
Four checks indicate that the floor is a property of the inputs rather than of the pipeline. (i) The identical command line, run with the base arm on the 230 HumanEval operator instances of random CRB, gives Pass@1 $0.174$ at $T=1$ and $0.313$ at $T=4$. (ii) Raising the budget to $T=16$ (17 forward passes) on the base-generated set leaves every arm's Pass@1 unchanged (0, 1 and 1 programs). (iii) The remasking rule does act: the first round remasks at least one body position in 21.9--30.6\% of the base-generated programs, depending on the arm, and the refiner changes 12.4--28.2\% of them, depending on arm and depth (Table~\ref{tab:natural_errors}); the edits almost never turn a failing program into a passing one. (iv) Restricting to failures that parse as Python (880 of 940 base-generated, 792 of 865 CDLM-generated) changes nothing: every arm repairs zero or one program at either depth.

\paragraph{Reading.}
This probe is uninformative about error awareness, and we do not read it either way: it neither supports nor contradicts the comparison of CDLM and MDLM on CRB. What it measures is a scope boundary. Confidence-gated in-place refinement addresses localized errors in an otherwise correct program; end-to-end generation failure of a 0.5B model is a different regime, in which the failing program is far from the canonical solution. This is the same boundary that puts insertions and deletions out of reach of a fixed-length remask-then-repredict decoder (Appendix~\ref{app:limitations}). Testing error awareness on natural errors requires failures that are close to a correct program, which at this scale are rare (1.3\% of base-generated failures within eight tokens); a stronger generator or a much larger pool of generations would be needed.

%% file: appendix/llada8b.tex
\providecommand{\mdfence}{\texttt{\textasciigrave\textasciigrave\textasciigrave}}

\section{Transfer to LLaDA-8B-Base}
\label{app:llada8b}

This appendix documents the LLaDA-8B-Base comparison of Section~\ref{sec:llada8b} (Table~\ref{tab:llada8b}): the training and evaluation protocol (Appendix~\ref{app:llada8b_setup}), the two Pass@1 conventions and the code-fence audit behind them (Appendix~\ref{app:llada8b_fence}), the confidence analysis and localization over all four datasets (Appendix~\ref{app:llada8b_conf}), the behavior of the mixture arm at $T=4$ on HumanEval+ (Appendix~\ref{app:llada8b_t4}), and a from-scratch generation check (Appendix~\ref{app:llada8b_guard}).
LLaDA-8B-Base~\citep{nie2025largelanguagediffusionmodels} is pretrained from scratch as a masked diffusion model, whereas Dream-7B-Base and Open-dCoder-0.5B are adapted from autoregressive checkpoints~\citep{ye2025dream7bdiffusionlarge, opendllm2025}.
As at 0.5B there are three arms: the base model, an absorbing-only arm (MDLM) and a mixture arm (CDLM); the two fine-tuned arms differ only in the training objective.
Each fine-tuned arm is a single training run (seed 42), and every evaluation is a single deterministic decoding run.

\subsection{Training and evaluation setup}
\label{app:llada8b_setup}

\paragraph{Adapters and corpus.}
Both arms train a LoRA adapter~\citep{hu2022lora} on \texttt{GSAI-ML/LLaDA-8B-Base} (Hugging Face revision \texttt{0f2787f2}, 8.02B parameters, bidirectional attention, mask token id 126336).
The adapter has rank 64, $\alpha=128$ (scaling $\alpha/r=2$) and dropout 0.05, and is placed on the seven linear projections inside each of the 32 transformer blocks (query, key, value, attention output, gate, up and down projections).
The token embeddings and the output layer are frozen.
The adapter has 167.77M trainable parameters, 2.05\% of the 8.18B parameters of the adapted model.
The corpus is Nemotron-SFT-Code~\citep{nvidia2025nvidianemotronnano2}, the coding corpus of the 0.5B runs (78 parquet shards).
Each document is tokenized, an end-of-text token is appended, and the result is split into chunks of at most 4096 tokens (chunks shorter than 16 tokens are dropped); documents are not packed together.

\paragraph{What a run reads.}
The trainer shuffles the shard files with seed 42 and gives rank $r\in\{0,1,2,3\}$ the files $r, r+4, r+8, \dots$, each read from its first row.
Its data loader runs two workers over a stream that is not split between them, so every sequence is delivered twice in a row, each time with an independent corruption draw.
A 2000-step run (24{,}000 sequences) therefore reads 12{,}000 distinct documents, the first 3{,}000 rows of four shards, with 3.55M tokens in total (296 per document on average, none longer than 4096 tokens), and sees each document twice.
The 200-step runs of Appendix~\ref{app:llada8b_conf} read the first 300 rows of the same four shards: 1{,}200 documents with 0.37M tokens.
These counts come from replaying the data stream of the reported runs with the trainer's own tokenizer and chunking.
Both arms read the same documents in the same order.

\paragraph{Optimization.}
Table~\ref{tab:llada8b_setup} lists the configuration.
The learning rate warms up linearly for 200 steps and then follows a cosine decay over a 20{,}000-step horizon; training stops at step 2000, where the learning rate is $2.94\times10^{-4}$, 98.0\% of the peak.
The 200-step runs are separate runs with the same configuration; they stop at the end of warmup, so their schedule is a prefix of the 2000-step schedule.
This schedule belongs to the 8B experiment and is not a replication of the 0.5B learning-rate schedule (Appendix~\ref{app:lr_horizon}); it is identical in both 8B arms.
The two arms differ only in the mixture probability and the noise weight: MDLM uses $\alpha=0$ and $\lambda_{\mathrm{noise}}=0$, CDLM uses $\alpha=0.1$ and $\lambda_{\mathrm{noise}}=0.1$, and both use clean-token weight 0.

\begin{table}[!htbp]
\centering
\small
\caption{LLaDA-8B-Base post-training configuration, identical in both arms except for the last two rows.}
\label{tab:llada8b_setup}
\begin{tabular}{lp{0.6\linewidth}}
\toprule
\textbf{Setting} & \textbf{Value} \\
\midrule
Base model & \texttt{GSAI-ML/LLaDA-8B-Base}, revision \texttt{0f2787f2} \\
Adapter & LoRA, rank 64, $\alpha=128$ (scaling 2), dropout 0.05, on the 7 linear projections of each block \\
Frozen & all base weights, including embeddings and output layer \\
Trainable parameters & 167.77M (2.05\%) \\
Corpus & Nemotron-SFT-Code: 12{,}000 documents, 3.55M tokens, each document seen twice \\
Sequence length & at most 4096 tokens, one document per sequence \\
Batch & 12 sequences (micro-batch 1, accumulation 3, 4 GPUs) \\
Steps & 2000 (onset analysis: separate 200-step runs) \\
Optimizer & AdamW~\citep{loshchilov2019adamw}, $\beta=(0.9,0.95)$, $\epsilon=10^{-8}$ \\
Weight decay, clipping & 0.01, gradient norm 1.0 \\
Learning rate & peak $3\times10^{-4}$, 200 warmup steps, cosine over 20{,}000 steps \\
Precision & bf16 base weights, fp32 adapter and optimizer states \\
Seed & 42 \\
Hardware & 4 A100-PCIe-40GB, about 1.4\,s per step, 46 min per arm \\
\midrule
Mixture probability $\alpha$ & MDLM 0, CDLM 0.1 \\
Noise weight $\lambda_{\mathrm{noise}}$ & MDLM 0, CDLM 0.1 \\
\bottomrule
\end{tabular}
\end{table}

\paragraph{Objective as implemented.}
For each sequence the trainer draws $t\sim\mathrm{Uniform}[0,1]$, sets $p=(1-10^{-3})\,t+10^{-3}$, and masks every position independently with probability $p$, which gives the masked set $\mathcal{M}$.
In the CDLM arm every remaining visible position is then replaced, independently with probability $\alpha$, by a token drawn uniformly from the vocabulary without the mask token and without the original token, which gives the replaced set $\mathcal{N}$.
With $\ell_i=-\log p_\theta(x_i\mid\bm{z})$ (LLaDA is bidirectional, so logits are not shifted), the loss of a micro-batch of $B$ sequences of length $L$ is
\[
\mathcal{L}_{\mathrm{8B}}
= \frac{1}{BL}\sum_{i\in\mathcal{M}}\frac{\ell_i}{p}
\;+\;
\lambda_{\mathrm{noise}}\,\frac{1}{|\mathcal{N}|}\sum_{i\in\mathcal{N}}\ell_i ,
\]
where the first sum runs over the masked positions of all $B$ sequences, each divided by the masking probability of its own sequence, and the second term is dropped when $\mathcal{N}$ is empty.
Clean visible positions receive no loss.
The first term is LLaDA's masked-diffusion loss; the second is the corruption term of Equation~\eqref{eq:mixture-loss}, normalized by the number of replaced positions.
The masking draws and the replacement draws come from separate random streams, so both arms see identical masking patterns.
The 0.5B runs implement the masked term differently (Appendix~\ref{app:objective_impl}), so the 0.5B and 8B objectives have the same structure but are not numerically the same loss.

\paragraph{Repair protocol.}
The repair inputs are the LLaDA-tokenized CRB files with $n_{\text{replace}}=1$ for the four datasets and three error types, restricted as everywhere in CRB to programs that fail their tests: 3{,}184 programs (HumanEval 224/131/142, HumanEval+ 255/103/153, MBPP 568/466/339 and MBPP+ 406/220/177 for operator/identifier/literal errors).
Refinement uses the procedure of Section~\ref{bg:refinement} (Algorithm~\ref{alg:confidence_refinement}), as in Table~\ref{tab:confidence_refinement_benchmark}: the function head is fixed and every body token is editable, positions with confidence at most $\tau=0.9$ are remasked and re-predicted greedily (temperature 0), with batch size 1 and $T\in\{1,4\}$ rounds, that is, 2 and 5 forward passes.
The adapter is merged into the bf16 base weights before refinement.
A program counts as repaired if it passes all tests of its dataset.
Pass@1 is reported per (dataset, error type) cell and as the macro, the unweighted mean over the 12 cells.
\emph{Raw} scores execute the refined program as produced (the protocol of Table~\ref{tab:confidence_refinement_benchmark}); \emph{de-fenced} scores first apply the rule of Appendix~\ref{app:llada8b_fence}.
All three arms were refined and scored with one frozen snapshot of the evaluation code.

\paragraph{Localization protocol.}
Localization is measured on the three HumanEval localization files with $n_{\text{replace}}=1$ (operator 250, identifier 136, literal 155 records; $n=541$).
One forward pass is run on the function head followed by the corrupted body, with no mask token in the input, and the confidence of each body position is $c_i=p_\theta(z_i\mid\bm{z})$, the probability of the token currently at position $i$, which is the quantity the remasking rule thresholds.
For each record the confidence gap is the mean confidence over clean body positions minus the confidence at the corrupted position, and Hit@$K$ (top-$K$ hit rate) records whether the corrupted position is among the $K$ lowest-confidence body positions.
Both are averaged over records, pooled over the three error types.
Appendix~\ref{app:llada8b_conf} extends the same protocol to all four datasets.

\paragraph{Localization records and repair inputs.}
The localization files and the repair inputs do not contain the same programs.
The repair inputs are the test-failing records of the evaluated CRB files.
The 250 operator localization records are exactly the records of the operator evaluated file, and the 224 of them that fail their tests are the operator repair inputs.
The identifier and literal localization files were regenerated after the evaluated files had been built, with a generator whose choice of replacement depended on Python's string-hash seed (Appendix~\ref{app:release}); only 46 of the 136 identifier and 89 of the 155 literal localization records coincide with records of the evaluated files.
In total, 351 of the 541 localization records are repair instances; the other 190 are 156 regenerated identifier or literal records and 34 records whose corruption does not make the program fail its tests.
All localization numbers in Table~\ref{tab:llada8b} and in this appendix are measured on the 541 localization records.
As a consistency check we recomputed the localization metrics, with the same released code and adapters, on exactly the 497 programs that the Pass@1 runs repaired (224 operator, 131 identifier, 142 literal). The conclusions are unchanged: the gap is $0.252$ (base), $0.404$ (MDLM) and $0.790$ (CDLM), and Hit@1 is $0.175$, $0.143$ and $0.612$ (Hit@5: $0.781$, $0.875$, $0.982$). CDLM's Hit@1 exceeds MDLM's (exact McNemar $p=1.5\times10^{-52}$) and base's ($p=2.0\times10^{-42}$), while MDLM and base are not significantly different ($p=0.16$); this last comparison also varies in sign across error types and subsets, so we read it only as ``no improvement''.

\subsection{Raw and de-fenced Pass@1}
\label{app:llada8b_fence}

\paragraph{The fence artifact.}
Both fine-tuned arms learn markdown code fences (\mdfence) from the fine-tuning corpus; the base model never emits one.
Under raw scoring a fence anywhere in a refined program is a syntax error.
The effect concentrates on HumanEval+ because of how its CRB bodies are formatted: all 511 HumanEval+ bodies begin with a blank line and end with two newlines, against 4 and 20 of the 497 HumanEval bodies and none of the MBPP and MBPP+ bodies.
Since every body token is editable, these whitespace positions are editable as well, and the fine-tuned arms fill them with a fence.
Table~\ref{tab:llada8b_fence} counts refined programs that contain a fence: none of the 3{,}184 base programs at either depth, against 78.7\% (MDLM) and 94.9\% (CDLM) of HumanEval+ programs at $T=1$, and at most 8.9\% at $T=1$ and 10.1\% at $T=4$ on the other three datasets.

\begin{table}[!htbp]
\centering
\small
\caption{Percentage of refined programs that contain a code fence (\mdfence), per dataset, pooled over the three error types ($n_{\text{replace}}=1$; 497, 511, 1{,}373 and 803 programs for HumanEval, HumanEval+, MBPP and MBPP+).}
\label{tab:llada8b_fence}
\begin{tabular}{l cccc cccc}
\toprule
& \multicolumn{4}{c}{$T=1$} & \multicolumn{4}{c}{$T=4$} \\
\cmidrule(lr){2-5}\cmidrule(lr){6-9}
\textbf{Arm} & HE & HE+ & MBPP & MBPP+ & HE & HE+ & MBPP & MBPP+ \\
\midrule
Base & 0.0 & 0.0 & 0.0 & 0.0 & 0.0 & 0.0 & 0.0 & 0.0 \\
MDLM & 5.6 & 78.7 & 0.4 & 0.5 & 9.9 & 88.5 & 1.5 & 1.6 \\
CDLM & 8.9 & 94.9 & 1.6 & 1.1 & 10.1 & 96.7 & 5.3 & 4.4 \\
\bottomrule
\end{tabular}
\end{table}

\paragraph{The de-fence rule.}
The rule truncates each refined program at the first occurrence of the substring \mdfence\ and strips trailing whitespace.
It is applied identically to all three arms.
(An earlier variant that truncated at the first line starting with a fence missed fences written inside a line; every number here uses the substring rule.)
The rule is justified by the fence counts, not by the scores it produces: fences occur only in the fine-tuned arms, and the rule removes no code from any base program.
None of the 6{,}368 base programs (24 cells) contains a fence, so on base the rule only strips trailing whitespace, which does not change the executed program, and the base macro is 0.551 at $T=1$ and 0.595 at $T=4$ under both conventions.
Three base cells nevertheless differ by one program between the raw and the de-fenced evaluation (Table~\ref{tab:llada8b_cells}); in all three the program text is the same in both evaluations and the outcome flips on HumanEval/139, a HumanEval+ problem whose tests time out intermittently.
On byte-identical programs this moved the base HumanEval+ operator cell from 0.616 to 0.612 (one program of 255), so cell differences of this size are evaluation noise.
Table~\ref{tab:confidence_refinement_benchmark} reports 0.552 at $T=1$ and 0.596 at $T=4$ for the same inputs; its cells agree with the base run here except in the HumanEval+ operator and literal cells, by at most the number of HumanEval/139 records they contain.

\paragraph{Results.}
Table~\ref{tab:llada8b_cells} gives every cell under both conventions.
Under raw scoring, MDLM falls below base (0.475 against 0.551 at $T=1$) and CDLM is at 0.530.
This deficit comes from the three HumanEval+ cells, where only 0.117 to 0.131 of MDLM's and 0.016 to 0.029 of CDLM's programs pass raw, and it is the fence artifact: on the other nine cells the raw 9-cell mean at $T=1$ is 0.533 (base), 0.592 (MDLM) and 0.699 (CDLM).
De-fenced, the macro at $T=1$ is 0.551 (base), 0.600 (MDLM) and 0.725 (CDLM).
At $T=1$, CDLM is the best of the three arms in all eight operator and identifier cells, ties base on HumanEval literal, and is below base in the other three literal cells.
Paired over the 3{,}184 programs (Table~\ref{tab:llada8b_paired}; pooled differences weight cells by size, unlike the macro), CDLM repairs more programs than MDLM under both conventions and at both depths, with exact McNemar $p\le1.1\times10^{-4}$ (Table~\ref{tab:llada8b_paired}; test as in Appendix~\ref{app:stats}).
MDLM repairs more programs than base under the de-fenced convention ($+0.062$ at $T=1$ and $+0.026$ at $T=4$).
We therefore read the raw MDLM deficit as a formatting effect and draw no conclusion from it about absorbing-only post-training.
The raw convention gives CDLM no advantage over base at $T=4$ ($-0.005$, $p=0.68$); Appendix~\ref{app:llada8b_t4} examines that depth.

\begin{table}[!htbp]
\centering
\small
\setlength{\tabcolsep}{2.2pt}
\caption{CRB Pass@1 of the three LLaDA-8B arms per (dataset, error type) cell ($n_{\text{replace}}=1$, $\tau=0.9$), raw and de-fenced. op, id and lit denote operator, identifier and literal errors; Macro is the unweighted mean over the 12 cells.}
\label{tab:llada8b_cells}
\begin{tabular}{ll ccc ccc ccc ccc c}
\toprule
& & \multicolumn{3}{c}{HumanEval} & \multicolumn{3}{c}{HumanEval+} & \multicolumn{3}{c}{MBPP} & \multicolumn{3}{c}{MBPP+} & \\
\cmidrule(lr){3-5}\cmidrule(lr){6-8}\cmidrule(lr){9-11}\cmidrule(lr){12-14}
$T$ & Arm & op & id & lit & op & id & lit & op & id & lit & op & id & lit & Macro \\
\midrule
\multicolumn{15}{l}{\textit{Raw}} \\
1 & Base & 0.621 & 0.351 & 0.775 & 0.616 & 0.379 & 0.824 & 0.533 & 0.307 & 0.646 & 0.515 & 0.314 & 0.734 & 0.551 \\
1 & MDLM & 0.638 & 0.527 & 0.697 & 0.129 & 0.117 & 0.131 & 0.606 & 0.442 & 0.655 & 0.640 & 0.473 & 0.650 & 0.475 \\
1 & CDLM & 0.763 & 0.756 & 0.704 & 0.016 & 0.029 & 0.026 & 0.708 & 0.609 & 0.622 & 0.749 & 0.714 & 0.667 & 0.530 \\
4 & Base & 0.674 & 0.412 & 0.817 & 0.682 & 0.437 & 0.843 & 0.558 & 0.333 & 0.673 & 0.596 & 0.368 & 0.746 & 0.595 \\
4 & MDLM & 0.603 & 0.519 & 0.683 & 0.082 & 0.068 & 0.065 & 0.602 & 0.446 & 0.637 & 0.645 & 0.505 & 0.667 & 0.460 \\
4 & CDLM & 0.772 & 0.710 & 0.704 & 0.012 & 0.019 & 0.007 & 0.680 & 0.564 & 0.605 & 0.751 & 0.705 & 0.689 & 0.518 \\
\midrule
\multicolumn{15}{l}{\textit{De-fenced}} \\
1 & Base & 0.621 & 0.351 & 0.775 & 0.612 & 0.379 & 0.830 & 0.533 & 0.307 & 0.646 & 0.515 & 0.314 & 0.734 & 0.551 \\
1 & MDLM & 0.665 & 0.542 & 0.732 & 0.604 & 0.485 & 0.706 & 0.606 & 0.442 & 0.655 & 0.640 & 0.473 & 0.655 & 0.600 \\
1 & CDLM & 0.817 & 0.802 & 0.775 & 0.737 & 0.777 & 0.712 & 0.708 & 0.616 & 0.622 & 0.749 & 0.714 & 0.667 & 0.725 \\
4 & Base & 0.674 & 0.412 & 0.817 & 0.686 & 0.437 & 0.843 & 0.558 & 0.333 & 0.673 & 0.596 & 0.368 & 0.746 & 0.595 \\
4 & MDLM & 0.661 & 0.534 & 0.739 & 0.639 & 0.495 & 0.732 & 0.602 & 0.446 & 0.637 & 0.645 & 0.505 & 0.672 & 0.609 \\
4 & CDLM & 0.821 & 0.756 & 0.761 & 0.384 & 0.388 & 0.353 & 0.681 & 0.582 & 0.605 & 0.751 & 0.709 & 0.689 & 0.623 \\
\bottomrule
\end{tabular}
\end{table}

\begin{table}[!htbp]
\centering
\small
\caption{Paired comparisons of the LLaDA-8B arms over the 3{,}184 CRB programs ($n_{\text{replace}}=1$): difference in the fraction of programs repaired (pooled over cells) and exact McNemar $p$-value.}
\label{tab:llada8b_paired}
\begin{tabular}{lc cc cc}
\toprule
& & \multicolumn{2}{c}{De-fenced} & \multicolumn{2}{c}{Raw} \\
\cmidrule(lr){3-4}\cmidrule(lr){5-6}
\textbf{Comparison} & $T$ & Difference & $p$ & Difference & $p$ \\
\midrule
CDLM $-$ MDLM & 1 & $+0.115$ & $1.8\times10^{-32}$ & $+0.072$ & $4.6\times10^{-15}$ \\
CDLM $-$ MDLM & 4 & $+0.038$ & $1.1\times10^{-4}$ & $+0.067$ & $3.4\times10^{-14}$ \\
CDLM $-$ Base & 1 & $+0.177$ & $2.0\times10^{-59}$ & $+0.052$ & $7.9\times10^{-6}$ \\
CDLM $-$ Base & 4 & $+0.064$ & $4.9\times10^{-9}$ & $-0.005$ & $0.68$ \\
MDLM $-$ Base & 1 & $+0.062$ & $1.2\times10^{-10}$ & $-0.020$ & $0.057$ \\
MDLM $-$ Base & 4 & $+0.026$ & $0.0070$ & $-0.072$ & $4.3\times10^{-12}$ \\
\bottomrule
\end{tabular}
\end{table}

\subsection{Confidence definition and localization}
\label{app:llada8b_conf}

\paragraph{Probability of the present token versus maximum probability.}
The remasking rule thresholds $c_i=p_\theta(z_i\mid\bm{z})$, the probability of the token that is present.
Table~\ref{tab:llada8b_conf} contrasts it with the maximum predicted probability $\max_{v\in\mathcal{V}}p_\theta(v\mid\bm{z})$, which measures how certain the model is about what the token should be, on the same 541 HumanEval records.
Under $c_i$ the gap grows from 0.265 (base) to 0.402 (MDLM) and 0.777 (CDLM), and Hit@1 is 0.165, 0.152 and 0.608.
Under the maximum probability CDLM moves from the largest gap to the smallest (0.197, 0.219 and 0.117), and its Hit@1 becomes the lowest (0.325, 0.460 and 0.153).
The reason is visible in the clean and corrupted columns.
At corrupted positions CDLM's mean maximum probability is 0.866 while the probability of the token actually present is 0.192; at clean positions the two are 0.983 and 0.969.
CDLM stays sharp about what a corrupted position should contain and assigns little probability to the token that is there, so its error awareness appears as a collapse of $p_\theta(z_i\mid\bm{z})$ and not as a loss of predictive certainty, and a score that ignores the present token cannot detect it.
This is the confidence the remasking rule has always used, and the definition the paper states in Section~\ref{bg:refinement}.

\begin{table}[!htbp]
\centering
\small
\setlength{\tabcolsep}{3.5pt}
\caption{Localization on CRB HumanEval ($n_{\text{replace}}=1$, $n=541$ records, three error types pooled) under two confidence scores: the probability of the present token $p_\theta(z_i\mid\bm{z})$, which the remasking rule uses, and the maximum predicted probability. Clean and Corr.\ are mean scores at clean and corrupted positions; Gap is their difference averaged per record. The 200-step rows are separate runs with the same configuration.}
\label{tab:llada8b_conf}
\begin{tabular}{lc ccccc cccc}
\toprule
& & \multicolumn{5}{c}{$p_\theta(z_i\mid\bm{z})$} & \multicolumn{4}{c}{$\max_{v}p_\theta(v\mid\bm{z})$} \\
\cmidrule(lr){3-7}\cmidrule(lr){8-11}
\textbf{Arm} & Steps & Clean & Corr. & Gap & Hit@1 & Hit@5 & Clean & Corr. & Gap & Hit@1 \\
\midrule
Base & 0 & 0.970 & 0.705 & 0.265 & 0.165 & 0.771 & 0.988 & 0.791 & 0.197 & 0.325 \\
MDLM & 2000 & 0.970 & 0.567 & 0.402 & 0.152 & 0.874 & 0.990 & 0.771 & 0.219 & 0.460 \\
CDLM & 2000 & 0.969 & 0.192 & 0.777 & 0.608 & 0.980 & 0.983 & 0.866 & 0.117 & 0.153 \\
\midrule
MDLM & 200 & 0.963 & 0.705 & 0.258 & 0.105 & 0.806 & 0.986 & 0.804 & 0.181 & 0.287 \\
CDLM & 200 & 0.962 & 0.173 & 0.790 & 0.530 & 0.970 & 0.989 & 0.889 & 0.099 & 0.196 \\
\bottomrule
\end{tabular}
\end{table}

\paragraph{Error types.}
Table~\ref{tab:llada8b_types} splits the 541 records by error type.
CDLM's gap is 0.794 for operator, 0.857 for identifier and 0.681 for literal errors, and its Hit@1 is lowest on literals (0.355), the error type on which its repair rate is below base in three of the four datasets (Table~\ref{tab:llada8b_cells}).

\begin{table}[!htbp]
\centering
\small
\caption{Confidence gap and Hit@1 under $p_\theta(z_i\mid\bm{z})$ per error type on CRB HumanEval ($n_{\text{replace}}=1$, 2000-step adapters).}
\label{tab:llada8b_types}
\begin{tabular}{l cc cc cc}
\toprule
& \multicolumn{2}{c}{Operator ($n=250$)} & \multicolumn{2}{c}{Identifier ($n=136$)} & \multicolumn{2}{c}{Literal ($n=155$)} \\
\cmidrule(lr){2-3}\cmidrule(lr){4-5}\cmidrule(lr){6-7}
\textbf{Arm} & Gap & Hit@1 & Gap & Hit@1 & Gap & Hit@1 \\
\midrule
Base & 0.269 & 0.220 & 0.133 & 0.096 & 0.376 & 0.135 \\
MDLM & 0.426 & 0.148 & 0.303 & 0.147 & 0.452 & 0.161 \\
CDLM & 0.794 & 0.660 & 0.857 & 0.801 & 0.681 & 0.355 \\
\bottomrule
\end{tabular}
\end{table}

\paragraph{Onset after 200 steps.}
The separation is already present after 200 steps, that is, after 1{,}200 documents (0.37M tokens) at the end of warmup.
The 200-step CDLM run has a gap of 0.790 and Hit@1 of 0.530, against 0.258 and 0.105 for the 200-step MDLM run and 0.265 and 0.165 for base (Table~\ref{tab:llada8b_conf}).
Between 200 and 2000 steps the MDLM gap rises from 0.258 to 0.402 and its Hit@1 from 0.105 to 0.152, while CDLM's gap stays near 0.78 (0.790 and 0.777) and its Hit@1 rises from 0.530 to 0.608.

\paragraph{All four datasets.}
On HumanEval, absorbing-only training leaves Hit@1 flat (0.165 to 0.152).
This does not hold on the other datasets.
Table~\ref{tab:llada8b_all4} applies the localization protocol to the localization files of all four datasets ($n=3{,}559$).
Pooled over the four datasets, MDLM raises Hit@1 from 0.173 to 0.393 and CDLM to 0.691, and on HumanEval+, MBPP and MBPP+ MDLM's Hit@1 is clearly above base.
The gain over base is about 2.3 times larger for CDLM ($+0.519$ against $+0.221$), CDLM has the larger gap and Hit@1 on every dataset, and the pooled gap is 0.227 (base), 0.379 (MDLM) and 0.758 (CDLM).
The HumanEval observation is therefore specific to that dataset; the general statement is that the mixture objective improves localization much more than absorbing-only training on the same data.

\begin{table}[!htbp]
\centering
\small
\caption{Localization on the CRB localization files of all four datasets ($n_{\text{replace}}=1$, three error types pooled, $p_\theta(z_i\mid\bm{z})$, 2000-step adapters). Hit@1 per dataset (541, 536, 1{,}573 and 909 records) and pooled over all 3{,}559 records.}
\label{tab:llada8b_all4}
\begin{tabular}{l cccc ccc}
\toprule
& \multicolumn{4}{c}{Hit@1} & \multicolumn{3}{c}{Pooled} \\
\cmidrule(lr){2-5}\cmidrule(lr){6-8}
\textbf{Arm} & HE & HE+ & MBPP & MBPP+ & Hit@1 & Hit@5 & Gap \\
\midrule
Base & 0.165 & 0.209 & 0.147 & 0.200 & 0.173 & 0.773 & 0.227 \\
MDLM & 0.152 & 0.345 & 0.443 & 0.480 & 0.393 & 0.892 & 0.379 \\
CDLM & 0.608 & 0.465 & 0.747 & 0.778 & 0.691 & 0.962 & 0.758 \\
\bottomrule
\end{tabular}
\end{table}

\paragraph{Relation to Table~\ref{tab:confidence_refinement_benchmark}.}
Table~\ref{tab:confidence_refinement_benchmark} reports for LLaDA-8B-Base at $n_{\text{replace}}=1$ a gap of 0.235, a top-1 hit rate of 0.178 and Pass@1 0.552 at $T=1$, against 0.265, 0.165 and 0.551 in Table~\ref{tab:llada8b}.
The Pass@1 values come from the same inputs and protocol and differ only through the timing-out test discussed in Appendix~\ref{app:llada8b_fence}.
The localization values differ in protocol.
Table~\ref{tab:confidence_refinement_benchmark} reads the confidence from the first forward pass of the repair runs on the 3{,}184 repair inputs, computes the gap within each (dataset, error type) cell from token-level confidences pooled across programs, and averages gap and hit rate over the 12 cells.
Table~\ref{tab:llada8b} uses the 541 HumanEval localization records and averages per record.
Recomputing the procedure of Table~\ref{tab:confidence_refinement_benchmark} from its stored per-cell outputs reproduces 0.235 and 0.178; restricted to its three HumanEval cells, the same procedure gives 0.245 and 0.164.
Over all four datasets the localization files give 0.227 and 0.173 for base (Table~\ref{tab:llada8b_all4}), close to Table~\ref{tab:confidence_refinement_benchmark}.
The difference is thus mainly HumanEval versus all four datasets, with a smaller contribution from the record set and the within-cell pooling.

\subsection{Refinement depth on HumanEval+}
\label{app:llada8b_t4}

CDLM's de-fenced macro falls from 0.725 at $T=1$ to 0.623 at $T=4$, while base rises from 0.551 to 0.595 and MDLM from 0.600 to 0.609.
Table~\ref{tab:llada8b_t4} shows that the fall is confined to HumanEval+, where CDLM goes from 0.742 to 0.375; on HumanEval, MBPP and MBPP+ its Pass@1 changes by $-0.026$ to $+0.007$, and the HumanEval operator cell, which shares its problems with HumanEval+, goes from 0.817 to 0.821.
Without HumanEval+, CDLM's 9-cell macro is 0.719 at $T=1$ and 0.706 at $T=4$ (base 0.533 and 0.575, MDLM 0.601 and 0.605).
Two observations bear on the HumanEval+ cells.

\paragraph{The threshold rule does not stop.}
At $T=4$ a program stops early once no editable position has confidence at or below $\tau$.
For 89.2\% of CDLM's HumanEval+ programs (456 of 511; 222 of 255 in the operator cell) some position is still at or below the threshold in the fourth round, so the program uses all five forward passes.
Running to the budget is common in general, for 50.8\% to 65.2\% of CDLM's programs on the other datasets and for 41.9\% to 66.6\% of the other arms' programs on every dataset, but it is most frequent for CDLM on HumanEval+.

\paragraph{The extra rounds write fences into the leading blank line.}
At $T=1$ CDLM's HumanEval+ fences are almost all trailing (467 of 511 programs), and the de-fence rule removes them cleanly.
At $T=4$ CDLM also fills the leading blank line of the HumanEval+ body with an opening fence in 221 of 511 programs (18 at $T=1$), wrapping the program in a fenced block.
The de-fence rule truncates at the first fence, so these 221 programs are scored as empty bodies and all fail.
MDLM does this in 12 of 511 HumanEval+ programs at $T=4$, and neither arm does it on more than 4 programs of any other dataset.
The HumanEval+ drop therefore mixes a scoring interaction between the fence the model writes and the de-fence rule with changes in the repaired code.
Fences aside, CDLM's edits to HumanEval+ bodies remain small at both depths: 1.85\% of body characters at $T=1$ and 2.33\% at $T=4$ (characters in the non-matching blocks of a character-level alignment of the buggy body and the refined body with fence markers removed, summed over programs and divided by the total buggy-body length).

\paragraph{Diagnostic re-scoring.}
To separate the two, we re-scored the recorded programs of both fine-tuned arms with a rule that deletes every line starting with a fence, and any remaining fence marker, instead of truncating at the first fence.
This rule was written after the leading fences were found and is used for no other number in the paper.
It leaves every base program identical to its de-fenced version, and programs whose text is the same under both rules changed outcome between the two evaluations in 7 of 12{,}469 cases (test timeouts).
Under this rule CDLM's HumanEval+ Pass@1 is 0.772 at $T=1$ and 0.701 at $T=4$ (de-fenced: 0.742 and 0.375), and its 12-cell macro is 0.732 and 0.707 (de-fenced: 0.725 and 0.623); MDLM's macro is 0.600 and 0.616, and base stays at 0.551 and 0.595.
Neither rule removes the language tag of an opening fence: 38 of CDLM's 511 HumanEval+ programs at $T=4$ (13 at $T=1$) contain a line consisting only of \texttt{python}, and all of them fail; deleting that line as well gives 0.787 at $T=1$ and 0.756 at $T=4$. Most of CDLM's de-fenced drop at $T=4$ thus comes from fence artifacts, and a decline of about 0.03 on HumanEval+ remains.

\begin{table}[!htbp]
\centering
\small
\caption{Refinement depth per dataset ($n_{\text{replace}}=1$, three error types). Pass@1 is de-fenced, averaged over the three cells of each dataset. ``All 5 passes'' is the share of programs that still remask a position in the fourth round at $T=4$. ``Leading fence'' counts programs whose first fence precedes all code, which the de-fence rule empties.}
\label{tab:llada8b_t4}
\begin{tabular}{ll c cc c cc}
\toprule
& & & \multicolumn{2}{c}{Pass@1} & All 5 passes & \multicolumn{2}{c}{Leading fence} \\
\cmidrule(lr){4-5}\cmidrule(lr){7-8}
\textbf{Arm} & \textbf{Dataset} & $n$ & $T=1$ & $T=4$ & (\%), $T=4$ & $T=1$ & $T=4$ \\
\midrule
Base & HumanEval & 497 & 0.582 & 0.634 & 66.6 & 0 & 0 \\
Base & HumanEval+ & 511 & 0.607 & 0.655 & 56.4 & 0 & 0 \\
Base & MBPP & 1373 & 0.495 & 0.521 & 63.5 & 0 & 0 \\
Base & MBPP+ & 803 & 0.521 & 0.570 & 53.2 & 0 & 0 \\
\midrule
MDLM & HumanEval & 497 & 0.647 & 0.645 & 44.5 & 0 & 4 \\
MDLM & HumanEval+ & 511 & 0.598 & 0.622 & 41.9 & 0 & 12 \\
MDLM & MBPP & 1373 & 0.568 & 0.562 & 48.7 & 0 & 0 \\
MDLM & MBPP+ & 803 & 0.589 & 0.607 & 42.2 & 0 & 0 \\
\midrule
CDLM & HumanEval & 497 & 0.798 & 0.779 & 65.2 & 0 & 4 \\
CDLM & HumanEval+ & 511 & 0.742 & 0.375 & 89.2 & 18 & 221 \\
CDLM & MBPP & 1373 & 0.649 & 0.623 & 63.7 & 0 & 0 \\
CDLM & MBPP+ & 803 & 0.710 & 0.717 & 50.8 & 0 & 0 \\
\bottomrule
\end{tabular}
\end{table}

We report this as an observed limitation rather than an explained mechanism.
With a fixed budget, threshold-based refinement keeps rewriting low-confidence positions; here the fine-tuned model fills whitespace positions with formatting learned from its corpus, which interacts with how the output is scored, and a decline of about 0.03 on HumanEval+ remains once all fence artifacts are removed.
An explicit convergence or stopping criterion, or restricting the editable set to positions that carry code, are natural remedies that we have not evaluated.

\subsection{From-scratch generation check}
\label{app:llada8b_guard}

\paragraph{Protocol.}
To see how post-training affects writing programs from scratch at this scale, each model generates one program per problem for HumanEval+ (164 problems) and MBPP+ (378 problems), with the lm-evaluation-harness tasks, prompts, stop sequences and pinned inputs of the 0.5B from-scratch evaluation (Appendix~\ref{app:scratch_eval}).
Decoding appends 128 mask tokens to the prompt and runs 128 steps of confidence-based decoding with a linear schedule (every step predicts all masked positions greedily, then remasks the lowest-confidence fraction of the 128 positions given by the schedule), temperature 0, batch size 1.
The generated tokens are cut at the first end-of-text token and at the task's stop sequences, and the code is extracted with the task's filter followed by the harness's sanitizer (the HumanEval+ filter already applies it; the MBPP+ filter only removes code fences).
HumanEval+ is scored with the EvalPlus tests~\citep{liu2023humanevalplus}.
The primary MBPP+ score uses the three original MBPP asserts shown in the prompt, which are the task's own targets as in the 0.5B evaluation (so it is not comparable with EvalPlus MBPP+ scores); the score on the EvalPlus tests and the score with the task filter alone are reported separately.
Execution uses the HF \texttt{code\_eval} metric with a 20\,s timeout per problem.
Besides the two Nemotron adapters of this appendix, Table~\ref{tab:llada8b_guard} includes the two OpenCodeInstruct adapters of Appendix~\ref{app:oci}, trained with the same trainer and settings on a public corpus.

\paragraph{Results.}
Every adapter is below base on every metric of Table~\ref{tab:llada8b_guard}; on the three tested metrics the differences from base range from $-0.152$ to $-0.283$, with exact McNemar $p\le4.1\times10^{-5}$ (Table~\ref{tab:llada8b_guard_paired}).
CDLM and MDLM do not differ significantly on either corpus: on Nemotron-SFT-Code the differences are $0.000$, $-0.024$ and $0.000$ ($p=1$, $0.42$, $1$), and on OpenCodeInstruct $-0.037$, $-0.037$ and $-0.045$ ($p=0.33$, $0.21$, $0.082$).
We therefore do not claim that the corrective objective improves from-scratch generation at 8B; the from-scratch gains of Section~\ref{sec:when_correction_helps} are not reproduced in this setting.
The check shows only that the two objectives do not differ significantly when trained on the same data; on OpenCodeInstruct CDLM is numerically lower by 0.037 to 0.045.
Absolute comparisons with the base model are also sensitive to output format: within the 128 generated tokens the base model ends only 35 of 164 HumanEval+ and 215 of 378 MBPP+ programs with an end-of-text token and relies on the stop sequences and the sanitizer otherwise, whereas every adapter ends all but at most two programs per task with one.
The comparison between the two fine-tuned arms of the same corpus, which share their output format, is less exposed to this effect.

\begin{table}[!htbp]
\centering
\small
\caption{From-scratch generation on HumanEval+ (164 problems) and MBPP+ (378 problems): Pass@1 of one greedy sample per problem, 128 tokens in 128 steps. MBPP+ is scored on the three prompt asserts (primary), on the EvalPlus tests, and on the prompt asserts with the task filter alone.}
\label{tab:llada8b_guard}
\begin{tabular}{ll c ccc}
\toprule
& & & \multicolumn{3}{c}{MBPP+} \\
\cmidrule(lr){4-6}
\textbf{Model} & \textbf{Corpus} & HumanEval+ & Primary & EvalPlus & Task filter \\
\midrule
LLaDA-8B-Base & none & 0.280 & 0.529 & 0.447 & 0.381 \\
\quad + MDLM & Nemotron-SFT-Code & 0.073 & 0.270 & 0.196 & 0.206 \\
\quad + CDLM & Nemotron-SFT-Code & 0.073 & 0.246 & 0.196 & 0.217 \\
\quad + MDLM & OpenCodeInstruct & 0.128 & 0.352 & 0.294 & 0.294 \\
\quad + CDLM & OpenCodeInstruct & 0.091 & 0.315 & 0.249 & 0.315 \\
\bottomrule
\end{tabular}
\end{table}

\begin{table}[!htbp]
\centering
\small
\caption{Paired differences on the generation check of Table~\ref{tab:llada8b_guard}, with exact McNemar $p$-values in parentheses.}
\label{tab:llada8b_guard_paired}
\begin{tabular}{l ccc}
\toprule
& & \multicolumn{2}{c}{MBPP+} \\
\cmidrule(lr){3-4}
\textbf{Comparison} & HumanEval+ & Primary & EvalPlus \\
\midrule
\multicolumn{4}{l}{\textit{Nemotron-SFT-Code}} \\
CDLM $-$ MDLM & $0.000$ ($1$) & $-0.024$ ($0.42$) & $0.000$ ($1$) \\
MDLM $-$ Base & $-0.207$ ($5.4\times10^{-9}$) & $-0.259$ ($8.2\times10^{-18}$) & $-0.251$ ($8.4\times10^{-20}$) \\
CDLM $-$ Base & $-0.207$ ($1.1\times10^{-9}$) & $-0.283$ ($1.2\times10^{-21}$) & $-0.251$ ($2.1\times10^{-19}$) \\
\midrule
\multicolumn{4}{l}{\textit{OpenCodeInstruct}} \\
CDLM $-$ MDLM & $-0.037$ ($0.33$) & $-0.037$ ($0.21$) & $-0.045$ ($0.082$) \\
MDLM $-$ Base & $-0.152$ ($4.1\times10^{-5}$) & $-0.177$ ($1.5\times10^{-10}$) & $-0.153$ ($1.2\times10^{-9}$) \\
CDLM $-$ Base & $-0.189$ ($3.4\times10^{-7}$) & $-0.214$ ($5.1\times10^{-15}$) & $-0.198$ ($7.5\times10^{-15}$) \\
\bottomrule
\end{tabular}
\end{table}

%% file: appendix/alpha_sweep.tex
\section{Mixture-Probability Sweep at \texorpdfstring{$T=1$}{T=1} and \texorpdfstring{$T=4$}{T=4}}
\label{app:alpha_sweep}

\begin{figure*}[t]
    \centering
    \includegraphics[width=0.95\linewidth]{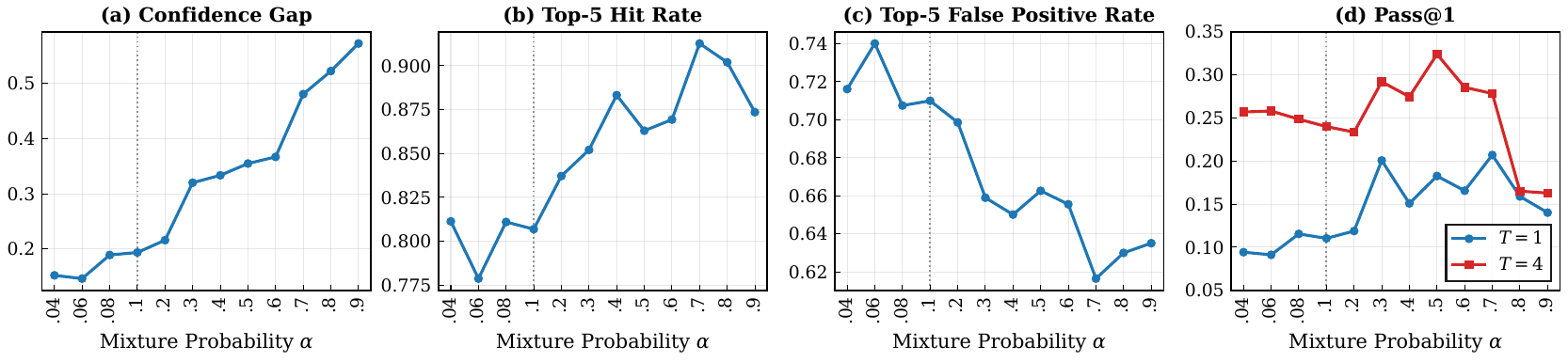}
    \caption{Effect of the mixture probability $\alpha$ (12 Open-dCoder-0.5B checkpoints, 2000 steps each; HumanEval CRB, mean over 3 error types $\times$ $n_{\mathrm{replace}}\in\{1,\ldots,5\}$, $\tau=0.9$). (a) Confidence gap. (b) Top-5 hit rate. (c) Top-5 false positive rate. (d) Pass@1 at $T=1$ and $T=4$. Panels (a)--(c) are read at the first forward pass. The dotted line marks $\alpha=0.1$, used in all other experiments; these checkpoints use a shorter learning-rate horizon than \textsc{CDLM-0.5B} (Appendix~\ref{app:alpha_sweep}).}
    \label{fig:mixture_prob_ablation}
\end{figure*}

This section gives the protocol and the full numbers behind Figure~\ref{fig:mixture_prob_ablation}.

\paragraph{Checkpoints.}
The sweep uses a separate family of twelve checkpoints, one for each of $\alpha=0.04$, $0.06$, $0.08$, $0.1$, $0.2$, $0.3$, \ldots, $0.9$. Each fine-tunes Open-dCoder-0.5B on the same Nemotron code corpus, and the training script is identical across the family except for $\alpha$ ($\lambda_{\mathrm{noise}}=0.1$, $\mathrm{ctw}=0$, global batch 12, sequence length 4096, seed 42, and 2000 training steps with the schedule length also set to 2000 steps, so the cosine learning-rate schedule decays fully within the run; Appendix~\ref{app:lr_horizon}). The \textsc{CDLM-0.5B} model of Section~\ref{sec:main_comparison} was instead saved at step 2000 of a run with a much longer schedule horizon. The two are therefore different checkpoints even at $\alpha=0.1$. $\alpha=0.1$ was fixed from preliminary experiments ($\alpha\in\{0.1, 0.2, \ldots, 1.0\}$) before the sweep in Figure~\ref{fig:mixture_prob_ablation} was run, and was not revisited; the sweep is a post-hoc sensitivity analysis.

\paragraph{Evaluation.}
Each checkpoint is evaluated on HumanEval CRB only, over 3 error types $\times$ $n_{\mathrm{replace}}\in\{1,\ldots,5\}$ (15 cells), with $\tau=0.9$ and greedy decoding. Every number in Table~\ref{tab:alpha_sweep} is the unweighted mean over the 15 cells. The confidence gap, the Top-5 hit rate and the Top-5 false positive rate are read from the first forward pass on the corrupted program, so they do not depend on $T$; Pass@1 is reported after $T=1$ and $T=4$ refinement steps (at most two and five forward passes). The Top-5 false positive rate is the fraction of the five lowest-confidence body positions, $\mathrm{Bottom}_5(c)$, that hold clean tokens, pooled over the programs of a cell. The confidence gap likewise pools all body positions of a cell, as in Table~\ref{tab:confidence_refinement_benchmark}. A program with $n_{\mathrm{replace}}=n$ has exactly $n$ corrupted positions, so the rate cannot fall below $1-n/5$; averaged over $n\in\{1,\ldots,5\}$ this floor is $0.4$, and only differences across $\alpha$ are informative. Operator corruptions are identical for all twelve checkpoints, whereas identifier and literal corruptions were drawn separately for each checkpoint, so each $\alpha$ is scored on 1{,}488 to 1{,}502 corrupted programs. For a given $\alpha$, $T=1$ and $T=4$ refine exactly the same programs.

\paragraph{Relation to the main results.}
Because of the different training horizon and the different aggregation, the sweep's $\alpha=0.1$ value (Pass@1 $0.110$ at $T=1$, HumanEval, mean over $n_{\mathrm{replace}}\in\{1,\ldots,5\}$) is not the main \textsc{CDLM-0.5B} number ($0.188$ at $T=1$, macro over the 12 dataset $\times$ error-type cells at $n_{\mathrm{replace}}=1$; Table~\ref{tab:ctw_ablation}). Comparisons below are between checkpoints of the sweep family.

\begin{table}[!htbp]
\centering
\small
\caption{Full mixture-probability sweep. Open-dCoder-0.5B fine-tuned for 2000 steps per $\alpha$, all other settings fixed; HumanEval CRB, unweighted mean over 3 error types $\times$ $n_{\mathrm{replace}}\in\{1,\ldots,5\}$, $\tau=0.9$. $N$: corrupted programs per $\alpha$, summed over the 15 cells. Gap, Top-5 hit and Top-5 FP (false positive) rate are measured at the first forward pass and do not depend on $T$. Rules separate the three regions discussed in the text. One training seed per $\alpha$.}
\label{tab:alpha_sweep}
\begin{tabular}{lcccccc}
\toprule
$\alpha$ & $N$ & Gap & Top-5 hit & Top-5 FP & Pass@1 $T{=}1$ & Pass@1 $T{=}4$ \\
\midrule
0.04 & 1{,}488 & 0.152 & 0.811 & 0.716 & 0.094 & 0.257 \\
0.06 & 1{,}495 & 0.146 & 0.779 & 0.740 & 0.091 & 0.258 \\
0.08 & 1{,}496 & 0.189 & 0.811 & 0.708 & 0.115 & 0.249 \\
0.1  & 1{,}502 & 0.194 & 0.807 & 0.710 & 0.110 & 0.240 \\
0.2  & 1{,}495 & 0.216 & 0.837 & 0.699 & 0.119 & 0.234 \\
\midrule
0.3  & 1{,}490 & 0.320 & 0.852 & 0.659 & 0.201 & 0.292 \\
0.4  & 1{,}489 & 0.333 & 0.883 & 0.650 & 0.151 & 0.275 \\
0.5  & 1{,}497 & 0.355 & 0.863 & 0.663 & 0.183 & 0.324 \\
0.6  & 1{,}490 & 0.367 & 0.869 & 0.656 & 0.166 & 0.286 \\
0.7  & 1{,}496 & 0.481 & 0.913 & 0.617 & 0.207 & 0.278 \\
\midrule
0.8  & 1{,}489 & 0.522 & 0.902 & 0.630 & 0.159 & 0.165 \\
0.9  & 1{,}490 & 0.572 & 0.874 & 0.635 & 0.140 & 0.163 \\
\bottomrule
\end{tabular}
\end{table}

\paragraph{Results.}
The confidence gap rises with $\alpha$, from $0.146$--$0.152$ at $\alpha\le0.06$ to $0.572$ at $\alpha=0.9$, and is monotone from $\alpha=0.06$ upward. Localization does not follow it: the Top-5 hit rate is highest at $\alpha=0.7$ ($0.913$) and the Top-5 false positive rate lowest there ($0.617$); both move the wrong way at $0.8$ and $0.9$. Pass@1 has the same three regions at both depths. For $\alpha\in[0.04,0.2]$, which contains $\alpha=0.1$, Pass@1 lies in $0.091$--$0.119$ at $T=1$ and $0.234$--$0.258$ at $T=4$. For $\alpha\in[0.3,0.7]$ it lies in $0.151$--$0.207$ at $T=1$ and $0.275$--$0.324$ at $T=4$. Within this family $\alpha=0.1$ is therefore not the best value, so the main results do not rest on a tuned $\alpha$; whether a larger $\alpha$ would also help \textsc{CDLM-0.5B} at its own training horizon was not tested.

We describe the good region as the interval $0.3$--$0.7$ rather than naming a peak. The $T=1$ maximum is at $\alpha=0.7$ and the $T=4$ maximum at $\alpha=0.5$, the $T=1$ curve is non-monotonic inside the interval ($0.201$, $0.151$, $0.183$, $0.166$, $0.207$ for $\alpha=0.3,\ldots,0.7$), and with one training seed per $\alpha$ the differences within $0.3$--$0.7$ cannot be separated from seed-to-seed variation.

Above $\alpha=0.7$ Pass@1 falls at both depths, but by different amounts. At $T=4$, $\alpha=0.8$ and $0.9$ reach $0.165$ and $0.163$, below every checkpoint with $\alpha\le0.2$. At $T=1$ the drop is milder: $\alpha=0.8$ ($0.159$) and $\alpha=0.9$ ($0.140$) remain above the low-$\alpha$ region. The loss is therefore concentrated in what additional refinement steps add: going from $T=1$ to $T=4$ gains $+0.006$ at $\alpha=0.8$ and $+0.023$ at $\alpha=0.9$, against $+0.071$ to $+0.167$ for every $\alpha\le0.7$. The trajectories differ from those of clean-token supervision (Section~\ref{sec:ctw}), where more programs leave refinement after the first pass. At $\alpha=0.8$ and $0.9$, $95.0\%$ and $97.0\%$ of programs use all four refinement steps at $T=4$ (some position is still at or below $\tau$ after each of the first four passes), against $57.5$--$88.8\%$ for $\alpha\le0.7$ ($57.7\%$ at $\alpha=0.1$); refinement keeps editing but repairs fewer programs.

%% file: appendix/oci.tex
\providecommand{\mdfence}{\texttt{\textasciigrave\textasciigrave\textasciigrave}}

\section{Replication on a Public Corpus (OpenCodeInstruct)}
\label{app:oci}

The controlled comparisons of Sections~\ref{sec:main_comparison} and~\ref{sec:llada8b} fine-tune on Nemotron-SFT-Code, a subset of Nemotron-Pretraining-SFT-v1~\citep{nvidia2025nvidianemotronnano2}.
That corpus is public but gated: access requires manual approval by NVIDIA under its data agreement, and the data cannot be redistributed with our release.
In addition, the documents a run reads depend on the order in which the shards are streamed and on the number of data-parallel ranks, so reconstructing the exact training subset requires the gated data and the recorded shard order.
To make both comparisons reproducible from openly available data, we retrained both arms of each comparison on OpenCodeInstruct~\citep{ahmad2025opencodeinstruct} with the released code (Appendix~\ref{app:release}), changing only the corpus.
We call the resulting models MDLM-OCI and CDLM-OCI.
They serve only as a replication; no result in the main text uses them.

\paragraph{Corpus.}
We use \texttt{nvidia/OpenCodeInstruct} at Hugging Face revision \texttt{8f3ba5ba} (CC BY 4.0, not gated; 5{,}000{,}000 rows in 50 parquet shards).
Every row is rendered as the plain text \texttt{input:}~$\langle$\textit{input}$\rangle$~\texttt{output:}~$\langle$\textit{output}$\rangle$, with single spaces as shown, and no row is filtered out.
The same rendered corpus is used at both scales.
Each arm is a single training run (seed 42), and every evaluation is a single deterministic decoding run.
Paired tests are exact McNemar tests on per-program outcomes pooled over cells, and confidence intervals are 10{,}000-resample paired bootstraps, as in Appendix~\ref{app:stats}.

\subsection{Open-dCoder-0.5B}
\label{app:oci_05b}

\paragraph{Training.}
Both arms continue training Open-dCoder-0.5B~\citep{opendllm2025} with the configuration of CDLM-0.5B and MDLM-0.5B (Appendix~\ref{app:hparams}), with only the corpus changed.
CDLM-OCI uses $\alpha=0.1$ and $\lambda_{\mathrm{noise}}=0.1$, MDLM-OCI uses $\alpha=0$ and $\lambda_{\mathrm{noise}}=0$, and both use clean-token weight 0.
The shared settings are AdamW with peak learning rate $3\times10^{-4}$, weight decay 0.01 and gradient clipping 1.0; the long-horizon cosine schedule stopped at step 2000, where the learning rate is still in linear warmup at $2.95\times10^{-5}$ (Appendix~\ref{app:lr_horizon}); a global batch of 12 packed sequences of 4096 tokens (3 per GPU on 4 GPUs); frozen embedding and output layers; and seed 42.
A run processes $2000\times12\times4096\approx98.3$M token positions and took about 19 minutes on 4 A100-PCIE-40GB GPUs.

\paragraph{Error localization and correction on CRB.}
Table~\ref{tab:oci_05b_crb} reports CRB for the two OpenCodeInstruct arms next to the base model and the two Nemotron-trained 0.5B models.
All five models refine the same input set, the released Open-dCoder CRB inputs (Appendix~\ref{app:release}), with the protocol of Section~\ref{bg:refinement} ($\tau=0.9$, temperature 0, batch size 1).
The Base and MDLM-0.5B values therefore differ slightly from Tables~\ref{tab:confidence_refinement_benchmark} and~\ref{tab:mixture_training_pass_at_1_comparison} (and the Base value from Table~\ref{tab:tau_sensitivity}), where each of these two models refined input files generated under its own name; those files contain the same operator instances but different identifier and literal instances (Appendix~\ref{app:release}).

\begin{table}[!htbp]
\centering
\small
\setlength{\tabcolsep}{5pt}
\caption{CRB for Open-dCoder-0.5B models trained on Nemotron-SFT-Code (Nemotron) and on OpenCodeInstruct (OCI). Pass@1 after $T$ refinement rounds, confidence gap, Hit@1 and Hit@5 (at least one corrupted position among the 1 or 5 lowest-confidence positions), macro-averaged over the dataset $\times$ error-type $[\times\, n_{\mathrm{replace}}]$ cells. All models refine the same inputs; the localization metrics do not depend on $T$. CDLM-0.5B's $T=4$ value is 0.285 here and 0.286 in Table~\ref{tab:ctw_ablation}; the difference is one program (HumanEval/139) whose tests run close to the 8\,s execution timeout.}
\label{tab:oci_05b_crb}
\begin{tabular}{ll cccc ccc}
\toprule
& & \multicolumn{4}{c}{Pass@1} & & & \\
\cmidrule(lr){3-6}
\textbf{Model} & \textbf{Corpus} & $T=1$ & $T=2$ & $T=3$ & $T=4$ & Gap & Hit@1 & Hit@5 \\
\midrule
\multicolumn{9}{l}{\textit{$n_{\mathrm{replace}}=1$: 12 cells, 3{,}213 test-failing programs}} \\
Base & none & 0.145 & 0.240 & 0.255 & 0.263 & 0.105 & 0.147 & 0.511 \\
MDLM-0.5B & Nemotron & 0.138 & 0.236 & 0.246 & 0.245 & 0.098 & 0.147 & 0.503 \\
CDLM-0.5B & Nemotron & 0.188 & 0.279 & 0.281 & 0.285 & 0.164 & 0.228 & 0.606 \\
MDLM-OCI & OCI & 0.140 & 0.223 & 0.235 & 0.239 & 0.099 & 0.145 & 0.520 \\
CDLM-OCI & OCI & \textbf{0.220} & \textbf{0.302} & \textbf{0.303} & \textbf{0.307} & \textbf{0.192} & \textbf{0.252} & \textbf{0.650} \\
\midrule
\multicolumn{9}{l}{\textit{$n_{\mathrm{replace}}\in\{1,\dots,5\}$: 60 cells, 9{,}257 test-failing programs}} \\
Base & none & 0.053 & 0.135 & 0.153 & 0.153 & 0.090 & 0.254 & 0.684 \\
MDLM-0.5B & Nemotron & 0.048 & 0.118 & 0.136 & 0.139 & 0.082 & 0.250 & 0.683 \\
CDLM-0.5B & Nemotron & 0.076 & 0.208 & 0.221 & 0.225 & 0.148 & 0.363 & 0.770 \\
MDLM-OCI & OCI & 0.048 & 0.128 & 0.145 & 0.150 & 0.085 & 0.258 & 0.699 \\
CDLM-OCI & OCI & \textbf{0.095} & \textbf{0.222} & \textbf{0.233} & \textbf{0.235} & \textbf{0.173} & \textbf{0.396} & \textbf{0.805} \\
\bottomrule
\end{tabular}
\end{table}

The advantage of the corrective objective replicates on the public corpus.
At $n_{\mathrm{replace}}=1$, CDLM-OCI repairs more programs than MDLM-OCI at every depth.
Pooled over the 3{,}213 test-failing programs, the difference is $+0.076$ at $T=1$ (95\% CI $[+0.065,+0.087]$, $p=6.0\times10^{-43}$), $+0.084$ at $T=2$ ($p=1.0\times10^{-29}$), $+0.080$ at $T=3$ ($p=3.3\times10^{-27}$) and $+0.078$ at $T=4$ ($[+0.064,+0.092]$, $p=5.6\times10^{-27}$).
Over $n_{\mathrm{replace}}\in\{1,\dots,5\}$ (9{,}257 programs) it is $+0.053$ at $T=1$ ($p=3.7\times10^{-93}$) and $+0.090$ at $T=4$ ($p=9.2\times10^{-104}$).
Localization moves with repair: the confidence gap rises from 0.099 (MDLM-OCI) to 0.192 (CDLM-OCI) and Hit@1 from 0.145 to 0.252.
In macro terms, the $T=1$ advantage of the mixture arm over the absorbing-only arm is $+0.079$ on OpenCodeInstruct and $+0.050$ on Nemotron-SFT-Code.

Changing the corpus does not move the absorbing-only arm at $T=1$: MDLM-OCI and MDLM-0.5B differ by $+0.001$ ($p=0.78$), and neither improves on the base model at any depth.
CDLM-OCI is above CDLM-0.5B by $+0.032$ at $T=1$ ($p=1.2\times10^{-14}$) and $+0.030$ at $T=4$ ($p=9.0\times10^{-7}$).
We do not attribute this difference to any property of the corpora: they differ in content, format and length, and each arm is a single run.

\paragraph{From-scratch code generation.}
Table~\ref{tab:oci_05b_codegen} evaluates the two OpenCodeInstruct arms with the from-scratch protocol of Table~\ref{tab:pure_completion_results} and Appendix~\ref{app:scratch_eval} (10 samples per problem, temperature 0.8, top-$k$ 200, 128 tokens in 128 steps, vanilla and ReMDM~\citep{wang2025remasking} decoding).
Relative to the matched absorbing-only arm, CDLM-OCI is higher on HumanEval and HumanEval+ under both decoders, with 95\% confidence intervals that exclude zero, and indistinguishable on MBPP and MBPP+.
Pass@10 differences between the two arms range from $-0.015$ to $+0.024$, and every 95\% interval contains zero.
On MBPP and MBPP+ both OpenCodeInstruct arms score lower than the base model, as both Nemotron-trained arms do (Table~\ref{tab:mbpp_results_appendix}), so the MBPP difference to the base model is present after fine-tuning on either corpus with either objective.

\begin{table}[!htbp]
\centering
\small
\caption{From-scratch generation Pass@1 of the Open-dCoder-0.5B models trained on OpenCodeInstruct, with the protocol of Table~\ref{tab:pure_completion_results}. Base is the recorded run of the untrained model behind Tables~\ref{tab:pure_completion_results} and~\ref{tab:mbpp_results_appendix}, shown to three decimals. The last column is CDLM-OCI minus MDLM-OCI with a paired bootstrap 95\% CI over problems (164 for HumanEval and HumanEval+, 500 for MBPP, 378 for MBPP+).}
\label{tab:oci_05b_codegen}
\begin{tabular}{ll ccc c}
\toprule
\textbf{Benchmark} & \textbf{Decoder} & Base & MDLM-OCI & CDLM-OCI & Difference [95\% CI] \\
\midrule
HumanEval  & vanilla & 0.188 & 0.171 & \textbf{0.216} & $+0.045$ [$+0.015$, $+0.077$] \\
HumanEval+ & vanilla & 0.167 & 0.153 & \textbf{0.196} & $+0.043$ [$+0.013$, $+0.073$] \\
MBPP       & vanilla & \textbf{0.157} & 0.121 & 0.116 & $-0.005$ [$-0.024$, $+0.013$] \\
MBPP+      & vanilla & \textbf{0.231} & 0.190 & 0.203 & $+0.013$ [$-0.010$, $+0.037$] \\
\midrule
HumanEval  & ReMDM & 0.182 & 0.160 & \textbf{0.221} & $+0.061$ [$+0.024$, $+0.099$] \\
HumanEval+ & ReMDM & 0.164 & 0.144 & \textbf{0.198} & $+0.054$ [$+0.018$, $+0.091$] \\
MBPP       & ReMDM & \textbf{0.160} & 0.122 & 0.119 & $-0.003$ [$-0.019$, $+0.013$] \\
MBPP+      & ReMDM & \textbf{0.232} & 0.184 & 0.195 & $+0.011$ [$-0.012$, $+0.033$] \\
\bottomrule
\end{tabular}
\end{table}

\subsection{LLaDA-8B-Base}
\label{app:oci_8b}

\paragraph{Training.}
Two LoRA adapters~\citep{hu2022lora} on LLaDA-8B-Base~\citep{nie2025largelanguagediffusionmodels} (revision \texttt{0f2787f2}) were trained with the trainer, launch flags and seed of the Nemotron adapters (Appendix~\ref{app:llada8b_setup}): 2000 steps, peak learning rate $3\times10^{-4}$ with 200 warmup steps and cosine decay over a 20{,}000-step horizon, global batch 12 sequences of at most 4096 tokens, LoRA rank 64, $\alpha=128$ and dropout 0.05, frozen embeddings and output layer, seed 42.
Only the corpus differs from the Nemotron adapters, and within the pair only $\alpha$ and $\lambda_{\mathrm{noise}}$ differ (0.1 and 0.1 for CDLM-OCI, 0 and 0 for MDLM-OCI).
The runs took 51 (MDLM-OCI) and 56 (CDLM-OCI) minutes on 4 A100-PCIE-40GB GPUs shared with other jobs.

\paragraph{What a run reads.}
The trainer reads the data as described in Appendix~\ref{app:llada8b_setup}: the shard files are shuffled with seed 42, rank $r$ reads files $r, r+4, \dots$ from their first row with one document per sequence, and each document is delivered twice, with independent corruption draws.
A 2000-step run therefore reads 12{,}000 distinct documents, the first 3{,}000 rows of shards 25, 23, 19 and 11 of OpenCodeInstruct, with 6.58M tokens (548 per document on average); the Nemotron runs read 12{,}000 documents with 3.55M tokens at the same positions of their stream.
OpenCodeInstruct stores its categories in long contiguous runs within each shard, so these documents are not a sample of the corpus: 9{,}000 of them are \emph{generic / evol-instruct} and 3{,}000 \emph{algorithmic / self-instruct}, whereas the full corpus is 38.9\% generic / evol-instruct, 33.4\% generic / self-instruct, 19.6\% algorithmic / self-instruct and 8.1\% algorithmic / evol-instruct.
Both arms read the same documents in the same order.

\paragraph{Results.}
Table~\ref{tab:oci_8b} uses the protocol of Table~\ref{tab:llada8b}: localization on the 541 records of the three HumanEval CRB files at $n_{\mathrm{replace}}=1$, with confidence $p_\theta(z_i\mid\bm{z})$; Pass@1 macro-averaged over the 12 dataset $\times$ error-type cells at $n_{\mathrm{replace}}=1$ (3{,}184 test-failing programs), with de-fenced Pass@1 as the primary convention and raw Pass@1 as in Appendix~\ref{app:llada8b_fence}.

\begin{table}[!htbp]
\centering
\small
\caption{LLaDA-8B-Base LoRA adapters trained on OpenCodeInstruct (2000 steps), with the protocol of Table~\ref{tab:llada8b}, which reports the Nemotron-trained adapters. Gap, Hit@1 and Hit@5 on the HumanEval localization files ($n=541$); Pass@1 macro over 12 cells at $n_{\mathrm{replace}}=1$, de-fenced (primary) and raw. The Base row is the run reported in Table~\ref{tab:llada8b}.}
\label{tab:oci_8b}
\begin{tabular}{l ccc cc cc}
\toprule
& & & & \multicolumn{2}{c}{Pass@1, de-fenced} & \multicolumn{2}{c}{Pass@1, raw} \\
\cmidrule(lr){5-6}\cmidrule(lr){7-8}
\textbf{Model} & Gap & Hit@1 & Hit@5 & $T=1$ & $T=4$ & $T=1$ & $T=4$ \\
\midrule
LLaDA-8B-Base & 0.265 & 0.165 & 0.771 & 0.551 & 0.595 & \textbf{0.551} & \textbf{0.595} \\
\quad + MDLM-OCI & 0.338 & 0.102 & 0.887 & 0.576 & 0.590 & 0.546 & 0.521 \\
\quad + CDLM-OCI & \textbf{0.835} & \textbf{0.593} & \textbf{0.978} & \textbf{0.741} & \textbf{0.726} & 0.518 & 0.506 \\
\bottomrule
\end{tabular}
\end{table}

The 8B result also replicates on the public corpus.
The mixture objective raises the confidence gap from 0.265 (base) to 0.835 and Hit@1 from 0.165 to 0.593, against 0.338 and 0.102 for the absorbing-only arm.
De-fenced Pass@1 is 0.741 at $T=1$ and 0.726 at $T=4$ for CDLM-OCI, 0.576 and 0.590 for MDLM-OCI, and 0.551 and 0.595 for the base model.
Pooled over the 3{,}184 test-failing programs, CDLM-OCI exceeds MDLM-OCI by $+0.170$ at $T=1$ (95\% CI $[+0.150,+0.191]$, $p=2.3\times10^{-59}$) and $+0.145$ at $T=4$ ($[+0.125,+0.166]$, $p=3.5\times10^{-44}$) de-fenced, and by $+0.042$ ($p=2.2\times10^{-4}$) and $+0.046$ ($p=1.7\times10^{-5}$) raw.

The raw macro nevertheless places CDLM-OCI below MDLM-OCI (0.518 against 0.546 at $T=1$).
This is the fence effect of Appendix~\ref{app:llada8b_fence}: at $T=1$, 99.6\% of CDLM-OCI's refined HumanEval+ programs contain a markdown fence (\mdfence), against 10.4\% for MDLM-OCI and none for the base model, so its three raw HumanEval+ cells are 0.004, 0.000 and 0.000 and weigh a quarter of the macro.
Without HumanEval+, the raw macro at $T=1$ is 0.691 (CDLM-OCI), 0.538 (MDLM-OCI) and 0.533 (base).

On the HumanEval localization files, MDLM-OCI's Hit@1 is 0.102 against 0.165 for the base model.
As for the Nemotron adapters (Appendix~\ref{app:llada8b_conf}), this pattern is specific to HumanEval: pooled over the localization files of all four datasets ($n=3{,}559$), Hit@1 is 0.173 for the base model, 0.409 for MDLM-OCI and 0.732 for CDLM-OCI, and the gap is 0.227, 0.329 and 0.809.
The gain in Hit@1 over the base model is $+0.559$ for the mixture objective and $+0.236$ for the absorbing-only objective, about 2.4 times larger.

CDLM-OCI does not show the large decline from $T=1$ to $T=4$ of the Nemotron CDLM adapter (Appendix~\ref{app:llada8b_t4}): its de-fenced HumanEval+ Pass@1 (mean of the three error types) is 0.769 at $T=1$ and 0.754 at $T=4$, against 0.742 and 0.375 for the Nemotron CDLM adapter.
Compared with the Nemotron CDLM adapter on the same programs, CDLM-OCI is $+0.014$ at $T=1$ ($p=0.10$) and $+0.067$ at $T=4$ ($p=1.4\times10^{-13}$), de-fenced.

The base row of Table~\ref{tab:oci_8b} is the run reported in Table~\ref{tab:llada8b}.
We re-ran the base model with the released pipeline in the same evaluation as the two OpenCodeInstruct adapters: all 6{,}368 refined programs (3{,}184 at each depth) were identical to those of the reported run, and test outcomes differed only for HumanEval/139, whose tests run close to the execution timeout; no macro value changed by more than 0.001.

\paragraph{From-scratch generation.}
Table~\ref{tab:llada8b_guard} includes the two OpenCodeInstruct adapters in the generation check of Appendix~\ref{app:llada8b_guard}.
Pass@1 on HumanEval+ and MBPP+ is 0.280 and 0.529 for the base model, 0.128 and 0.352 for MDLM-OCI, and 0.091 and 0.315 for CDLM-OCI.
Both fine-tuned arms are below the base model (MDLM-OCI: $-0.152$, $p=4.1\times10^{-5}$, and $-0.177$, $p=1.5\times10^{-10}$; CDLM-OCI: $-0.189$, $p=3.4\times10^{-7}$, and $-0.214$, $p=5.1\times10^{-15}$).
CDLM-OCI and MDLM-OCI do not differ significantly ($-0.037$, $p=0.33$, and $-0.037$, $p=0.21$).
As with the Nemotron adapters, we make no claim that the corrective objective improves from-scratch generation at 8B.

\paragraph{Scope.}
Each arm at each scale is a single training run, so run-to-run variance of these replications is not measured.
At 8B the training data are the first rows of four category-grouped shards rather than a sample of OpenCodeInstruct.
Absolute numbers are not comparable across the two corpora; the replication concerns the comparison between the two objectives on identical data, which favors the mixture objective at both scales and on both corpora.

%% file: appendix/concurrent.tex
\section{Comparison with Concurrent Corrective Methods}
\label{app:concurrent}

This appendix compares CDLM with two concurrent approaches to error correction in DLMs: a learned self-correction head (PRISM, Appendix~\ref{app:prism}) and error localization from the temporal dynamics of decoding (Appendix~\ref{app:temporal}; related work in Appendix~\ref{app:rw_temporal}).
Both studies use the CRB protocol of Section~\ref{sec:experiments}: confidence threshold $\tau=0.9$, temperature $0$, and $T$ remask-then-repredict rounds, which cost $T+1$ full-sequence forward passes.
Confidence is $c_i=p_\theta(z_i\mid\bm{z})$, the probability the model assigns to the token currently at position $i$.
Hit@$K$ (top-$K$ hit rate) is the fraction of programs with at least one corrupted position among the $K$ positions ranked most suspicious, which for $c_i$ are the $K$ lowest-confidence positions.

\subsection{Self-Correction Head (PRISM)}
\label{app:prism}

\paragraph{Method.}
\citet{kim2025finetuningmaskeddiffusionprovable} (PRISM) fine-tune a masked diffusion model jointly with a small quality head that estimates, for each token, the probability that it is correct; at inference, tokens with low estimated quality are sent back to the mask token.
The head (LayerNorm, a $d\times d$ linear layer, GELU and a $d\times 1$ linear layer, followed by a sigmoid) reads the backbone's last hidden state.
Each training step takes two forward passes.
The first predicts the masked positions of a masked sequence; $K$ of them are then filled with the model's own greedy predictions, which gives a partially decoded sequence.
The second pass scores that sequence, and the head is trained with a binary cross-entropy loss, at the $K$ newly filled positions only, to predict whether each filled token equals the ground truth.
A masked-token cross-entropy term with weight $0.5$ is added as a regularizer.

\paragraph{Training.}
No PRISM checkpoints are released, so we reimplemented the method from the authors' code and trained it in the CDLM setup: the same base checkpoint (Open-dCoder-0.5B), corpus (Nemotron-SFT-Code), optimizer (AdamW, peak learning rate $3\times10^{-4}$, weight decay $0.01$, cosine decay, gradient clipping at $1.0$), seed ($42$), global batch size ($12$ packed sequences of $4{,}096$ tokens; the PRISM run accumulated gradients over $12$ micro-batches on one GPU), frozen layers (input embeddings and output head; the rest of the backbone is trained together with the head) and number of steps ($2{,}000$).
The PRISM-specific hyperparameters are the authors' defaults ($K=8$ per $1{,}024$ tokens, regularizer weight $0.5$).
We use their \texttt{random} token-sampling variant (the configuration used in their own Sudoku and OpenWebText experiments); the semi-autoregressive variant produced degenerate supervision in our code-repair setting.

One training setting differs: the length of the learning-rate schedule.
Our description of this comparison during the review discussion called the schedule identical, which was not accurate.
The CDLM and MDLM checkpoints are step $2{,}000$ of runs whose cosine schedule spans far more steps, so at step $2{,}000$ they were still in a $20{,}345$-step linear warmup (learning rate $2.95\times10^{-5}$, $9.83\%$ of peak).
The PRISM run set its schedule length to $2{,}000$ steps ($2$ warmup steps, then cosine decay to $3\times10^{-6}$ at step $2{,}000$).
Over the same $2{,}000$ updates, the summed learning rate of the PRISM run is therefore about ten times that of the CDLM and MDLM runs.
The comparison below does not control for this difference.

\paragraph{Adapting PRISM to in-place repair.}
PRISM's released sampler applies the head only inside the semi-autoregressive block currently being generated, and it skips any block that contains no mask token.
A CRB input is a complete program with no masked positions, so the released sampler returns it unchanged.
We made two adaptations, both in PRISM's favor, since without them the method makes no edit at all: (i) we score every editable (non-prompt) position of the complete program, instead of only the decoded tokens of a block that still contains masks; and (ii) we drop the semi-autoregressive block restriction.
The score itself is computed as in their code.
It replaces $c_i$ in the remasking step of our refinement loop; the threshold test, step budget and stopping rule are unchanged, so within the refinement loop the two methods differ in the score that decides what to remask and in the fine-tuned backbone that re-predicts the remasked positions.
Open-dCoder inherits the Qwen2 convention that the hidden state at position $i-1$ predicts token $i$, so we read the head with the same one-position shift as the logits.
A sanity check supports the implementation: during training, the head's accuracy rises from $0.45$ at the first step to about $0.85$ within $40$ steps, with about two thirds of the filled tokens labeled correct.

\paragraph{Two operating points.}
The head's score is a sigmoid output on a different scale from a token probability, so the same numerical threshold implies a different edit budget.
We therefore evaluate the head at two operating points.
(a)~$\tau=0.9$, the threshold of every CDLM result. Here the head remasks $5.72\%$ of the editable tokens in the first round, against $2.87\%$ for CDLM.
(b)~Exact edit parity. PRISM's own sampler does not threshold; it remasks a fixed number of the lowest-scoring tokens.
We run the head in this ranking mode and set the number of tokens it remasks, for every program and every round, to the number CDLM remasked in that program and round.
Both methods then make the same number of edits on the same programs and differ in which tokens they select and in the backbone that refills them.
We verified the match from the logged refinement histories: the per-program, per-round edit counts are identical in all $24$ cells ($12$ dataset $\times$ error-type cells at $T=1$ and at $T=4$).
For example, both methods make $663$ edits in total on HumanEval operator errors at $T=1$, $1{,}611$ at $T=4$, and $468$ on HumanEval identifier errors at $T=1$.

\paragraph{Evaluation.}
Both head configurations are run on exactly the $3{,}213$ corrupted programs on which CDLM is scored ($n_{\mathrm{replace}}=1$, $12$ cells) and are scored by the same evaluator.
The Base, MDLM and CDLM rows of Table~\ref{tab:prism_h2h} are the runs of the main comparison (Section~\ref{sec:main_comparison}); the Base row reproduces the Open-dCoder-0.5B entries of Table~\ref{tab:confidence_refinement_benchmark} at $n_{\mathrm{replace}}=1$.
Base and MDLM are scored on their own CRB draws ($3{,}230$ and $3{,}220$ programs), which coincide with CDLM's for operator errors and differ in part for identifier and literal errors (Appendix~\ref{app:stats_exact}).

\begin{table}[!htbp]
\centering
\small
\caption{Self-correction head (PRISM) versus CDLM on CRB with the Open-dCoder-0.5B backbone, $n_{\mathrm{replace}}=1$, macro-averaged over the $12$ dataset $\times$ error-type cells. Gap (mean score of clean positions minus mean score of corrupted positions, pooled over the positions of each cell as in Table~\ref{tab:confidence_refinement_benchmark}; averaged per program as in Tables~\ref{tab:stats_seeds} and~\ref{tab:oci_05b_crb}, CDLM's gap is $0.164$) and Hit@1 are computed from the first-round scores: $c_i$ for Base, MDLM and CDLM, and the head's sigmoid output for PRISM. Remask is the percentage of editable tokens remasked in the first round. $^{\dagger}$Equal to CDLM's by construction: the head remasks, in every program and round, as many tokens as CDLM.}
\label{tab:prism_h2h}
\begin{tabular}{lccccc}
\toprule
 & \multicolumn{2}{c}{Pass@1} & & & \\
\cmidrule(lr){2-3}
Model & $T=1$ & $T=4$ & Gap & Hit@1 & Remask (\%) \\
\midrule
Base (Open-dCoder-0.5B) & 0.143 & 0.251 & 0.106 & 0.147 & 3.09 \\
MDLM (absorbing-only) & 0.137 & 0.237 & 0.095 & 0.140 & 2.95 \\
CDLM (ours) & \textbf{0.188} & \textbf{0.286} & \textbf{0.169} & \textbf{0.228} & 2.87 \\
\midrule
PRISM head, $\tau=0.9$ & 0.132 & 0.167 & 0.087 & 0.111 & 5.72 \\
PRISM head, exact edit parity & 0.107 & 0.134 & 0.087 & 0.111 & 2.87$^{\dagger}$ \\
\bottomrule
\end{tabular}
\end{table}

\paragraph{Results.}
At $\tau=0.9$ the head remasks about twice as many tokens as CDLM ($5.72\%$ against $2.87\%$) and repairs fewer programs (Pass@1 $0.132$ against $0.188$ at $T=1$, and $0.167$ against $0.286$ at $T=4$).
At exact edit parity it reaches $0.107$ and $0.134$.
CDLM, in turn, remasks fewer tokens than the base model ($3.09\%$) and MDLM ($2.95\%$) while repairing more programs, so its gain does not come from editing more.
Gap and Hit@1 are identical for the two head configurations because they are computed from the first-round scores, which depend only on the model and the input and not on the rule that turns scores into edits.
They therefore measure localization independently of the operating point: the head's Hit@1 ($0.111$) and gap ($0.087$) are below those of the untrained base model's $c_i$ ($0.147$ and $0.106$), while CDLM raises both ($0.228$ and $0.169$).
A CRB program at $n_{\mathrm{replace}}=1$ contains a single corrupted token, so every remasked clean token is a chance to introduce a new error, and imprecise localization costs more as rounds accumulate.
From $T=1$ to $T=4$, Pass@1 rises by $+0.108$ for the base model and by $+0.098$ for CDLM, but by $+0.034$ for the head at $\tau=0.9$ and by $+0.027$ for the head at exact edit parity (differences computed before rounding).

\paragraph{Scope of the comparison.}
We read these results as a statement about task fit rather than about the quality of PRISM, and the reasons are in its favor.
PRISM targets self-correction during generation: its head decides which just-generated tokens to send back to the mask token while a sequence is still being decoded.
CRB instead presents a complete program that the model did not write and asks which of its visible tokens are wrong.
PRISM's released sampler skips any block that contains no mask token, so it returns a fully visible corrupted program unchanged, and its head is only ever supervised on tokens the model has just generated inside a sequence that still contains masks.
A CRB input differs from that training distribution both in mask density (it has no masks) and in the source of its errors (an external edit rather than the model's own predictions).
Applying the method to in-place repair required the two adaptations above, both made in its favor.
We make no claim about their method in the setting their paper actually evaluates.

\paragraph{Cost.}
Independently of accuracy, the two approaches differ in cost.
PRISM adds a quality head with $806{,}401$ parameters at Open-dCoder's hidden size of $896$, a second forward pass per training step, and supervision constructed from a model rollout compared against the ground truth.
CDLM adds no parameters, uses one forward pass per training step, and needs no data construction, since its targets are the clean tokens.
At inference both use one forward pass per round, because the head shares the backbone's pass.

\subsection{Temporal Trajectory Features}
\label{app:temporal}

Temporal-dynamics methods (Appendix~\ref{app:rw_temporal}) read a signal from the denoising trajectory.
In-place revision has no such trajectory.
The input is a fully visible corrupted program with no masked positions, repeated forward passes on it are deterministic and return identical outputs, and the positions to edit must be chosen from a single pass before any edit is made.
A trajectory therefore has to be induced before these methods apply.
We induce one by unrolling the refinement loop itself, and ask whether features of the resulting trajectory localize corrupted tokens better than single-pass confidence.

\paragraph{Setup.}
We use the logged per-round histories of the $T=4$ CRB runs at $\tau=0.9$, which record, after each of the five forward passes, the token at every editable position and its confidence $c_i$.
The study covers $540$ runs and $83{,}332$ programs: $9$ models, $4$ datasets (HumanEval, HumanEval+, MBPP, MBPP+), $3$ error types and $n_{\mathrm{replace}}\in\{1,\ldots,5\}$; no program was excluded.
The nine models are the Open-dCoder-0.5B base model, its MDLM and CDLM fine-tunes at $2{,}000$ steps, a CDLM checkpoint trained for $10{,}000$ steps, the clean-token-supervised ablation CDLM$+$ctw (Section~\ref{sec:ctw}), LLaDA-8B-Base, Dream-7B-Base, and Fast-dLLM-v2 at $1.5$B and $7$B parameters~\citep{wu2025fastdllmv2efficientblockdiffusion}.
For each editable position $i$ we compute, over the five passes, the flip rate (fraction of rounds in which the token at $i$ changed), the convergence step (last round in which it changed), the variance of $c_i$, the minimum of $c_i$, and the least-squares slope of $c_i$ against the round index.
The baseline is the single-pass confidence $c_i^{(0)}$ that the remasking rule already uses.
Each feature defines a ranking of positions in two orientations (a low or a high value marks a suspicious position), and we report the better orientation.
We score the rankings with Hit@1 and Hit@5 as in Section~\ref{bench:error_identification}, breaking ties uniformly at random and averaging over five draws, and with AUROC, the probability that a clean position ranks as less suspicious than a corrupted one (ties count one half).
All metrics are computed per program and averaged over programs.

\begin{table}[!htbp]
\centering
\small
\caption{Trajectory features versus single-pass confidence for error localization on CRB HumanEval: all nine models, three error types and $n_{\mathrm{replace}}\in\{1,\ldots,5\}$ ($13{,}446$ programs). Each feature is shown in the better of its two orientations. Passes: forward passes needed to compute the feature.}
\label{tab:temporal_features}
\begin{tabular}{lcccc}
\toprule
Feature & Passes & Hit@1 & Hit@5 & AUROC \\
\midrule
Single-pass confidence $c_i^{(0)}$ & 1 & 0.359 & 0.772 & 0.891 \\
Flip rate & 5 & 0.150 & 0.480 & 0.740 \\
Convergence step & 5 & 0.142 & 0.489 & 0.742 \\
Confidence variance & 5 & 0.253 & 0.622 & 0.840 \\
Minimum confidence & 5 & 0.296 & 0.708 & 0.893 \\
\midrule
Confidence slope (rising $=$ suspicious) & 5 & 0.434 & 0.752 & 0.735 \\
\bottomrule
\end{tabular}
\end{table}

\paragraph{Results.}
On HumanEval (Table~\ref{tab:temporal_features}), none of flip rate, convergence step, confidence variance and minimum confidence improves on single-pass confidence in Hit@1 or Hit@5, although each needs five times as many forward passes.
Minimum confidence matches single-pass confidence in AUROC ($0.893$ against $0.891$) while trailing it on both hit rates.
We also computed the mean and the last-pass confidence and two z-scored combinations of features (one of which includes single-pass confidence); none of them exceeds single-pass confidence on any of the three metrics on HumanEval.

Table~\ref{tab:temporal_datasets} repeats the comparison on all four datasets.
On MBPP and MBPP+, single-pass confidence has the highest Hit@1 apart from confidence slope and also the highest AUROC.
HumanEval+ is an exception: its single-pass Hit@1 is only $0.176$, and confidence variance ($0.269$) and minimum confidence ($0.189$) exceed it, as do minimum confidence and confidence variance in AUROC ($0.895$ and $0.881$ against $0.880$).
This exception comes from the inputs.
In all $14{,}273$ HumanEval+ programs of the study, the editable body begins with a blank line (against $163$ of $13{,}446$ HumanEval programs and none in MBPP or MBPP+), and the first body token is the single-pass argmin in $61.6\%$ of HumanEval+ programs ($1.7\%$ on HumanEval, at most $0.5\%$ on MBPP and MBPP+), although it is never a corrupted position.
Pooled over all four datasets ($83{,}332$ programs), single-pass confidence has the highest Hit@1 apart from confidence slope, and minimum confidence matches it in AUROC ($0.874$ for both).

\begin{table}[!htbp]
\centering
\small
\setlength{\tabcolsep}{2.5pt}
\caption{Hit@1 and AUROC of the features of Table~\ref{tab:temporal_features} on each CRB dataset and pooled over all four (all nine models, three error types, $n_{\mathrm{replace}}\in\{1,\ldots,5\}$). Programs: $13{,}446$ (HumanEval), $14{,}273$ (HumanEval+), $36{,}922$ (MBPP), $18{,}691$ (MBPP+), $83{,}332$ (pooled).}
\label{tab:temporal_datasets}
\begin{tabular}{lcccccccccc}
\toprule
 & \multicolumn{2}{c}{HumanEval} & \multicolumn{2}{c}{HumanEval+} & \multicolumn{2}{c}{MBPP} & \multicolumn{2}{c}{MBPP+} & \multicolumn{2}{c}{Pooled} \\
\cmidrule(lr){2-3}\cmidrule(lr){4-5}\cmidrule(lr){6-7}\cmidrule(lr){8-9}\cmidrule(lr){10-11}
Feature & Hit@1 & AUROC & Hit@1 & AUROC & Hit@1 & AUROC & Hit@1 & AUROC & Hit@1 & AUROC \\
\midrule
Single-pass & 0.359 & 0.891 & 0.176 & 0.880 & 0.324 & 0.873 & 0.340 & 0.857 & 0.308 & 0.874 \\
Flip rate & 0.150 & 0.740 & 0.169 & 0.765 & 0.135 & 0.714 & 0.161 & 0.704 & 0.149 & 0.725 \\
Convergence step & 0.142 & 0.742 & 0.095 & 0.756 & 0.133 & 0.721 & 0.146 & 0.707 & 0.131 & 0.727 \\
Conf.\ variance & 0.253 & 0.840 & 0.269 & 0.881 & 0.215 & 0.799 & 0.236 & 0.783 & 0.235 & 0.816 \\
Min.\ confidence & 0.296 & 0.893 & 0.189 & 0.895 & 0.279 & 0.868 & 0.297 & 0.856 & 0.270 & 0.874 \\
\midrule
Slope (rising) & 0.434 & 0.735 & 0.406 & 0.728 & 0.401 & 0.703 & 0.403 & 0.697 & 0.408 & 0.711 \\
\bottomrule
\end{tabular}
\end{table}

\paragraph{Per-model breakdown.}
Table~\ref{tab:temporal_models} breaks the HumanEval result down by model.
For each model it takes the best of nine trajectory-derived scores (the five features above, mean and last-pass confidence, and the two combinations) chosen after the fact, which favors the trajectory features.
They raise Hit@1 only on the five Open-dCoder-0.5B checkpoints, by $+0.001$ to $+0.045$, and on four of these five only through the combination that includes single-pass confidence.
The best purely temporal score adds $+0.005$ for the base model and $+0.045$ for the $10{,}000$-step CDLM checkpoint and lowers Hit@1 for the other three.
On all four larger models every trajectory-derived score lowers Hit@1; even the best one is $0.029$ to $0.155$ below single-pass confidence.
In AUROC, minimum confidence improves on single-pass confidence for the five 0.5B checkpoints ($+0.031$ to $+0.052$), while for the four larger models every trajectory-derived score falls below it (the best by $0.033$ to $0.041$).

\begin{table}[!htbp]
\centering
\small
\setlength{\tabcolsep}{4pt}
\caption{Per-model breakdown on CRB HumanEval (three error types, $n_{\mathrm{replace}}\in\{1,\ldots,5\}$, $1{,}484$ to $1{,}505$ programs per model). ``Best traj.''\ is the highest of nine trajectory-derived scores, each in its primary orientation, chosen per model after the fact. Superscripts name the score: c, z-scored combination of single-pass confidence, flip rate and confidence variance; m, minimum confidence; s, convergence step.}
\label{tab:temporal_models}
\begin{tabular}{lcccccc}
\toprule
 & \multicolumn{3}{c}{Hit@1} & \multicolumn{3}{c}{AUROC} \\
\cmidrule(lr){2-4}\cmidrule(lr){5-7}
Model & Single-pass & Best traj. & $\Delta$ & Single-pass & Best traj. & $\Delta$ \\
\midrule
Open-dCoder-0.5B (base) & 0.242 & 0.272$^{\mathrm{c}}$ & +0.031 & 0.840 & 0.892$^{\mathrm{m}}$ & +0.052 \\
MDLM, 2,000 steps & 0.263 & 0.267$^{\mathrm{c}}$ & +0.004 & 0.840 & 0.882$^{\mathrm{m}}$ & +0.042 \\
CDLM, 2,000 steps & 0.383 & 0.395$^{\mathrm{c}}$ & +0.012 & 0.873 & 0.906$^{\mathrm{m}}$ & +0.034 \\
CDLM, 10,000 steps & 0.433 & 0.478$^{\mathrm{m}}$ & +0.045 & 0.839 & 0.886$^{\mathrm{m}}$ & +0.047 \\
CDLM$+$ctw & 0.472 & 0.472$^{\mathrm{c}}$ & +0.001 & 0.852 & 0.884$^{\mathrm{m}}$ & +0.031 \\
\midrule
LLaDA-8B-Base & 0.175 & 0.144$^{\mathrm{s}}$ & $-$0.031 & 0.944 & 0.905$^{\mathrm{c}}$ & $-$0.038 \\
Dream-7B-Base & 0.101 & 0.071$^{\mathrm{s}}$ & $-$0.029 & 0.921 & 0.887$^{\mathrm{c}}$ & $-$0.033 \\
Fast-dLLM-v2-1.5B & 0.542 & 0.398$^{\mathrm{c}}$ & $-$0.144 & 0.949 & 0.908$^{\mathrm{c}}$ & $-$0.041 \\
Fast-dLLM-v2-7B & 0.622 & 0.467$^{\mathrm{m}}$ & $-$0.155 & 0.961 & 0.926$^{\mathrm{c}}$ & $-$0.035 \\
\bottomrule
\end{tabular}
\end{table}

\paragraph{The confidence-slope exception.}
Confidence slope, read in the orientation in which rising confidence marks a suspicious position, is the one feature with a higher HumanEval Hit@1 than single-pass confidence ($0.434$ against $0.359$).
The refinement rule explains why.
A corrupted token has low confidence in the first pass, the policy remasks it and refills it with the model's own prediction, and the new token has high confidence.
Rising confidence therefore marks positions that the first pass had already flagged.
Restricting the slope ranking to positions whose first-pass confidence is below $\tau$ gives Hit@1 $0.439$ (one tie-breaking draw; the unrestricted ranking gives $0.435$ under the same draw), so the restriction loses nothing: the feature reorders the single-pass shortlist rather than adding candidates to it.
Of the positions it ranks first, $72.1\%$ already had first-pass confidence below $\tau$ and $83.0\%$ had their token changed by the policy.
Its ranking over all positions is much weaker than that of single-pass confidence (AUROC $0.735$ against $0.891$), because positions the policy never remasks have slopes near zero and are left unordered.
What the slope detects is that the model prefers a different token at position $i$.
That information is a disagreement margin, $\max_{v\in\mathcal{V}}p_\theta(v\mid\bm{z})-p_\theta(z_i\mid\bm{z})$, which is available from the single forward pass that already produces $c_i=p_\theta(z_i\mid\bm{z})$.
The Hit@1 gain of the slope is therefore in principle obtainable at the cost of one pass.
The logged histories store only $c_i$, so we do not evaluate the margin here, and this result does not show that trajectories carry information beyond one-step confidence.

\paragraph{Limitations.}
First, the trajectory is produced by the $\tau$-gated policy itself.
A token changes only if the policy remasked it, which happens only when its confidence is below $\tau$, so flip rate and convergence step are thresholded functions of the single-pass signal rather than an independent probe.
An unconditional re-scoring of every position at every round cannot be recovered from the logged runs and would require new forward passes.
Second, when no position falls below $\tau$ the program never changes, and every trajectory feature is then constant and carries no information.
The fraction of programs with such a frozen trajectory (all four datasets) is $18.3\%$ for the Open-dCoder-0.5B base model, $20.0\%$ for MDLM, $18.5\%$ for CDLM at $2{,}000$ steps, $34.9\%$ for CDLM at $10{,}000$ steps, $43.0\%$ for CDLM$+$ctw, $3.3\%$ for LLaDA-8B-Base, and $0.0\%$ for Dream-7B-Base ($1$ of $9{,}260$ programs) and both Fast-dLLM-v2 models.
The large frozen fraction of CDLM$+$ctw is the trajectory-level counterpart of the observation in Section~\ref{sec:ctw} that clean-token supervision leaves fewer positions below $\tau$; the CDLM values here are those of the main checkpoints, whose learning-rate schedule differs from that of CDLM$+$ctw (Appendix~\ref{app:lr_horizon}).
On the $11{,}400$ HumanEval programs whose trajectory changes at least once, the population most favorable to trajectory features, single-pass confidence still has the highest Hit@1 ($0.393$; the best purely trajectory-based feature, minimum confidence, reaches $0.319$, and the combination that includes single-pass confidence $0.342$) apart from confidence slope ($0.503$), and minimum confidence again matches it in AUROC ($0.913$ against $0.911$).
Third, $p_\theta(z_i\mid\bm{z})$ does not drift across re-evaluations at the program level unless the policy has remasked something: across all $83{,}332$ programs, the confidence variance is positive at some position exactly when the flip rate is positive at some position, without exception.
At the level of a single position, confidence can still move when the position itself was never remasked, because its context changed.

%% file: appendix/release.tex
\section{Reproducibility and Release}
\label{app:release}

This appendix lists what is released, where, and under which licenses, and states which results the release reproduces exactly and which it does not.

\paragraph{Code.}
The release is the tag \texttt{v1.0.4-corrective-training} (commit \texttt{7053a1c}) of the repository \url{https://github.com/zhangshuibai/CDLM}.
It contains the CRB generator and the refinement and scoring pipeline; launchers for the CRB evaluation and the from-scratch code-generation evaluation of the 0.5B models, which check their code, inputs, model and package versions before running; the 0.5B trainer, a snapshot of Open-dLLM~\citep{opendllm2025} and VeOmni~\citep{ma2025veomni} with the mixture objective added, together with the resolved configurations of the runs behind CDLM-0.5B and MDLM-0.5B; the LLaDA-8B LoRA trainer with its CRB, localization and generation-check scripts; and the Sudoku experiment.
The release does not include the fine-tuning code of the 16B MATH experiment or of the ParallelBench experiment.
The repository is MIT-licensed, except the training framework derived from VeOmni and Open-dLLM and the bundled copy of Open-dLLM, which are under the Apache License 2.0, and two third-party Sudoku puzzle generators (MIT).

\paragraph{Models.}
Table~\ref{tab:release_artifacts} lists the released checkpoints with the revisions the release pins.
The model repositories in the table are MIT-licensed; the 0.5B models are derivatives of Open-dCoder-0.5B (Apache License 2.0), and LLaDA-8B-Base is MIT-licensed.
In our repositories, commits after the first upload changed only model cards (and, for the two 0.5B OpenCodeInstruct models, the \texttt{torch\_dtype} field of \texttt{config.json}), so the weights at the pinned revisions are the evaluated ones; the repository README lists the full revision hashes and the sha256 digest of every weight file.
CDLM-0.5B is also published as \texttt{Shuibai12138/CDLM-0.5B} with identical weights; the CRB pipeline selects the diffusion code path from the model name, so the evaluation scripts expect the \texttt{Open-Dcoder} repository name.
The checkpoints of the $\alpha$ sweep (\texttt{Shuibai12138/open-dcoder-ablation-}$\langle\alpha\rangle$, twelve values of $\alpha$) and of CDLM$+$ctw (\texttt{Shuibai12138/open-dcoder-ablation-0.1-ctw0.1}) are public as well.
The checkpoints of the seed study and of the Sudoku experiment are not released.

\begin{table}[!htbp]
\centering
\small
\caption{Released models and data on the Hugging Face Hub. Repositories without an owner prefix belong to \texttt{Shuibai12138}. Revisions are the pinned commits of the release (first eight hexadecimal digits). The two Nemotron adapter repositories also hold the 200-step runs in \texttt{step200/}. OCI: OpenCodeInstruct.}
\label{tab:release_artifacts}
\begin{tabular}{@{}lll@{}}
\toprule
\textbf{Artifact} & \textbf{Repository} & \textbf{Revision} \\
\midrule
\multicolumn{3}{@{}l}{\textit{Open-dCoder-0.5B, full fine-tuning}} \\
Base model (third party) & \texttt{fredzzp/open-dcoder-0.5B} & \texttt{d0d86d5b} \\
MDLM-0.5B & \texttt{Open-Dcoder-0.5B-baseline-mdm-step2000} & \texttt{fa5eef96} \\
CDLM-0.5B & \texttt{Open-Dcoder-0.5B-mixture-mdm-step2000} & \texttt{5a7170e7} \\
MDLM-OCI & \texttt{Open-Dcoder-0.5B-MDLM-OpenCodeInstruct} & \texttt{535b3693} \\
CDLM-OCI & \texttt{Open-Dcoder-0.5B-CDLM-OpenCodeInstruct} & \texttt{8eb87fe2} \\
\midrule
\multicolumn{3}{@{}l}{\textit{LLaDA-8B-Base, LoRA adapters}} \\
Base model (third party) & \texttt{GSAI-ML/LLaDA-8B-Base} & \texttt{0f2787f2} \\
MDLM, Nemotron & \texttt{LLaDA-8B-MDLM-LoRA} & \texttt{97cb772d} \\
CDLM, Nemotron & \texttt{LLaDA-8B-CDLM-LoRA} & \texttt{84fb1d23} \\
MDLM-OCI & \texttt{LLaDA-8B-MDLM-LoRA-OpenCodeInstruct} & \texttt{b68c4c8d} \\
CDLM-OCI & \texttt{LLaDA-8B-CDLM-LoRA-OpenCodeInstruct} & \texttt{713278bc} \\
\midrule
\multicolumn{3}{@{}l}{\textit{Data}} \\
CRB inputs (canonical) & \texttt{crb-paper-inputs} (dataset) & \texttt{a0003763} \\
OpenCodeInstruct (third party) & \texttt{nvidia/OpenCodeInstruct} (dataset) & \texttt{8f3ba5ba} \\
\bottomrule
\end{tabular}
\end{table}

\paragraph{CRB inputs.}
The canonical CRB instances are the released input files, not the generator.
They are the files of the dataset \url{https://huggingface.co/datasets/Shuibai12138/crb-paper-inputs} (CC BY 4.0), pinned at revision
\begin{center}
\small
\texttt{Shuibai12138/crb-paper-inputs} @ \texttt{a00037635943c930fa5d8e0961c7ca26127ca53f},
\end{center}
which holds the corrupted programs that were refined in the CRB evaluations of the main text and of Appendices~\ref{app:llada8b}, \ref{app:concurrent} and~\ref{app:oci}, with their error positions, original tokens and test outcomes, and the raw generator output they were produced from.
Its main groups of files are:
the pinned Open-dCoder set (60 files, $n_{\mathrm{replace}}\in\{1,\dots,5\}$, 9{,}778 corrupted programs of which 9{,}257 fail their tests), which CDLM-0.5B refines in all of its CRB results except those of Appendix~\ref{app:robustness}, and which both arms of Table~\ref{tab:ctw_ablation}, the MDLM row of Table~\ref{tab:tau_sensitivity}, the seed-study checkpoints, the self-correction-head runs of Appendix~\ref{app:prism} and the models of Appendix~\ref{app:oci} also refine;
the per-model 0.5B sets, refined by the base model and MDLM-0.5B in Tables~\ref{tab:confidence_refinement_benchmark} and~\ref{tab:mixture_training_pass_at_1_comparison}, the Base and MDLM rows of Table~\ref{tab:prism_h2h} and Figures~\ref{fig:error_localization} and~\ref{fig:pass_at_1_comparison}, by the base model in Table~\ref{tab:tau_sensitivity}, by CDLM$+$ctw in the comparisons of Appendix~\ref{app:stats} with the main checkpoint, and by the $\alpha$-sweep checkpoints in Figure~\ref{fig:mixture_prob_ablation};
the LLaDA-8B-Base set, whose 12 files at $n_{\mathrm{replace}}=1$ (3{,}184 test-failing programs) are the repair inputs of the 8B experiment, together with the raw files its localization metrics read;
and the Dream-7B-Base and Fast-dLLM-v2 sets (Appendix~\ref{app:temporal}).
The plausible-error variant, its matched random control and the natural-error sets of Appendix~\ref{app:robustness} are not part of this dataset.
The 0.5B evaluation launcher pins the earlier revision \texttt{21cae174} of the same dataset; no data file of that revision changed, and the later revision only adds the LLaDA-8B localization files and corrects the dataset card.
The source benchmarks keep their licenses (HumanEval: MIT; MBPP: CC BY 4.0; HumanEval+ and MBPP+: Apache License 2.0).
An earlier upload, \texttt{Shuibai12138/crb-datasets}, is not the input set of the paper and should not be used to reproduce it.

\paragraph{Regenerating CRB.}
The released generator iterates its candidate replacements in sorted order and is deterministic given its seed.
With the default seed it regenerates the operator files of the released sets byte for byte.
It does not regenerate the identifier and literal files: these were produced by an earlier version that iterated a Python set of candidate strings, so their random stream depended on the per-process string-hash seed (\texttt{PYTHONHASHSEED}), which was not recorded.
For the same reason the per-model 0.5B sets share their operator instances but not their identifier and literal instances (Appendix~\ref{app:stats_exact}), and the LLaDA-8B localization and repair files coincide only partly (Appendix~\ref{app:llada8b_setup}).
A set built with the released generator is a new and reproducible sample of the benchmark; the paper's numbers are reproduced from the released files.

\paragraph{What the release reproduces exactly.}
Evaluation is reproducible bit for bit at both scales.
On an A100 GPU in the pinned evaluation environment, the released CRB launcher reproduces the recorded run of CDLM-0.5B at $n_{\mathrm{replace}}=1$ exactly: every refined program, step count, step-0 confidence and per-step history of the 12{,}852 refinements (3{,}213 programs at four depths).
The released code-generation launcher reproduces all 1{,}640 recorded HumanEval completions of CDLM-0.5B with vanilla decoding.
For the LLaDA-8B adapters, refinement and localization of a fixed adapter at batch size 1 give identical outputs across GPUs, sharding and freshly built environments: re-running the 24 CRB cells of the base model reproduced all 6{,}368 refined programs of the reported sweep, and the adapters downloaded from the Hub reproduce the reported localization values exactly.
Batched inference is not bit-identical to batch size 1, so every CRB evaluation uses batch size 1.
Test execution is the only nondeterministic step: one program (HumanEval/139) runs close to the 8\,s execution timeout and can pass in one run and time out in another, which changes a 12-cell macro by at most 0.001.
This is why a rerun gives 0.285 for CDLM-0.5B at $T=4$ (Table~\ref{tab:oci_05b_crb}) where Table~\ref{tab:ctw_ablation} reports 0.286.

\paragraph{What the release does not reproduce exactly.}
Training is not bitwise reproducible over a full run at either scale, because some GPU backward kernels are nondeterministic.
For the 0.5B trainer in the pinned environment, fresh runs reproduce the recorded losses of steps 1 and 2 bitwise and differ slightly afterwards.
The released CDLM-0.5B and MDLM-0.5B weights are tied to the logged runs by a replay: loaded into the released code and trained for one more step with learning rate $10^{-30}$ on the recorded data stream, each gives a step-2001 loss bitwise equal to the one its original run logged.
No complete 2000-step rerun has been compared with the released 0.5B checkpoints.
For the 8B LoRA trainer, a run launched from the release reproduces the recorded step-1 loss and its components exactly for both arms, while the step-20 loss differs from the recorded one by $2.5\times10^{-3}$ (CDLM-OCI) and $1.3\times10^{-3}$ (MDLM-OCI) relative; across the reruns we compared, step-20 differences ranged from $5\times10^{-4}$ to $7\times10^{-3}$ relative.
Run-to-run variance of the final 8B metrics has not been measured, and the reported 8B numbers are reproduced exactly from the released adapters, not from retraining.

\paragraph{Environments and pinned inputs.}
Each part of the release has a pinned environment built on Python 3.11 and PyTorch 2.5.0.
The 0.5B trainer pins its direct requirements and, through a constraints file, all 83 packages of a verified environment; this environment differs from that of the original seed-42 runs only in the versions of \texttt{transformers}, \texttt{tokenizers} and \texttt{datasets}, and reproduces their data stream and early losses as described above.
The LLaDA-8B scripts pin all 70 packages of the environment that produced every 8B number (\texttt{transformers} 4.57.0, \texttt{peft} 0.17.1).
The evaluation launchers pin their own environment and the revisions of the evaluation datasets and of the \texttt{code\_eval} metric.
The two base models are used at the following revisions:
\begin{center}
\small
\begin{tabular}{@{}ll@{}}
\texttt{GSAI-ML/LLaDA-8B-Base} & \texttt{0f2787f2d87eac5eed8a087d5ecd24277e6255b2} \\
\texttt{fredzzp/open-dcoder-0.5B} & \texttt{d0d86d5b99960c05258bb1f8265dd91564dbac67} \\
\end{tabular}
\end{center}
The LLaDA-8B trainer and evaluators pin the first revision, which every reported 8B run loaded.
The second has been the Hub's default revision of Open-dCoder-0.5B since before the 0.5B runs; the 0.5B training launchers load the default revision, and a local copy of this revision can be passed instead to pin it.
The Nemotron-SFT-Code data are read at revision \texttt{af7991c5} of \texttt{nvidia/Nemotron-Pretraining-SFT-v1}, with the shard order of the 0.5B runs recorded in the release, and OpenCodeInstruct at revision \texttt{8f3ba5ba}.
Nemotron-Pretraining-SFT-v1 is gated and is not redistributed; it is needed only to retrain the Nemotron models, and Appendix~\ref{app:oci} replicates both comparisons on OpenCodeInstruct (CC BY 4.0) for that reason.

%% file: checklist.tex
\section*{NeurIPS Paper Checklist}

\begin{enumerate}

\item {\bf Claims}
    \item[] Question: Do the main claims made in the abstract and introduction accurately reflect the paper's contributions and scope?
    \item[] Answer: \answerYes{}
    \item[] Justification: The abstract and the contributions list in the Introduction state three claims: (i) CRB as a controlled executable benchmark; (ii) a correction-oriented post-training principle that supervises visible corrupted tokens but not clean ones, with the clean-token comparison run at 0.5B only (Sections~\ref{sec:main_comparison}--\ref{sec:robustness}, including the LLaDA-8B transfer); (iii) improved generation on HumanEval(+) at 0.5B, MATH at 16B and the high-entropy ParallelBench average under remask decoding, stated together with the MBPP(+) regression relative to base and the absence of a from-scratch gain at 8B. Each claim is substantiated by the corresponding section and its scope is stated there.

\item {\bf Limitations}
    \item[] Question: Does the paper discuss the limitations of the work performed by the authors?
    \item[] Answer: \answerYes{}
    \item[] Justification: A ``Limitations'' paragraph appears in the Conclusion section, with a longer discussion in Appendix~\ref{app:limitations}: (a) fixed-length, in-place editing, a property of the remask-based decoder that CRB inherits; (b) localized errors only, since a model's own failing generations are far from a correct program; (c) ablations only at 0.5B and single training runs at 8B and 16B; (d) two of the three probed backbones are adapted from autoregressive checkpoints; (e) Sudoku uses an oracle editable set; (f) threshold-based refinement can run to its full step budget; (g) no from-scratch generation gain at 8B and an MBPP(+) regression at 0.5B.

\item {\bf Theory assumptions and proofs}
    \item[] Question: For each theoretical result, does the paper provide the full set of assumptions and a complete (and correct) proof?
    \item[] Answer: \answerNA{}
    \item[] Justification: The paper is empirical and does not include formal theorems or proofs. The training objective (Eq.~\ref{eq:mixture-loss}) and refinement procedure (Algorithm~\ref{alg:confidence_refinement}) are precisely specified.

    \item {\bf Experimental result reproducibility}
    \item[] Question: Does the paper fully disclose all the information needed to reproduce the main experimental results of the paper to the extent that it affects the main claims and/or conclusions of the paper (regardless of whether the code and data are provided or not)?
    \item[] Answer: \answerYes{}
    \item[] Justification: Appendix~\ref{app:hparams} lists, for the 0.5B, LLaDA-8B and 16B runs, the base checkpoint, data, steps, batch layout, optimizer, learning-rate schedule and horizon, mixture parameters and seed wherever they were recorded (entries not recorded for the 16B runs, mostly for ParallelBench, are marked ``--''). Appendices~\ref{app:oci}, \ref{app:prism} and~\ref{app:sudoku_exp} give the OpenCodeInstruct, PRISM and Sudoku settings, and Appendix~\ref{app:objective_impl} gives the loss of each trainer. Decoding settings are in Sections~\ref{sec:phenomenon} and~\ref{sec:experiments} and Appendix~\ref{app:scratch_eval}; Appendix~\ref{app:release} states which results the release reproduces exactly.

\item {\bf Open access to data and code}
    \item[] Question: Does the paper provide open access to the data and code, with sufficient instructions to faithfully reproduce the main experimental results, as described in supplemental material?
    \item[] Answer: \answerYes{}
    \item[] Justification: Code (CRB generator, refinement and scoring pipeline, 0.5B trainer, LLaDA-8B LoRA trainer and evaluation launchers) is released at \url{https://github.com/zhangshuibai/CDLM} (tag \texttt{v1.0.4-corrective-training}); the exact CRB instances of the main-text, LLaDA-8B and OpenCodeInstruct evaluations are released as the Hugging Face dataset \texttt{Shuibai12138/crb-paper-inputs} (the plausible-error and natural-error sets of Appendix~\ref{app:robustness} are not included), and the 0.5B and 8B checkpoints on the Hugging Face Hub (Appendix~\ref{app:release}). The fine-tuning code of the 16B MATH and ParallelBench experiments is not released.

\item {\bf Experimental setting/details}
    \item[] Question: Does the paper specify all the training and test details (e.g., data splits, hyperparameters, how they were chosen, type of optimizer) necessary to understand the results?
    \item[] Answer: \answerYes{}
    \item[] Justification: Section~\ref{sec:experiments} specifies the controlled training setup (matched data, optimizer, schedule and step count between MDLM and CDLM) and the decoding threshold; Appendix~\ref{app:hparams} gives all hyperparameters, and Appendices~\ref{app:llada8b}, \ref{app:scratch_eval} and~\ref{app:cdlm_parallelbench} the 8B, from-scratch generation and ParallelBench protocols.

\item {\bf Experiment statistical significance}
    \item[] Question: Does the paper report error bars suitably and correctly defined or other appropriate information about the statistical significance of the experiments?
    \item[] Answer: \answerYes{}
    \item[] Justification: Appendix~\ref{app:stats} reports two variance components for the 0.5B comparisons. Finite-problem-sample uncertainty: exact McNemar tests and 10{,}000-resample paired bootstrap 95\% confidence intervals over per-problem outcomes, with Benjamini--Hochberg correction across cells. Training-seed variance: three training seeds per arm, reported as mean $\pm$ standard deviation (not standard error). The paired tests cover one deterministic (greedy) decoding run per arm; the 8B and 16B comparisons are single training runs (the 8B Pass@1 comparisons and the Hit@1 comparison on the 497 repaired programs carry exact McNemar tests, Appendix~\ref{app:llada8b}), and ParallelBench has no per-sample outcomes to pair.

\item {\bf Experiments compute resources}
    \item[] Question: For each experiment, does the paper provide sufficient information on the computer resources (type of compute workers, memory, time of execution) needed to reproduce the experiments?
    \item[] Answer: \answerYes{}
    \item[] Justification: Appendix~\ref{app:hparams} (compute paragraph and Table~\ref{tab:hparams}) gives the hardware of each run. Each 2000-step 0.5B run of the main comparison took 57--65 minutes of training on 4 A100-40GB GPUs (the OpenCodeInstruct runs about 19 minutes, CDLM$+$ctw 30 minutes, the additional seeds 2 GPUs; the $\alpha$ sweep used 4 H200 GPUs). Each 2000-step LLaDA-8B LoRA run took 46--56 minutes on 4 A100-40GB GPUs. The 16B MATH runs used 8 H100-80GB GPUs for one epoch (9{,}306 steps); their wall-clock time and the hardware of the 16B ParallelBench runs were not recorded. The project also included preliminary and discarded runs that are not reported.
    
\item {\bf Code of ethics}
    \item[] Question: Does the research conducted in the paper conform, in every respect, with the NeurIPS Code of Ethics \url{https://neurips.cc/public/EthicsGuidelines}?
    \item[] Answer: \answerYes{}
    \item[] Justification: The work uses publicly available models and datasets (one gated dataset used under its license, and one commercial API model, gpt-4.1-nano, used as a baseline under its terms of use), involves no human subjects, and adheres to the NeurIPS Code of Ethics.

\item {\bf Broader impacts}
    \item[] Question: Does the paper discuss both potential positive societal impacts and negative societal impacts of the work performed?
    \item[] Answer: \answerYes{}
    \item[] Justification: An ``Impact'' paragraph at the end of the Conclusion section discusses both directions: better error awareness may make DLMs more reliable, while they can still produce incorrect outputs and need appropriate safeguards.
    
\item {\bf Safeguards}
    \item[] Question: Does the paper describe safeguards that have been put in place for responsible release of data or models that have a high risk for misuse (e.g., pre-trained language models, image generators, or scraped datasets)?
    \item[] Answer: \answerNA{}
    \item[] Justification: The released artifact is a code-correction benchmark (CRB) and post-training recipe applied to small to medium-scale code models; we do not release new generative models with materially elevated misuse risk relative to the pretrained checkpoints we build on.

\item {\bf Licenses for existing assets}
    \item[] Question: Are the creators or original owners of assets (e.g., code, data, models), used in the paper, properly credited and are the license and terms of use explicitly mentioned and properly respected?
    \item[] Answer: \answerYes{}
    \item[] Justification: Every third-party asset is cited where it is used. Datasets: HumanEval (MIT), MBPP (CC BY 4.0), HumanEval+ and MBPP+ (Apache 2.0), MATH (MIT) and its MATH-500 split, ParallelBench (public test set from its authors), Nemotron-Math-v2 (CC BY 4.0), OpenCodeInstruct (CC BY 4.0) and the code subset of Nemotron-Pretraining-SFT-v1 (gated, NVIDIA data license; used under its terms and not redistributed). Models: Open-dCoder-0.5B (Apache 2.0), LLaDA-8B-Base (MIT), Dream-v0-Base-7B (Apache 2.0), LLaDA2.0-mini-preview and LLaDA2.0-mini (Apache 2.0), and Fast-dLLM-v2 (temporal study). gpt-4.1-nano is accessed through the OpenAI API under its terms of use. Pinned versions are in Appendix~\ref{app:release}.

\item {\bf New assets}
    \item[] Question: Are new assets introduced in the paper well documented and is the documentation provided alongside the assets?
    \item[] Answer: \answerYes{}
    \item[] Justification: The new assets are the Code Revision Benchmark (CRB), documented in Section~\ref{sec:crb} and released with its generator and the instances used in the main-text, LLaDA-8B and OpenCodeInstruct evaluations, and the 0.5B and 8B checkpoints; Appendix~\ref{app:release} lists repositories, pinned revisions and licenses.

\item {\bf Crowdsourcing and research with human subjects}
    \item[] Question: For crowdsourcing experiments and research with human subjects, does the paper include the full text of instructions given to participants and screenshots, if applicable, as well as details about compensation (if any)?
    \item[] Answer: \answerNA{}
    \item[] Justification: This work involves no crowdsourcing or human-subject research.

\item {\bf Institutional review board (IRB) approvals or equivalent for research with human subjects}
    \item[] Question: Does the paper describe potential risks incurred by study participants, whether such risks were disclosed to the subjects, and whether Institutional Review Board (IRB) approvals (or an equivalent approval/review based on the requirements of your country or institution) were obtained?
    \item[] Answer: \answerNA{}
    \item[] Justification: This work involves no human-subject research and therefore does not require IRB review.

\item {\bf Declaration of LLM usage}
    \item[] Question: Does the paper describe the usage of LLMs if it is an important, original, or non-standard component of the core methods in this research? Note that if the LLM is used only for writing, editing, or formatting purposes and does \emph{not} impact the core methodology, scientific rigor, or originality of the research, declaration is not required.
    \item[] Answer: \answerYes{}
    \item[] Justification: \texttt{gpt-4.1-nano}~\citep{openai2025gpt41} is used only as the autoregressive baseline in the AR-vs-DLM comparison (Appendix~\ref{app:ar_baseline}); it is not part of the proposed method. LLM coding agents were used to implement and orchestrate evaluation pipelines and to assist with writing; no reported number comes from an LLM judgment (Appendix~\ref{app:llm_use}).

\end{enumerate}